\def\year{2021}\relax
\documentclass[letterpaper]{article} 
\usepackage{aaai21}  
\usepackage{times}  
\usepackage{helvet} 
\usepackage{courier}  
\usepackage[hyphens]{url}  
\usepackage{graphicx} 
\usepackage{natbib}  
\usepackage{caption} 

\usepackage{amsmath} 
\usepackage{nicefrac}       
\usepackage{microtype}      
\usepackage{graphicx}
\usepackage{caption}
\usepackage{amsmath}
\usepackage{amssymb}
\usepackage{multirow}
\usepackage{booktabs}
\usepackage{bm}
\usepackage{mathtools}
\usepackage{lscape} 
\usepackage{listings} 

\usepackage[utf8]{inputenc} 
\usepackage[T1]{fontenc}    
\usepackage{url}            
\usepackage{booktabs}       
\usepackage{amsfonts}       
\usepackage{nicefrac}       
\usepackage{microtype}      
\usepackage{graphicx}
\usepackage{caption}
\usepackage{amsmath}
\usepackage{amssymb}
\usepackage{multirow}
\usepackage{bm}
\usepackage{mathtools}
\usepackage{amsmath}
\usepackage{subfig}
\usepackage[ruled,vlined]{algorithm2e}
\usepackage{graphicx}
\usepackage{caption}
\usepackage{enumerate} 
\usepackage[export]{adjustbox}
\usepackage{subfig}
\usepackage{tabularx}
\usepackage[bottom]{footmisc}
\usepackage{tablefootnote}

\usepackage[switch]{lineno} 
\usepackage{enumitem,kantlipsum}
\SetKw{KwBy}{by}
\usepackage{mathtools}

\makeatletter
\define@key{Gin}{resolution}{\pdfimageresolution=#1\relax}
\newcommand*\bigcdot{\mathpalette\bigcdot@{0.65}}
\newcommand*\bigcdot@[2]{\mathbin{\vcenter{\hbox{\scalebox{#2}{$\m@th#1\bullet$}}}}}
\makeatother

\nocopyright
\title{Why have a Unified Predictive Uncertainty? \\ Disentangling it using Deep Split Ensembles}
\author {
    Utkarsh Sarawgi\textsuperscript{\rm 1}, Wazeer Zulfikar\textsuperscript{\rm 1}, Rishab Khincha\textsuperscript{\rm 1, 2}, Pattie Maes\textsuperscript{\rm 1} \\
}

\affiliations{
    \textsuperscript{\rm 1}Massachusetts Institute of Technology, USA \\
    \textsuperscript{\rm 2}BITS Pilani Goa Campus, India \\
    \{utkarshs, wazeer, rkhincha, pattie\}@mit.edu
}

\begin{document}

\maketitle 

\begin{abstract}
Understanding and quantifying uncertainty in black box Neural Networks (NNs) is critical when deployed in real-world settings such as healthcare. Recent works using Bayesian and non-Bayesian methods have shown how a unified predictive uncertainty can be modelled for NNs. Decomposing this uncertainty to disentangle the granular sources of heteroscedasticity in data provides rich information about its underlying causes. We propose a conceptually simple non-Bayesian approach, \textit{deep split ensemble}, to disentangle the predictive uncertainties using a multivariate Gaussian mixture model. The NNs are trained with clusters of input features, for uncertainty estimates per cluster. We evaluate our approach on a series of benchmark regression datasets, while also comparing with unified uncertainty methods. Extensive analyses using dataset shits and empirical rule highlight our inherently well-calibrated models. Our work further demonstrates its applicability in a multi-modal setting using a benchmark Alzheimer's dataset and also shows how deep split ensembles can highlight hidden modality-specific biases. The minimal changes required to NNs and the training procedure, and the high flexibility to group features into clusters makes it readily deployable and useful. The source code is available at \\{\lstinline|https://github.com/wazeerzulfikar/deep-split-ensembles|}

\end{abstract}

\section{Introduction}\label{intro}
\begin{figure*}[t]
\centering
\subfloat[Unified]{\includegraphics[width=0.24\linewidth, height=2.9cm, valign=t]{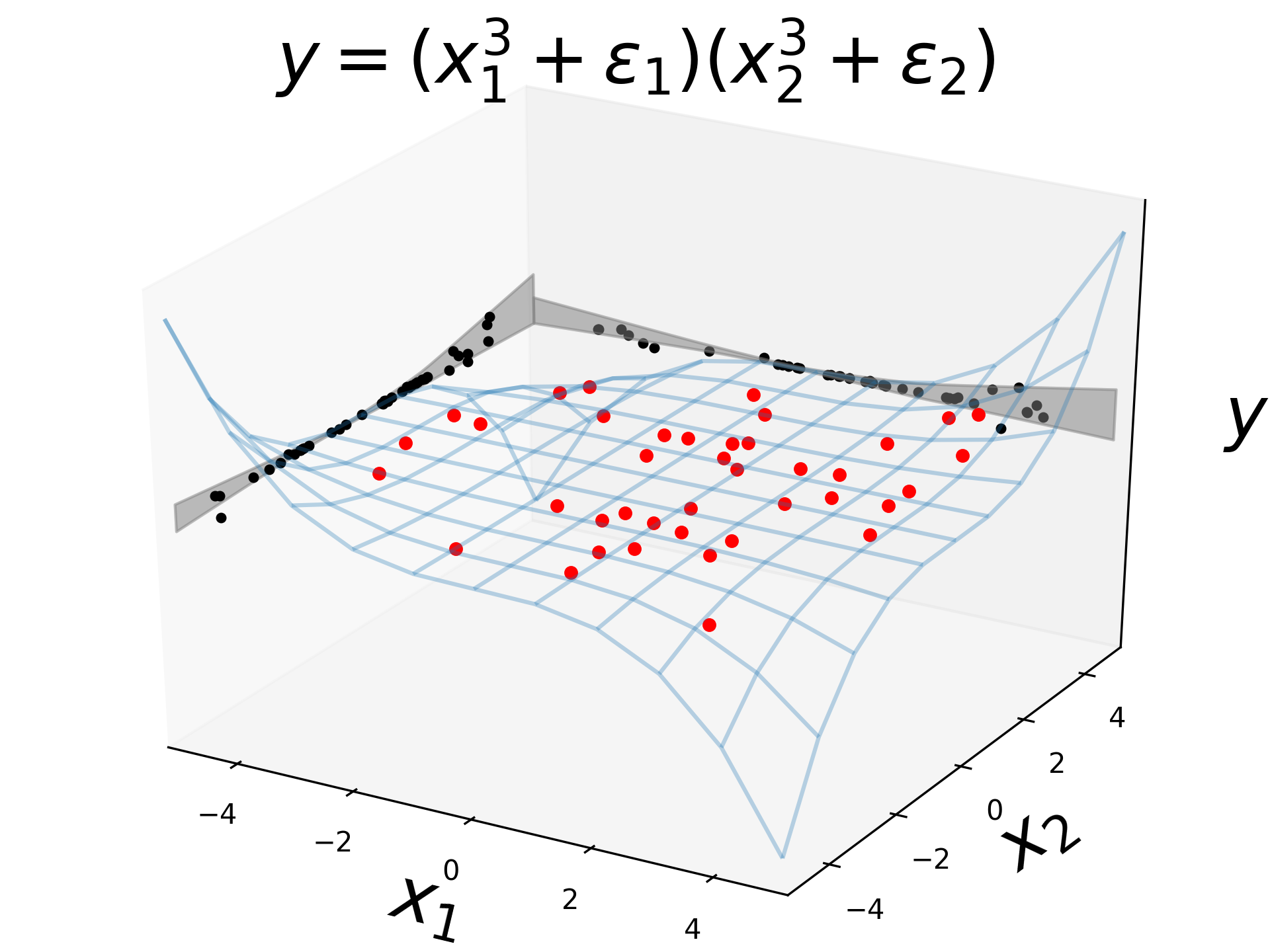}}
    ~
\subfloat[Disentangled]{\includegraphics[width=0.24\linewidth, height=2.9cm, valign=t]{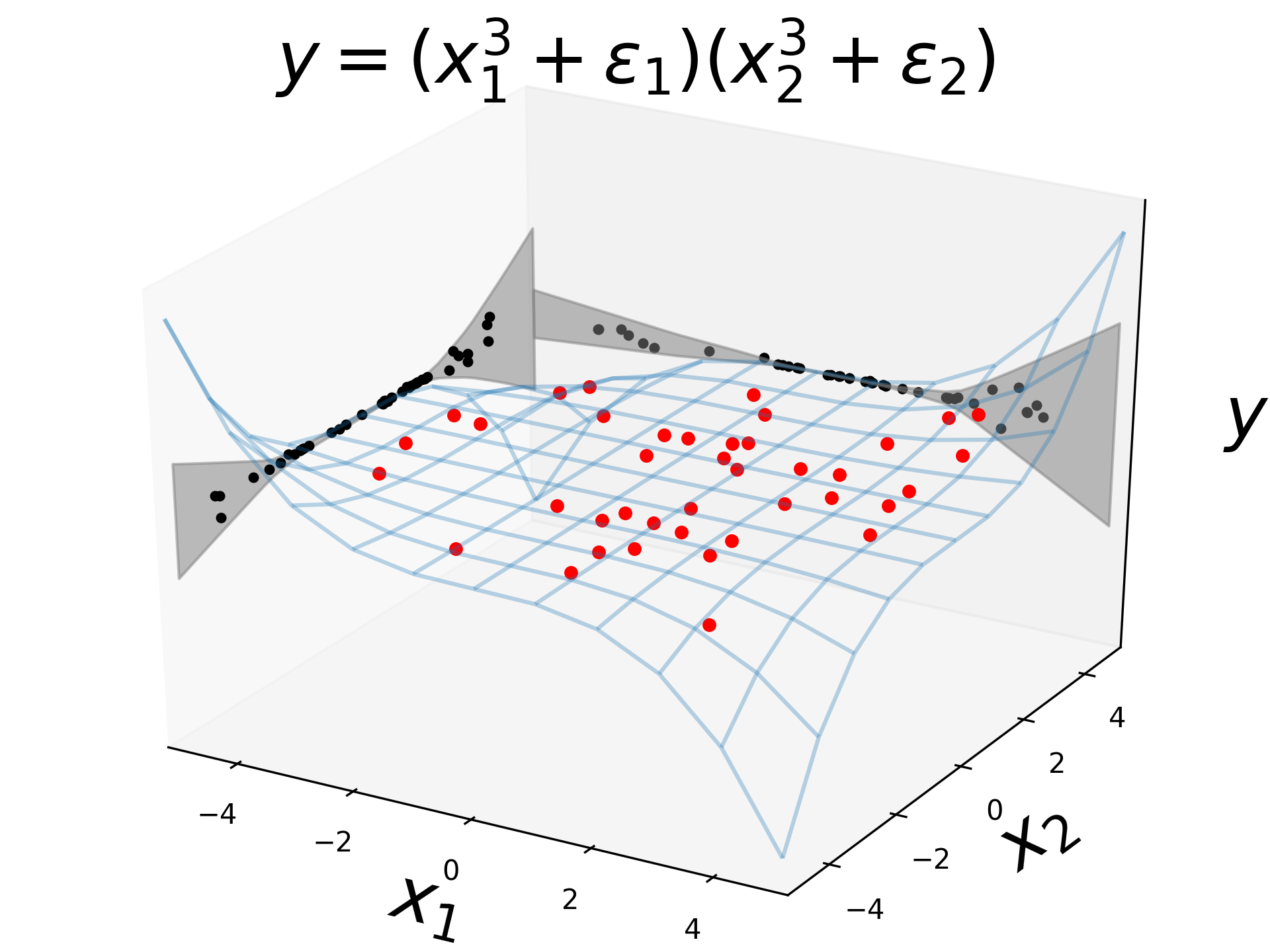}}
    ~
\subfloat[Unified]{\includegraphics[width=0.24\linewidth, height=2.9cm, valign=t]{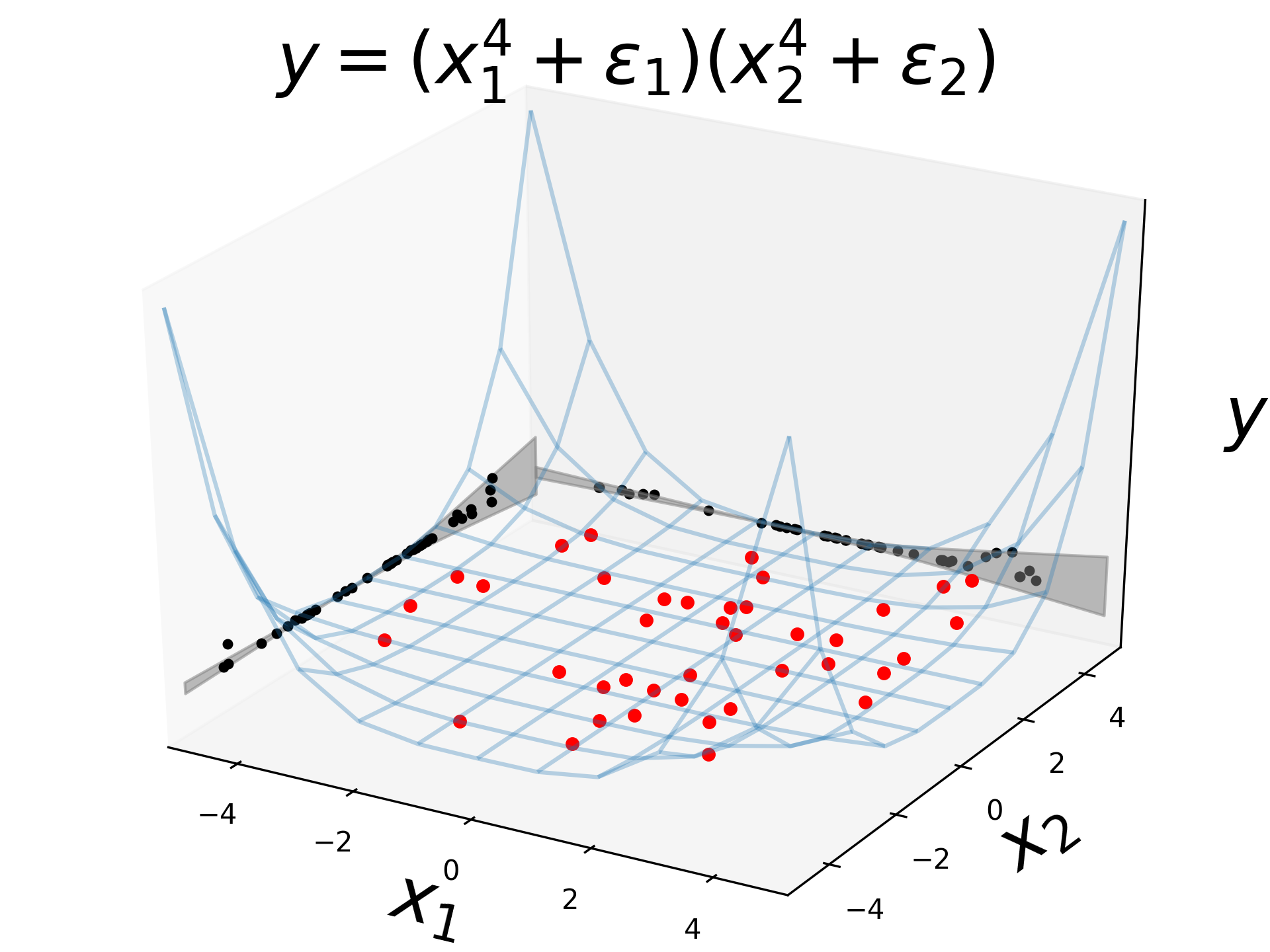}}
    ~
\subfloat[Disentangled]{\includegraphics[width=0.24\linewidth, height=2.9cm, valign=t]{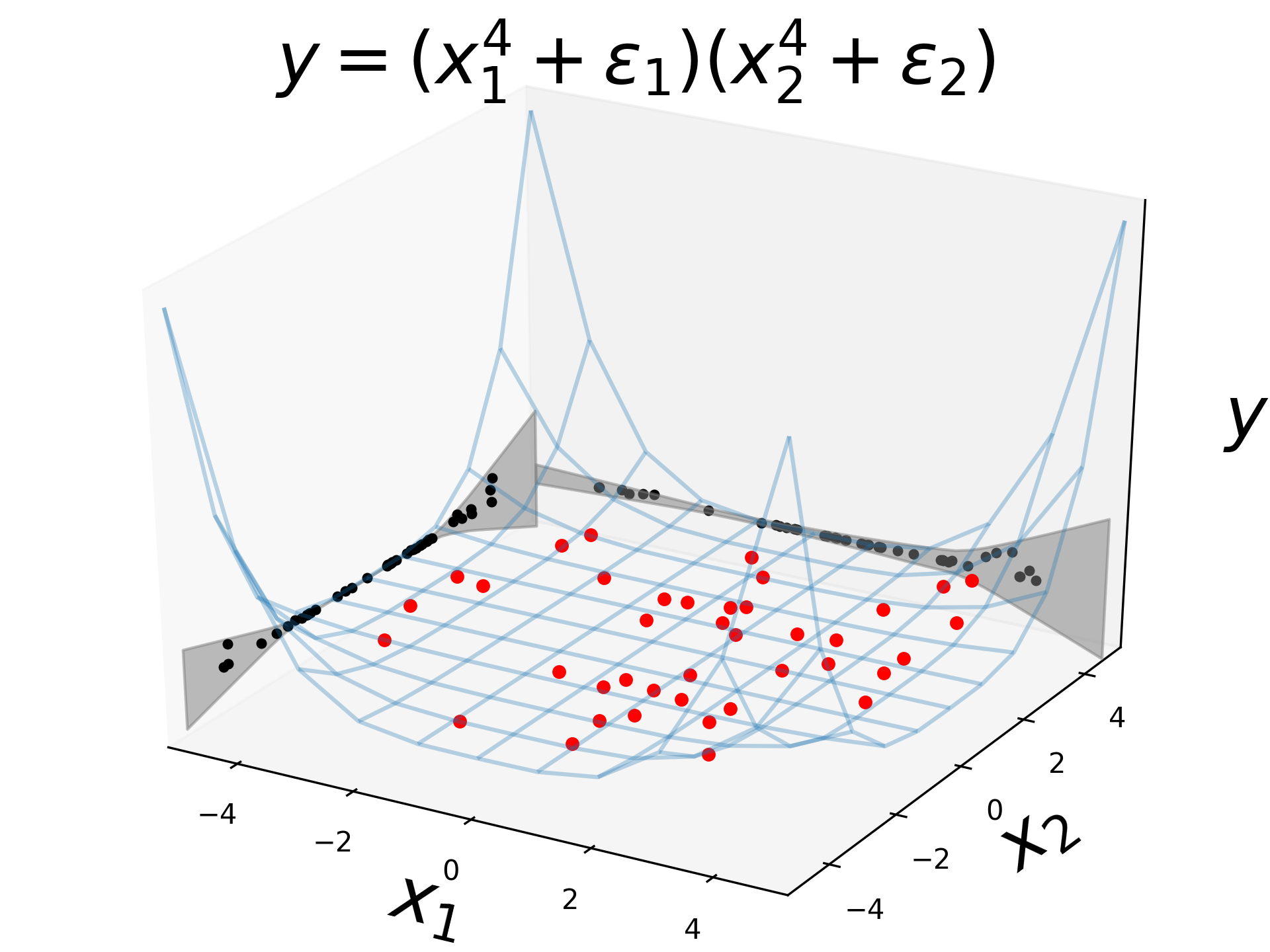}}
\caption{Examples of modelling unified vs disentangled uncertainties on 2 two-dimensional toy regression datasets, $y=(x_1^3+\epsilon_1)(x_2^3+\epsilon_2)$ and $y=(x_1^4+\epsilon_1)(x_2^4+\epsilon_2)$ where $\epsilon_1, \epsilon_2 \thicksim \mathcal{N}(0, 3^2)$. The red points are the observed noisy training samples and the black points are their projections on the respective axes. The grey regions show the predicted mean along with three standard deviations. The disentangled uncertainties have `different' grey regions on the two dimensions (as opposed to unified), and are able to `contain' the black points better. This illustrates how decomposed uncertainties can capture disentangled information about the individual noise in the input features.}
\label{toy_results}
\end{figure*}
Vast developments across a variety of machine learning tasks have led to extensive deployment of neural networks (NNs) in safety-critical applications ranging from medical diagnosis to self-driving cars \cite{lecun2015deep}. For reliable, fair and aware models in many regression tasks, the point prediction is not sufficient; the uncertainty or the confidence of that prediction must also be estimated by the model. Understanding what a model does not know is critical to using machine learning systems and mitigating plausible biases and risks in decision making \cite{gal2016uncertainty}.

Heteroscedasticity can be modelled as the changing variability of the random disturbance in output values given the input features; in other terms, the probabilistic variability introduced by the stochastic data generation processes. A `unified' predictive uncertainty would be a single estimate modelled for all input features together. Multiple probabilistic methods have been proposed to quantify the same. Bayesian approximation techniques such as variational inference (VI) \cite{graves2011practical, blundell2015varinf}, expectation propagation \cite{hernandez2015probabilistic}, dropout-based VI \cite{kingma2015dropout, gal2016dropout}, NNs as Gaussian processes \cite{lee2017deep}, deterministic VI \cite{wu2018deterministic}, Bayesian model averaging in low-dimensional parameter subspaces \cite{izmailov2020subspace}, and approximate Bayesian ensembling \cite{pearce2020uncertainty} have been shown to be quite useful in modelling the uncertainties in NNs. Non-Bayesian approaches \cite{osband2016risk, lakshminarayanan2017simple, dusenberry2020analyzing, jain2020maximizing} that involve bootstrapping and ensembling multiple probabilitic NNs have shown performances comparable to Bayesian methods with reduced computational costs and modifications to the training procedure. \citet{ashukha2020pitfalls} performed a broad study of ensembling techniques in context of uncertainty estimation. \citet{Qiu2020Quantifying} proposed a framework using residual estimation with an I/O kernel (RIO) to estimate uncertainty in any pretrained standard NN. Almost all previous works \cite{mackay1992laplace, kay1999statistics, welling2011stochastic, kendall2017uncertainties, shridhar2018uncertainty, snoek2019can} including the ones above estimate a unified predictive uncertainty. However, a single `unified' uncertainty would fundamentally be unable to distinguish the granular sources of heteroscedasticity in data, which is critical in applications such as healthcare as it can provide rich information about the underlying causes. `Disentangled' predictive uncertainties would be able to separate these tied sources with granular uncertainty estimates.

We propose a conceptually simple non-Bayesian approach, \textit{deep split ensemble}, to disentangle the predictive uncertainties using a multivariate Gaussian mixture model while training NNs with clusters of correlated features. These correlations can be statistical, or based on different input modalities (multi-modal), domain knowledge or user needs. Figure \ref{toy_results} shows application on simple examples using a multi-dimensional toy regression dataset (Section \ref{toy_sec}), highlighting an advantage of modelling disentangled predictive uncertainties over unified uncertainties.

Recent works have shown how NNs are usually overconfident at predicting probability estimates representative of the true likelihood \cite{guo2017calibration}. One can use the model's confidence on a target distribution to compare it with its accuracy and adjust the predictions \cite{platt1999calib, guo2017calibration, kuleshov2018accurate}. However, the distribution over this observed data may shift and eventually be very different once a model is deployed in practice. Robustness of uncertainty estimation under these conditions of distributional shift is necessary for the safe deployment of machine learning systems \cite{amodei2016concrete, varshney2017safety, kumar2019verified, thiagarajan2020building}.  \citet{snoek2019can} recently showed how post-hoc calibration can fail under even a mild shift in the data. We show that modelling disentangled predictive uncertainties using our approach produces inherently well-calibrated estimates per cluster of features, without any post-hoc calibration. We assess it using a granular feature-wise distributional shift. This helps address the critical and practical concerns of risk, uncertainty, and trust in a model’s output.


\paragraph{Summary of contributions:}
\begin{enumerate}
    \item To our knowledge, we are the first to propose a method to disentangle unified predictive uncertainties with NNs.
    \item We perform a rigorous and comprehensive evaluation on the inherent calibration of our models under dataset shifts on benchmark regression datasets.
    \item To demonstrate the applicability of our method, we extend it to include domain knowledge, and to a multi-modal setting to highlight the potential hidden modality-specific biases.
\end{enumerate}
\section{Deep Split Ensembles: Disentangling predictive uncertainties}
\subsection{Notation and setup}\label{setup}
Let $\mathbf{x} \in \mathbb{R}^d$ represent a set of $d$-dimensional input features and $y \in \mathbb{R}$ denote the real-valued label for regression.  Given a training dataset $\mathcal{D} = \{(\mathbf{x}_n, y_n)\}_{n=1}^N$ consisting of N i.i.d. samples, we model the probabilistic predictive distribution $p_\theta (y|\mathbf{x})$ using a neural network with parameters $\theta$.

We split the set of $d$ input features of $\mathbf{x}$ into $k$ exhaustive clusters, $k\neq1$, each containing $m_i$ number of features, s.t. $\forall^{k}_{i=1} m_i < d$ and $\sum_{i=1}^k m_i = d$. Features are non-overlapping i.e. a particular feature belongs to only one cluster. Let $\mathbf{c}^i_n \in \mathbb{R}^{m_i}$ denote $i^{th}$ cluster containing $m_i$ input features of $n^{th}$ data point $\mathbf{x}_n$. Thus, $\{(\mathbf{c}^i_n, y_n)\}_{n=1}^N$ represents  $i^{th}$ input feature cluster and corresponding label for N samples. Label $y_n$ is the same across any input cluster $c^i_n$ corresponding to the $n^{th}$ data point (Equation \ref{params2}). For clusters with one feature each, $k = d \implies \forall^{k}_{i=1} m_i = 1$.
\begin{figure}[h]
\centering
\subfloat{\includegraphics[width=\linewidth, height=4cm, valign=t]{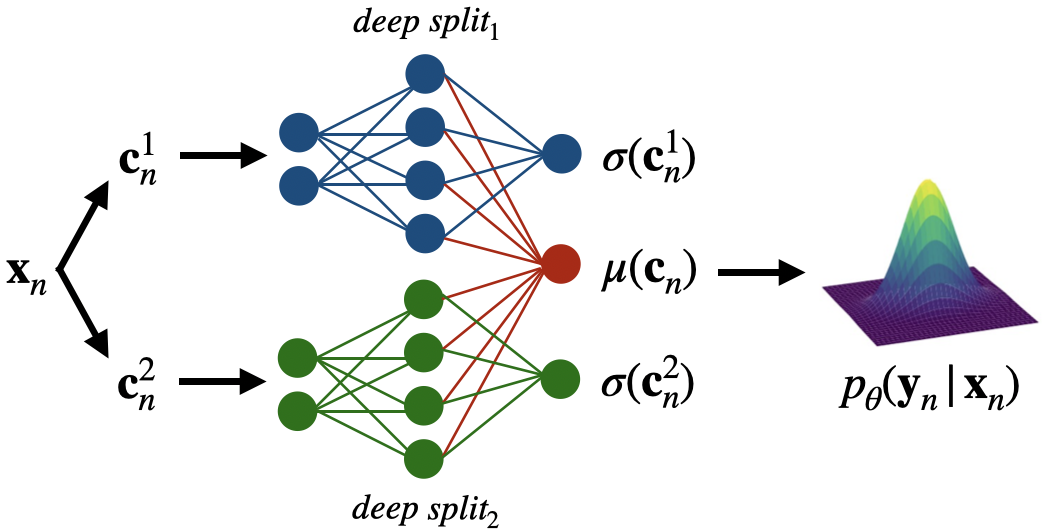}}
\caption{A representative Deep Split NN model architecture}
\label{gauss_model_c}
\end{figure}
\subsection{Defining `Deep Split Ensemble' with multivariate Gaussian mixture}\label{motivation}

\citet{jacobs1991adaptive, xu1995alternative} showed how local expert networks can be explicitly trained on differing input features and used a gating network to combine into a mixture of experts model. \citet{hinton1999products} introduced the product of experts model in which several individual probabilistic models are combined together to model the data. \citet{williams2002products} further considered each expert as a Gaussian for a richer structure. Recent works have shown improvements in performance of ensembles by training them jointly with a unified loss as compared to post-hoc ensembling of independent models \cite{lee2015m, furlanello2018born, dutt2020coupled}. Such mixture of experts have been widely used to predict the target value only. We model each expert to predict an uncertainty given the corresponding input features, while also contributing to a single target value prediction (Figure \ref{gauss_model_c}), trained with a unified loss (Section \ref{training}). This allows for disentangled predictive uncertainties per cluster of input features, as opposed to a single unified predictive uncertainty.

Our method forms clusters of input features (Section \ref{training}), and splits NNs proportionately. Each split, referred to as a `deep split', takes in a cluster of input features $\mathbf{c}^i_n$ and models its output as the predicted mean $\mu_\theta(\mathbf{c}^i_n)$ and the standard deviation $\sigma_\theta(\mathbf{c}^i_n)$ parameterizing a Gaussian distribution. However, only one common mean $\mu_\theta(\mathbf{c}_n)$ is learnt across the deep splits (Equation \ref{params1}). This is because while we aim to disentangle the predictive uncertainties, we still want the model to learn the regression value using all the input features together. We combine these Gaussian output distributions of deep splits to parametrize a multivariate Gaussian (MVN) $p_\theta(\mathbf{y}_n|\mathbf{x}_n)$; $\mathbf{y}_n \thicksim \mathcal{N}\left(\boldsymbol\mu_\theta(\mathbf{x}_n),  \Sigma_\theta(\mathbf{x}_n)\right)$, using the common mean and the covariance matrix $\Sigma_\theta(\mathbf{x}_n)$ represented as a diagonal matrix of the $k$ individual variances (Equations \ref{params2} and \ref{mv}). We call this entire model `deep split NN' (Figure \ref{gauss_model_c}).


\begin{multline}
\label{params1}
\boldsymbol\mu_\theta(\mathbf{x}_n) = [\mu_\theta(\mathbf{c}^1_n),\dots,\mu_\theta(\mathbf{c}^k_n)]^{T}\text{; }\\\forall_{i=1}^k\hspace{2mm}\mu_\theta(\mathbf{c}^i_n) = \mu_\theta(\mathbf{c}_n)
\end{multline}

\begin{multline}
\label{params2}
\Sigma_\theta(\mathbf{x}_n) = \text{diag}(\sigma^2_\theta(\mathbf{c}^1_n),\dots,\sigma^2_\theta(\mathbf{c}^k_n)) \\
\text{}\mathbf{y}_n = [y_n^1,\dots,y_n^k]^T\text{\hspace{2mm}}\text{;\hspace{2mm}}\forall_{i=1}^k\hspace{2mm} y_n^i = y_n
\end{multline}

\begin{multline}
\label{mv}
p_\theta(\mathbf{y}_n|\mathbf{x}_n) = \frac{1}{\sqrt{(2\pi)^{k}|\Sigma_\theta(\mathbf{x}_n)|}} \times\\ 
\text{ exp} \left(-\frac{1}{2}\left(\mathbf{y}_n-\boldsymbol\mu_\theta(\mathbf{x}_n)\right)^T
\Sigma_\theta(\mathbf{x}_n)^{-1}(\mathbf{y}_n-\boldsymbol\mu_\theta(\mathbf{x}_n))\right)
\end{multline}

The weight update equations of a network modelling a Gaussian output distribution, as derived by \citet{nix1994estimating}, show that the learning rate $\eta$ is affected by variations in $\sigma^2(\mathbf{x})$; $\eta$ is effectively amplified for input patterns where $\sigma^2(\mathbf{x})$ is smaller than average compared to patterns where $\sigma^2(\mathbf{x})$ is larger than average. While this behavior is noted across the datapoints, it is further applicable in our network across deep splits modelling different $\sigma^2(\boldsymbol c^i)$, assisting cluster-wise gradient backpropagation through our model. This biases the allocation of the network's resources towards lower-noise regions, discounting regions of the input space where the network is producing larger than average uncertainties, thus implementing a form of robust regression.

\citet{lakshminarayanan2017simple, snoek2019can, dusenberry2020analyzing, dietterich2000ensemble, pearce2020uncertainty} have shown improved performance with an ensemble of an NN initialized with random parameters, compared to the NN performance. We train a parallel ensemble of our deep split NNs while uniformly weighing the predictions across the ensemble. This forms a mixture of uniformly-weighted multivariate Gaussians.

We call the above method the `deep split ensemble', and train it using a simple procedure.
\begin{table*}[t]
\small
  \centering
  \begin{tabular}{lcccccccc}
    \toprule
    Datasets & \multicolumn{4}{c}{RMSE} & \multicolumn{4}{c}{NLL} \\
    \cmidrule(r){2-5}
    \cmidrule(r){6-9}
     & \multirow{2}{*}{RIO} & Deep & Anchored & Deep Split & \multirow{2}{*}{RIO} & Deep & Anchored & Deep Split\\
     &  & Ensembles & Ensembling & Ensembles & & Ensembles & Ensembling & Ensembles\footnotemark\\ 
    \midrule
     Boston & --  & 3.28 $\pm$ 1.00 & 3.09 $\pm$ 0.17 & \textbf{2.53 $\pm$ 0.15} & -- & 2.41 $\pm$ 0.25 & 2.52 $\pm$ 0.05 & \textbf{2.23 $\pm$ 0.04} \\
     Concrete & 5.97 $\pm$ 0.48  & 6.03 $\pm$ 0.58 & 4.87 $\pm$ 0.11 & \textbf{4.40 $\pm$ 0.10} & 3.24 $\pm$ 0.10 & 3.06 $\pm$ 0.18 & 2.97 $\pm$ 0.02 & \textbf{2.85 $\pm$ 0.02} \\
     Energy & 0.70 $\pm$ 0.38 & 2.09 $\pm$ 0.29 & \textbf{0.35 $\pm$ 0.01} & 0.41 $\pm$ 0.02 & 1.03 $\pm$ 0.35 & 1.38 $\pm$ 0.22 & 0.96 $\pm$ 0.13 & \textbf{0.28 $\pm$ 0.11} \\
     Kin8nm & -- & 0.09 $\pm$ 0.00 & \textbf{0.07 $\pm$ 0.00} & 0.19 $\pm$ 0.00 & -- & \textbf{-1.20 $\pm$ 0.02} & -1.09 $\pm$ 0.01 & -0.20 $\pm$ 0.02 \\
     Naval & --  & \textbf{0.00 $\pm$ 0.00} & \textbf{0.00 $\pm$ 0.00} & \textbf{0.00 $\pm$ 0.00} & -- & -5.63 $\pm$ 0.05 & \textbf{-7.17 $\pm$ 0.03} & -5.28 $\pm$ 0.02 \\
     Power & 4.05 $\pm$ 0.12 & 4.11 $\pm$ 0.17 & 4.07 $\pm$ 0.04 & \textbf{4.04 $\pm$ 0.05}  & 2.81 $\pm$ 0.03 & 2.79 $\pm$ 0.04 & 2.83 $\pm$ 0.01 & \textbf{2.78 $\pm$ 0.01} \\
     Protein & 4.08 $\pm$ 0.06 & 4.71 $\pm$ 0.06 & 4.36 $\pm$ 0.02 & \textbf{4.05 $\pm$ 0.03} & 2.82 $\pm$ 0.01 & 2.83 $\pm$ 0.02 & 2.89 $\pm$ 0.01 & \textbf{2.76 $\pm$ 0.00} \\
     Wine & 0.67 $\pm$ 0.03 & 0.64 $\pm$ 0.04 & 0.63 $\pm$ 0.01 & \textbf{0.60 $\pm$ 0.02} & 1.09 $\pm$ 0.10 & 0.94 $\pm$ 0.12 & 0.95 $\pm$ 0.01 & \textbf{0.89 $\pm$ 0.02} \\
     Yacht & 1.46 $\pm$ 0.49 & 1.58 $\pm$ 0.48 & \textbf{0.57 $\pm$ 0.05} & 0.86 $\pm$ 0.07 & 1.79 $\pm$ 0.88 & 1.18 $\pm$ 0.21 & \textbf{0.37 $\pm$ 0.08} & 0.90 $\pm$ 0.09 \\
    \bottomrule
  \end{tabular}
  \caption{Results on UCI regression benchmark datasets comparing RMSE and NLL}
  \label{base_comp_table}
\end{table*}

\subsection{Training procedure}\label{training}
\paragraph{Feature clustering and NN splitting:} The input feature space is required to be split into $k$ exhaustive clusters. For splitting, we use hierarchical clustering based on Pearson correlation distance, since we want to estimate predictive uncertainties for clusters of similar features. The dendograms thus obtained upon hierarchical clustering with complete linkage are thresholded relative to the maximum distance to obtain feature clusters (details in Appendix \ref{appendix_hc}). Note that we are clustering features, which should not be confused with clustering datapoints. The NN is then split to train using all feature clusters (Section \ref{motivation} and Figure \ref{gauss_model_c}). We also show splitting based on domain knowledge (Section \ref{human}) and across input modalities (Section \ref{multimodal}). 
\paragraph{Training criterion:} The deep split ensemble is trained with the clusters of input features of all training datapoints and their corresponding ground truth labels using a proper scoring rule $l(\theta, \mathbf{x}, \mathbf{y})$. We optimize for the negative log-likelihood (NLL) of the joint distribution $p_\theta(\mathbf{y}|\mathbf{x})$ according to Equation \ref{NLL}.
\begin{multline}\label{NLL}
    -\log p_\theta(\mathbf{y}_n|\mathbf{x}_n) = \frac{1}{2}\big(\log \Sigma_\theta(\mathbf{x})\\ + \left(\mathbf{y}-\boldsymbol\mu_\theta(\mathbf{x})\right)^T \Sigma_\theta(\mathbf{x})^{-1}\left(\mathbf{y}-\boldsymbol\mu_\theta(\mathbf{x})\right)\big) + \text{ constant}
\end{multline}
\paragraph{Parallel ensembling:} As discussed in Section \ref{motivation}, we train a parallel ensemble of our deep split models initialized with random NN parameters, while uniformly weighing the predictions across the ensemble. This forms a mixture of uniformly-weighted multivariate Gaussians $p_E(\mathbf{y}_n|\mathbf{x}_n)$ as shown in Equation \ref{mix}, where $E$ is the total number of models in the ensemble. The mean $\boldsymbol\mu_E(\mathbf{x}_n)$ and variance $\Sigma_E(\mathbf{x}_n)$ of such a mixture $E^{-1}\sum_e \mathcal{N}\left(\boldsymbol\mu_{\theta_e(\mathbf{x}_n)}, \Sigma_{\theta_e(\mathbf{x}_n)}\right)$ are shown in Equations \ref{mu_mix} and \ref{sigma_mix} respectively (refer to Appendix \ref{appendix_musigma} for derivation). For ease of computing quantiles and predictive probabilities, we approximate this ensemble prediction as a Gaussian whose mean and variance are that of the mixture. For a fair comparison with other uncertainty estimation methods that use ensembles of 5 models \cite{lakshminarayanan2017simple, pearce2020uncertainty}, we use a value of $E=5$ for our experiments (Section \ref{exp_and_results}). Refer to Appendix \ref{appendix_variants_results} for results with $E=1$ and $E=10$
\begin{equation}
\label{mix}
p_E(\mathbf{y}_n|\mathbf{x}_n) = E^{-1} \sum_{e=1}^E {p_\theta}_e(\mathbf{y}_n|\mathbf{x}_n)
\end{equation}

\begin{equation}
\label{mu_mix}
\boldsymbol\mu_E(\mathbf{x}_n) = E^{-1} \sum_{e=1}^E{\boldsymbol\mu_\theta}_e(\mathbf{x}_n)
\end{equation}

\begin{multline}
\label{sigma_mix}
\Sigma_E(\mathbf{x}_n) =\text{diag}(\sigma^2_E(\mathbf{c}^1_n),\dots,\sigma^2_E(\mathbf{c}^k_n))\text{\hspace{3mm}}\\
\forall_{i=1}^k\hspace{3mm}\sigma^2_E(\mathbf{c}^i_n) = E^{-1}\sum_{e=1}^E\big({\sigma_\theta}_e^2(\mathbf{c}^i_n) + {\mu_\theta}_e^2(\mathbf{c}^i_n)\big) - \mu_E^2(\mathbf{c}^i_n)
\end{multline}

\section{Experiments and results}\label{exp_and_results}
\subsection{Experimental setup and evaluation metrics}\label{exp_setup}
For a fair comparison with current state-of-the-art methods for predictive uncertainty estimation using NNs on benchmark regression tasks, we use the same experimental setup. The network consists of 50 hidden units with ReLU for smaller datasets split into 20 train-test folds and 100 units with ReLU for the larger `Protein' dataset (5 folds). See Appendix \ref{appendix_regdata} for other training hyperparameters. We measure the NLL, a proper scoring rule and widely used metric for evaluating predictive uncertainty \cite{quinonero2005evaluating}. We also compute the root mean squared error (RMSE) to measure the performance of the single mean prediction of our model. 
\footnotetext{NLLs of Deep Split Ensembles in Table \ref{base_comp_table} are averaged over feature clusters of corresponding datasets. Refer to Appendix \ref{appendix_nlls} for an exhaustive list of cluster-wise predictive uncertainty estimates for all datasets.}
\subsection{Regression datasets: Toy regression and UCI regression benchmarks}\label{toy_sec}
To highlight the need for disentangled uncertainties, we first evaluate the performance of our method on an extension of the toy regression dataset setup and used to evaluate probabilistic backpropagation (PBP) \cite{hernandez2015probabilistic}, deep ensembles \cite{lakshminarayanan2017simple}, and anchored ensembling \cite{pearce2020uncertainty} which consists of 20 examples drawn from $y=(x^3+\epsilon)$ where $\epsilon \thicksim \mathcal{N}(0, 3^2)$. The extended multi-dimensional toy regression datasets, contain 40 examples each drawn from $y=(x_1^3+\epsilon_1)(x_2^3+\epsilon_2)$ and $y=(x_1^4+\epsilon_1)(x_2^4+\epsilon_2)$ where $\epsilon_1, \epsilon_2 \thicksim \mathcal{N}(0, 3^2)$. It can be observed that in the case of the unified uncertainty estimates, the underlying heteroscedasticity along each input feature can not be captured individually. However, using the same model architecture, our method can produce `different' uncertainties for each input feature (Figure \ref{toy_results}).

We then evaluate and compare our approach to current state-of-the-art methods for predictive uncertainty estimation using NNs on UCI regression benchmark datasets (see Appendix \ref{appendix_regdata} for details on datasets). Table \ref{base_comp_table} shows the comparison with the latest and competitive methods - RIO \cite{Qiu2020Quantifying}, deep ensembles \cite{lakshminarayanan2017simple} and anchored ensembling \cite{pearce2020uncertainty}; see Appendix \ref{appendix_vipbp} for a full comparison with other methods - VI \cite{graves2011practical}, PBP \cite{hernandez2015probabilistic}, MC-dropout \cite{gal2016dropout}, deterministic VI (DVI) \cite{wu2018deterministic} and subspace inference (SI) \cite{izmailov2020subspace}. We observe that our method outperforms the existing methods on multiple datasets, while also disentangling the predictive uncertainties.

We also highlight the performance of our proposed MVN model trained with a unified loss, as compared to post-hoc ensembling (vanilla mixture of experts) of independent cluster-wise NN models. This serves as the baseline for a comparison of disentangled uncertainty estimates through the corresponding RMSE and cluster-wise NLL. For each dataset, we train deep ensemble per input cluster (DEPC) and anchored ensembling per input cluster (AEPC) following their respective training procedure. This results in a prediction and an uncertainty estimate per cluster. We then average the predictions across the clusters for the final prediction. Hence, we have an NLL per cluster and a single RMSE for the final prediction. Table \ref{depc_aepc} shows that deep split ensembles outperform DEPC and AEPC on NLL of every cluster as well as RMSE. We attribute this to the joint training of cluster-wise experts in deep split ensembles, as compared to the independent training of cluster-wise experts in DEPC and AEPC.


\begin{table*}[t]
\centering
\begin{tabular}{lccccccc}
\toprule
Datasets & \multicolumn{3}{c}{RMSE} & Clusters & \multicolumn{3}{c}{Cluster-wise NLL} \\
\cmidrule(r){2-4}
\cmidrule(r){6-8}
 & DEPC & AEPC & Deep Split Ens. & & DEPC & AEPC & Deep Split Ens. \\
 \midrule
\multirow{3}{*}{Boston} 
 & \multirow{3}{*}{5.11 $\pm$ 1.06} & \multirow{3}{*}{4.93 $\pm$ 1.03} & \multirow{3}{*}{\textbf{2.53 $\pm$ 0.15}} & 1 & 2.91 $\pm$ 0.16 & 3.87 $\pm$ 0.82 & \textbf{2.23 $\pm$ 0.04} \\
 & & & & 2 & 2.82 $\pm$ 0.16 & 3.99 $\pm$ 0.94 & \textbf{2.20 $\pm$ 0.03} \\
 & & & & 3 & 3.29 $\pm$ 0.10 & 4.23 $\pm$ 1.06 & \textbf{2.26 $\pm$ 0.05} \\ \midrule
\multirow{3}{*}{Concrete} 
 & \multirow{3}{*}{10.24 $\pm$ 0.85} 
 & \multirow{3}{*}{10.40 $\pm$ 0.93} 
 & \multirow{3}{*}{\textbf{4.40 $\pm$ 0.10}} 
 & 1 & 3.77 $\pm$ 0.05 & 5.75 $\pm$ 0.61 & \textbf{2.84 $\pm$ 0.02} \\
 & & & & 2 & 3.79 $\pm$ 0.09 & 5.68 $\pm$ 0.60 & \textbf{2.85 $\pm$ 0.02} \\
 & & & & 3 & 3.80 $\pm$ 0.05 & 5.83 $\pm$ 0.61 & \textbf{2.87 $\pm$ 0.01} \\ 
 \bottomrule
\end{tabular}
\caption{Results of deep ensemble per input cluster (DEPC), anchored ensembling per input cluster (AEPC), and deep split ensembles on UCI regression benchmark datasets\footnotemark  comparing RMSE and cluster-wise NLL.}
\label{depc_aepc}
\end{table*}
\begin{figure*}[ht!]
\centering
\subfloat{%
    \includegraphics[width=0.01\linewidth, height=2.5cm, valign=t]{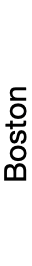}}\hspace{0.005mm}
    ~   
\subfloat{%
    \includegraphics[width=0.23\linewidth, valign=t]{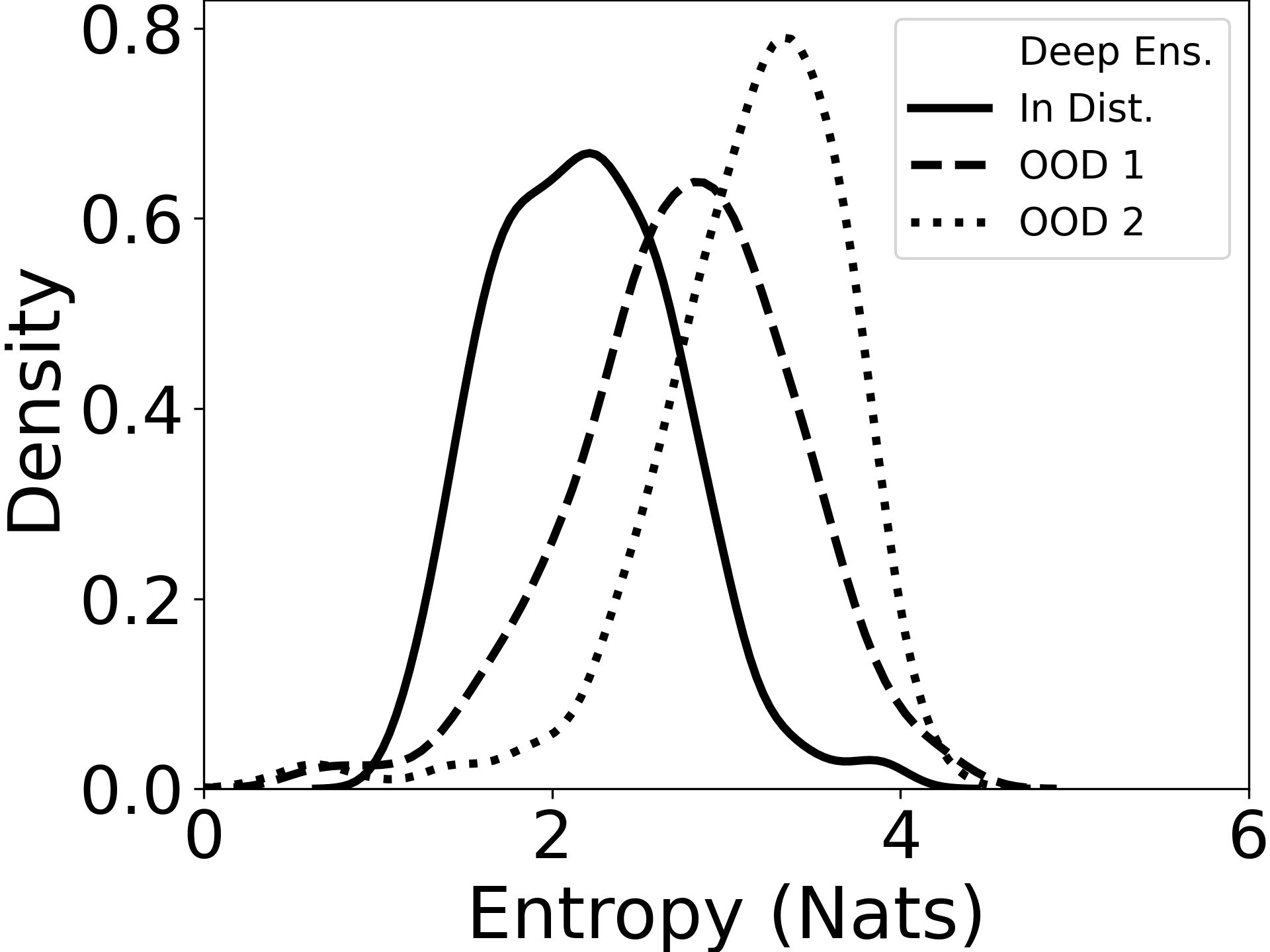}}
    ~
\subfloat{%
    \includegraphics[width=0.23\linewidth, valign=t]{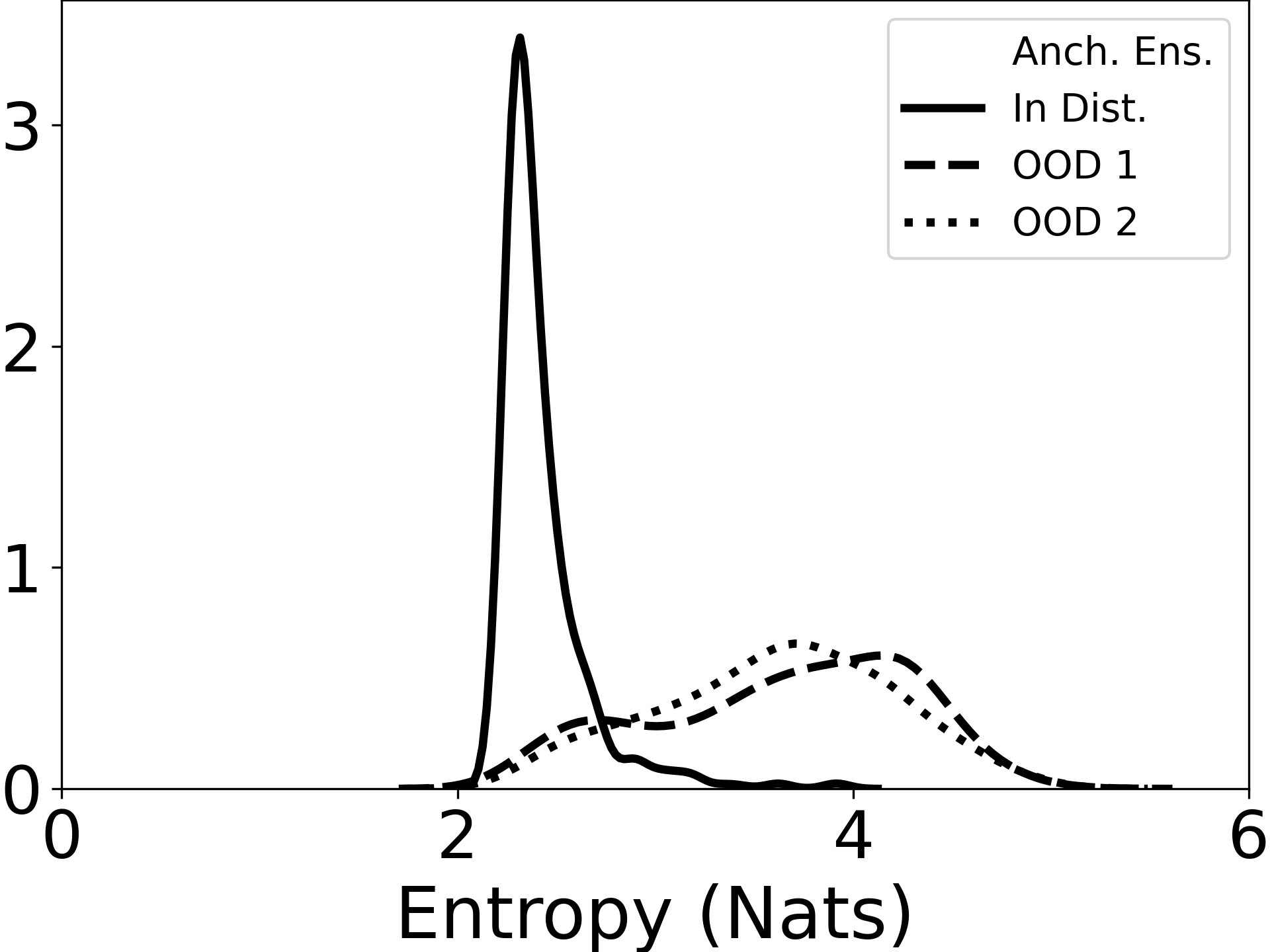}}
    ~
\subfloat{%
    \includegraphics[width=0.23\linewidth, valign=t]{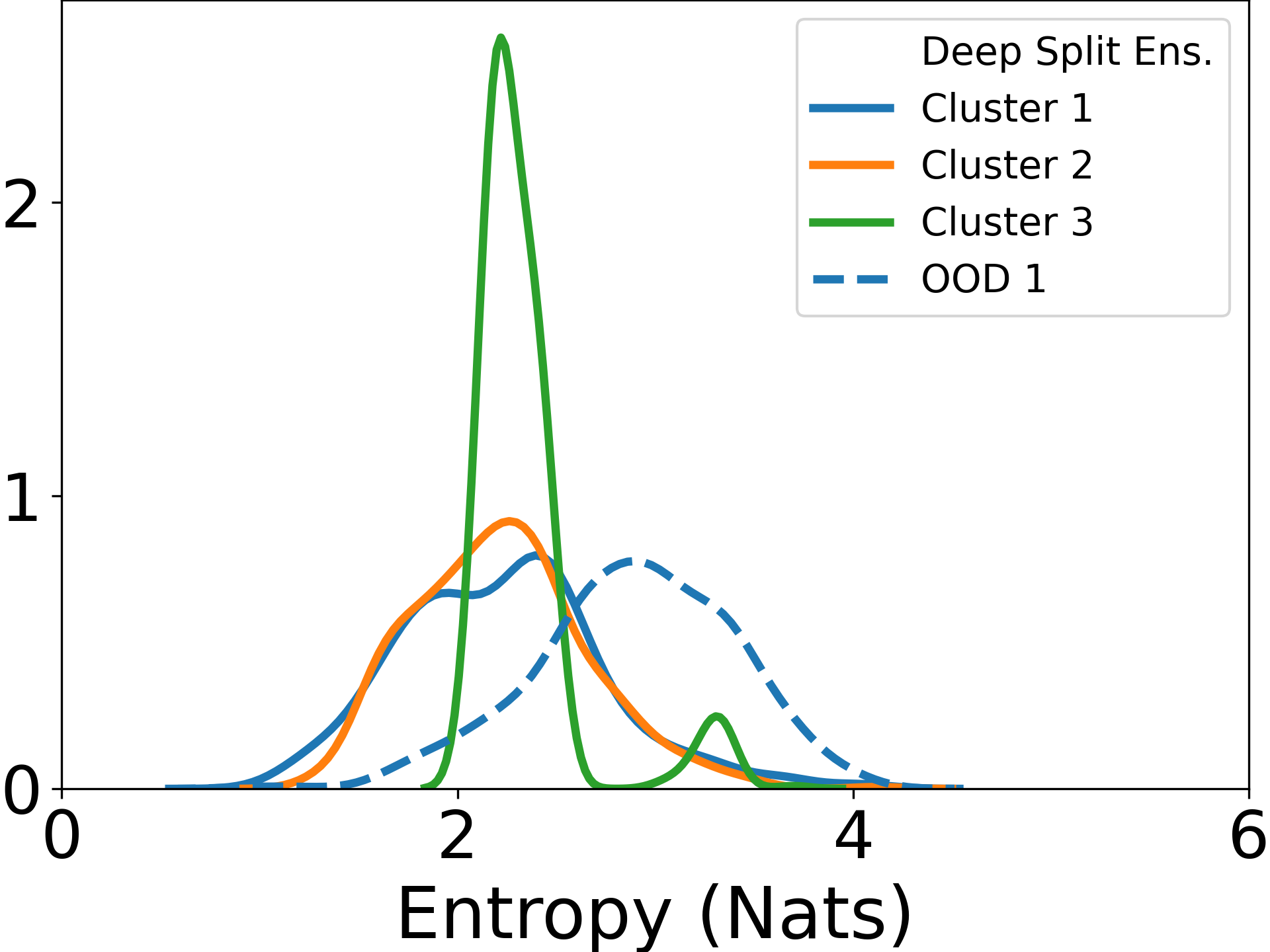}}
    ~
\subfloat{%
    \includegraphics[width=0.23\linewidth, valign=t]{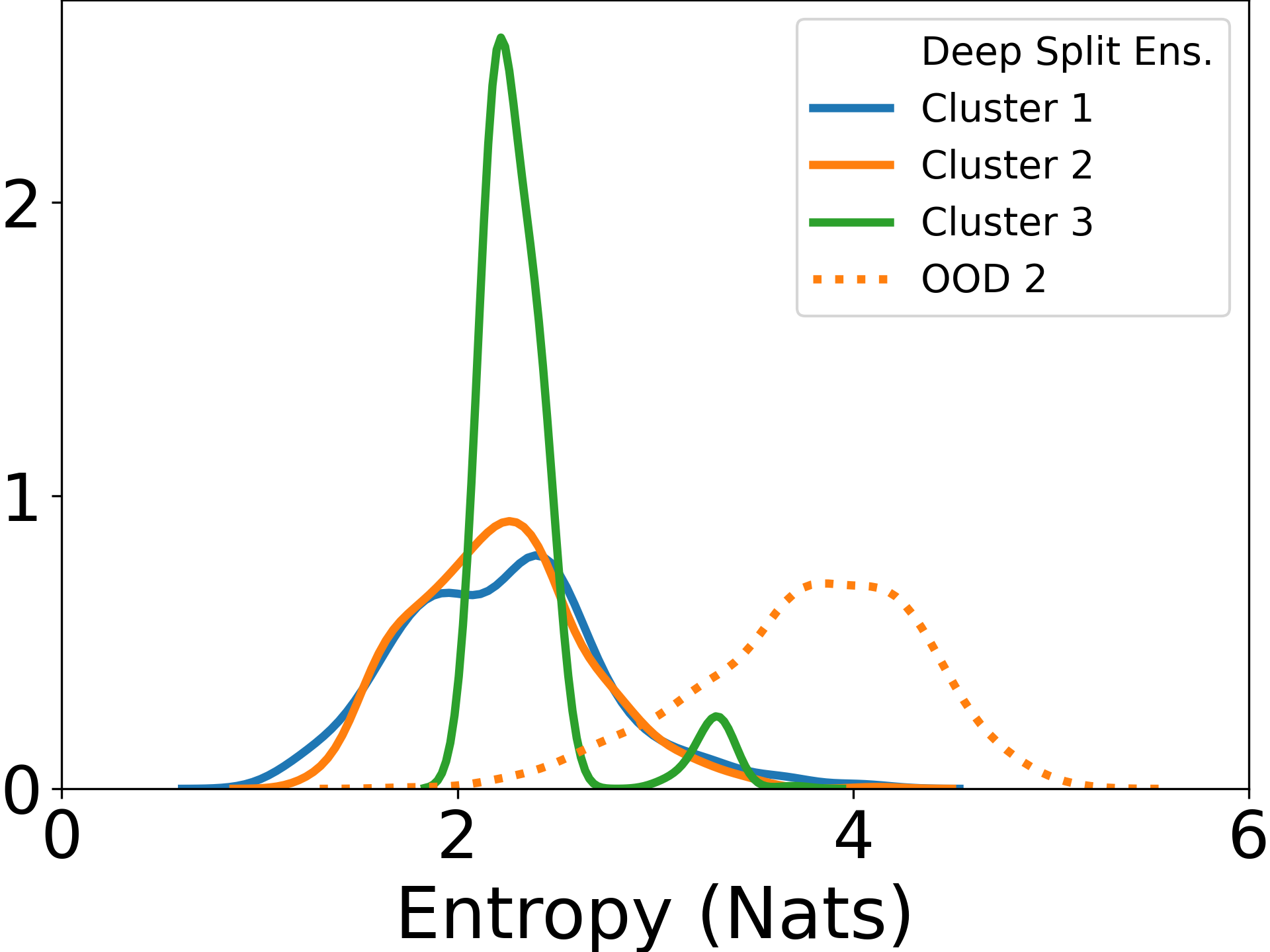}}

\subfloat{%
    \includegraphics[width=0.01\linewidth, height=2.8cm, valign=t]{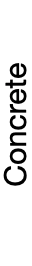}}\hspace{0.005mm}
    ~
\subfloat{%
    \includegraphics[width=0.23\linewidth, valign=t]{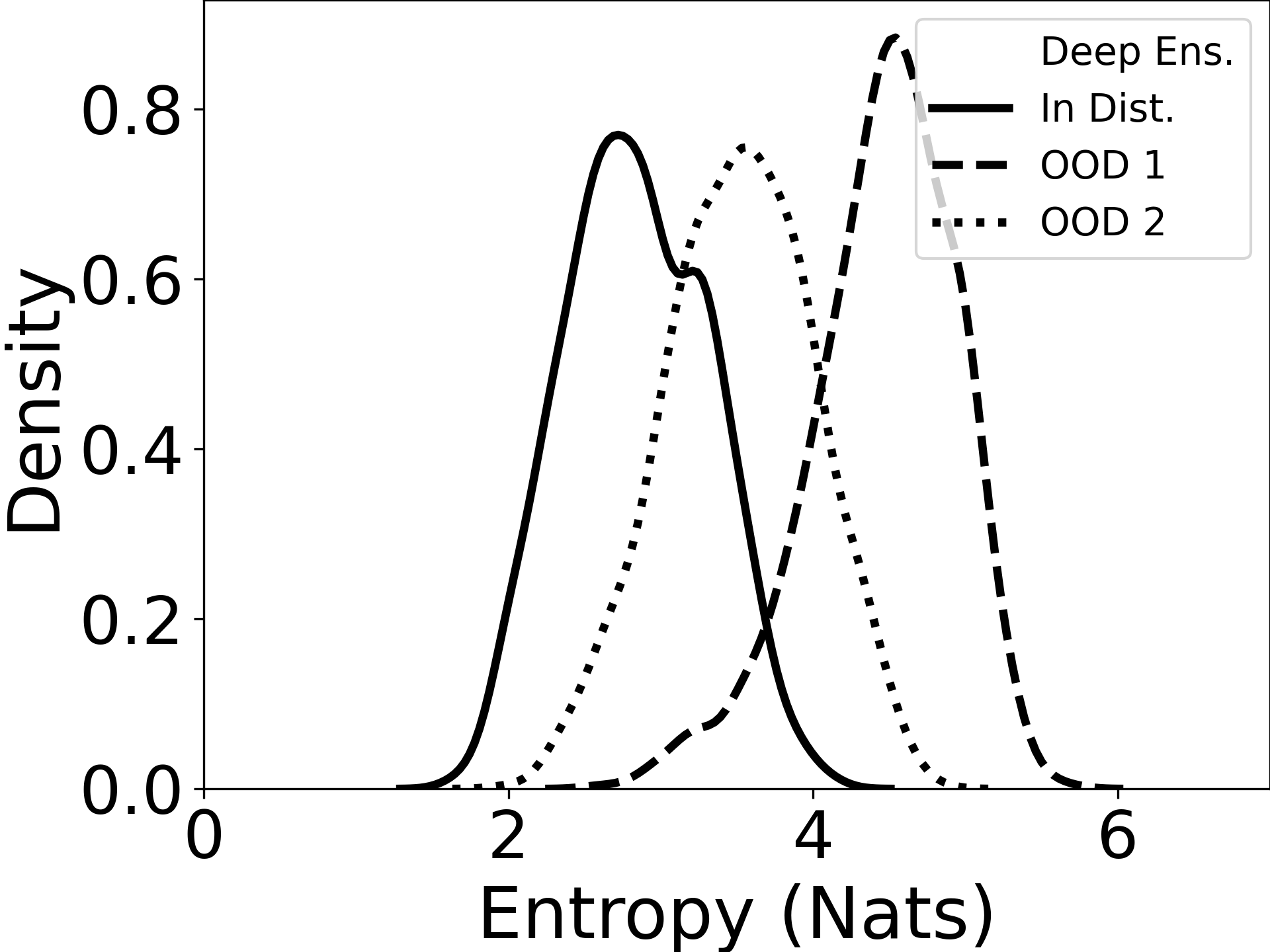}}
    ~
\subfloat{%
    \includegraphics[width=0.23\linewidth, valign=t]{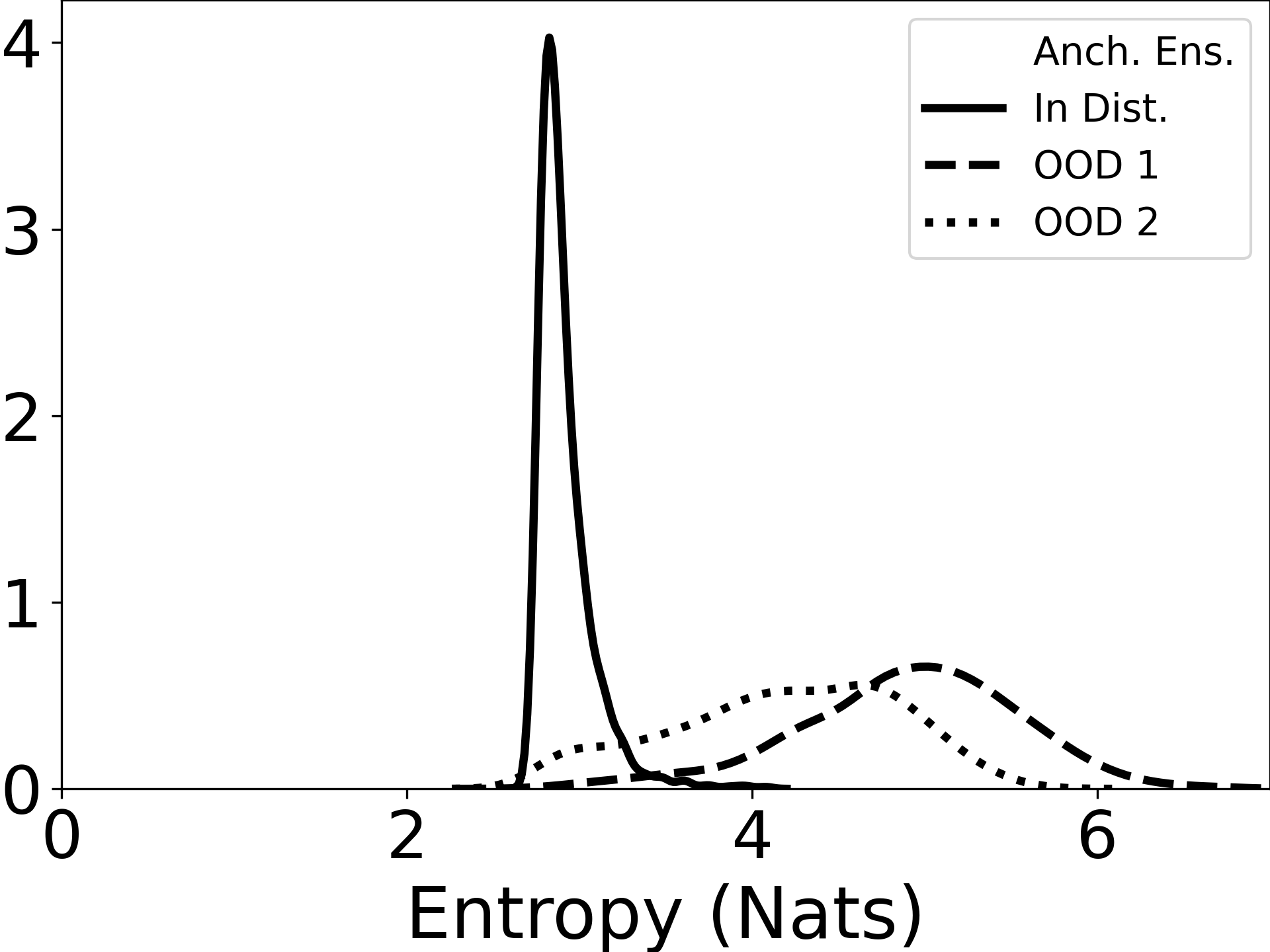}}
    ~
\subfloat{%
    \includegraphics[width=0.23\linewidth, valign=t]{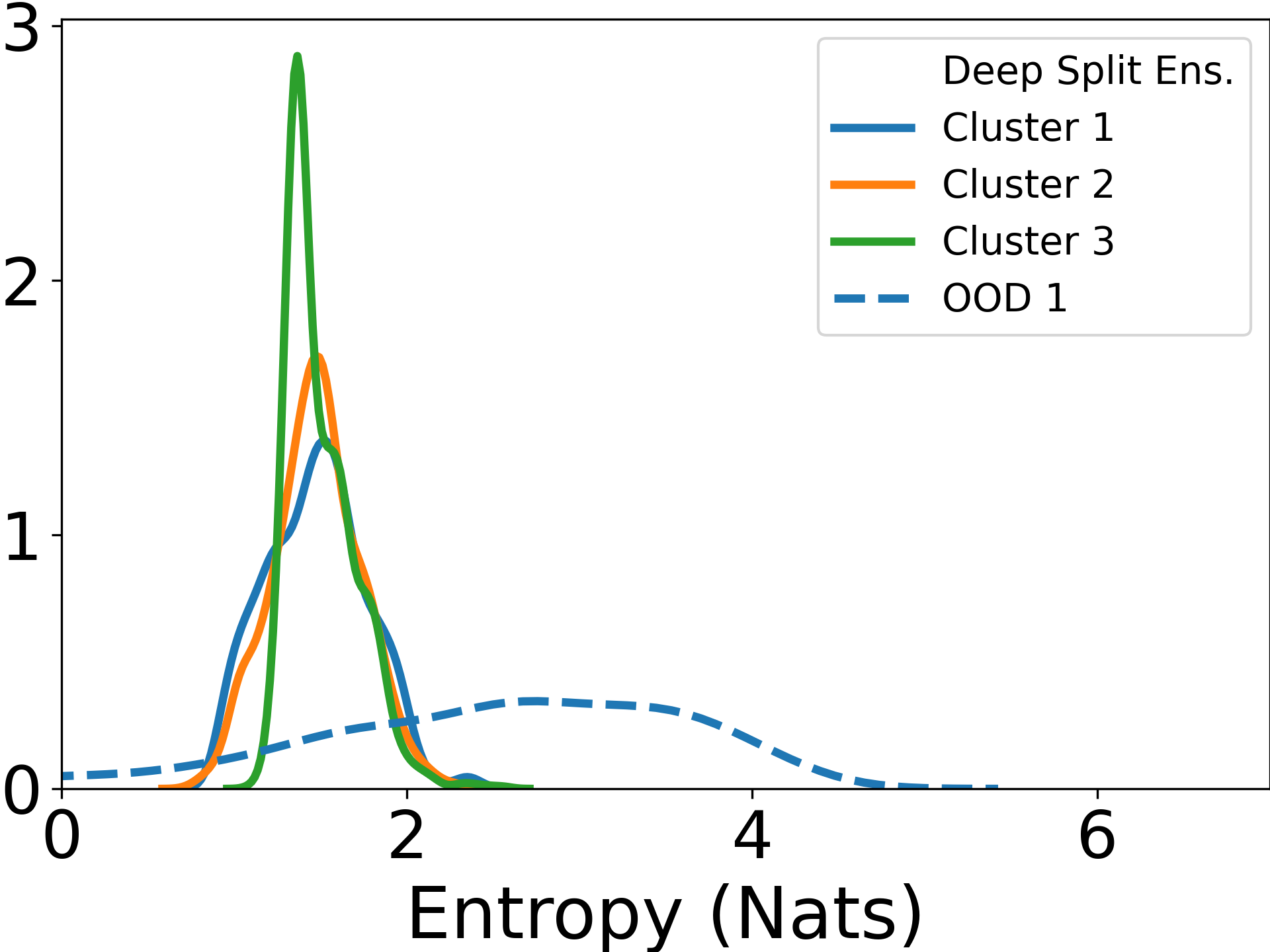}}
    ~
\subfloat{%
    \includegraphics[width=0.23\linewidth, valign=t]{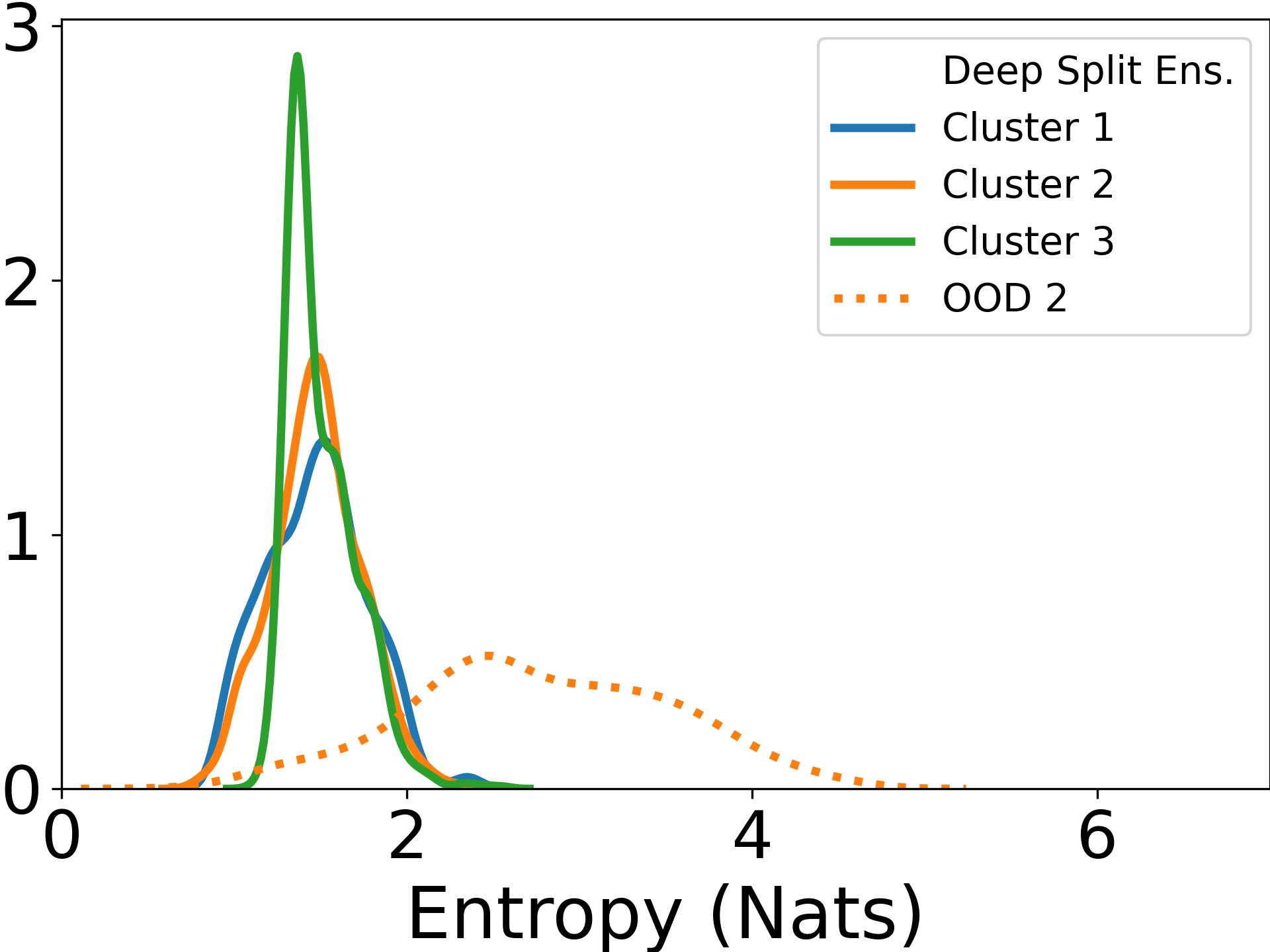}}
\caption{Entropy plots for `Boston' and `Concrete' datasets using hierarchical clustering (Section \ref{training}). The first two columns show the kernel density estimation (KDE) of entropy for in- distribution i.e. $\mathcal{N}(0, 1)$ and out-of-distribution samples, obtained with unified uncertainty estimation using deep ensemble and anchored ensembling respetively. The last two columns show `cluster-wise' KDE of entropy for in-distribution and out-of-distribution samples, obtained with disentangled uncertainty estimation using deep split ensembles. OOD 1 and OOD 2 refer to introducing dataset shift by inducing noise sampled from $\mathcal{N}(6, 2^2)$ into 2 random input features; the features correspond to different clusters for deep split ensembles. See Appendix \ref{appendix_entropy} for similar analyses for all other datasets.}
\label{entropy_plots}
\end{figure*}
\subsection{Calibration and uncertainty evaluation}\label{sec_calibration}
As a consequence of modelling disentangled predictive uncertainties during the training of NNs using NLL, we observe that our approach produces cluster-wise inherently well-calibrated models. Moreover, given that our model estimates disentangled uncertainties, we are able to assess the calibration of our model in a granular cluster-wise fashion. We assess our models without any post-hoc calibration. We first demonstrate this using entropy plots with out-of-distribution samples, and then using cluster-wise calibration curves using empirical rule.
\paragraph{Entropy analyses with out-of-distribution samples:} In real-world settings, there are often dataset shifts where the observed target data distribution may shift and eventually be very different once a model is deployed. Subsequently, the predictions need to exhibit higher uncertainty when this occurs. To assess it, we intentionally introduce a dataset shift by inducing noise, sampled from Gaussian distributions with shifted means and variances, into a random feature of a cluster and measure the corresponding clusters' predicted entropy (Figure \ref{entropy_plots}). We observe an increase in the entropy of only the noisy cluster while entropies of other clusters remain intact with deep split ensembles. However, unified uncertainty estimation methods like deep ensembles and anchored ensembling show an increase in the entropy corresponding to all features together. The disentanglement in OOD behaviour is an inherent characteristic of our method that cannot be observed in existing methods estimating unified uncertainties. Consequently, deep split ensembles can help better trace dataset shifts and pinpoint the noisy feature clusters during test time.
\begin{figure*}[ht!]
\centering
\subfloat{%
    \includegraphics[width=0.01\linewidth, height=2.5cm, valign=t]{figures/entropy_plots/boston_t.png}}\hspace{0.005mm}
    ~   
\subfloat{%
    \includegraphics[width=0.23\linewidth, valign=t]{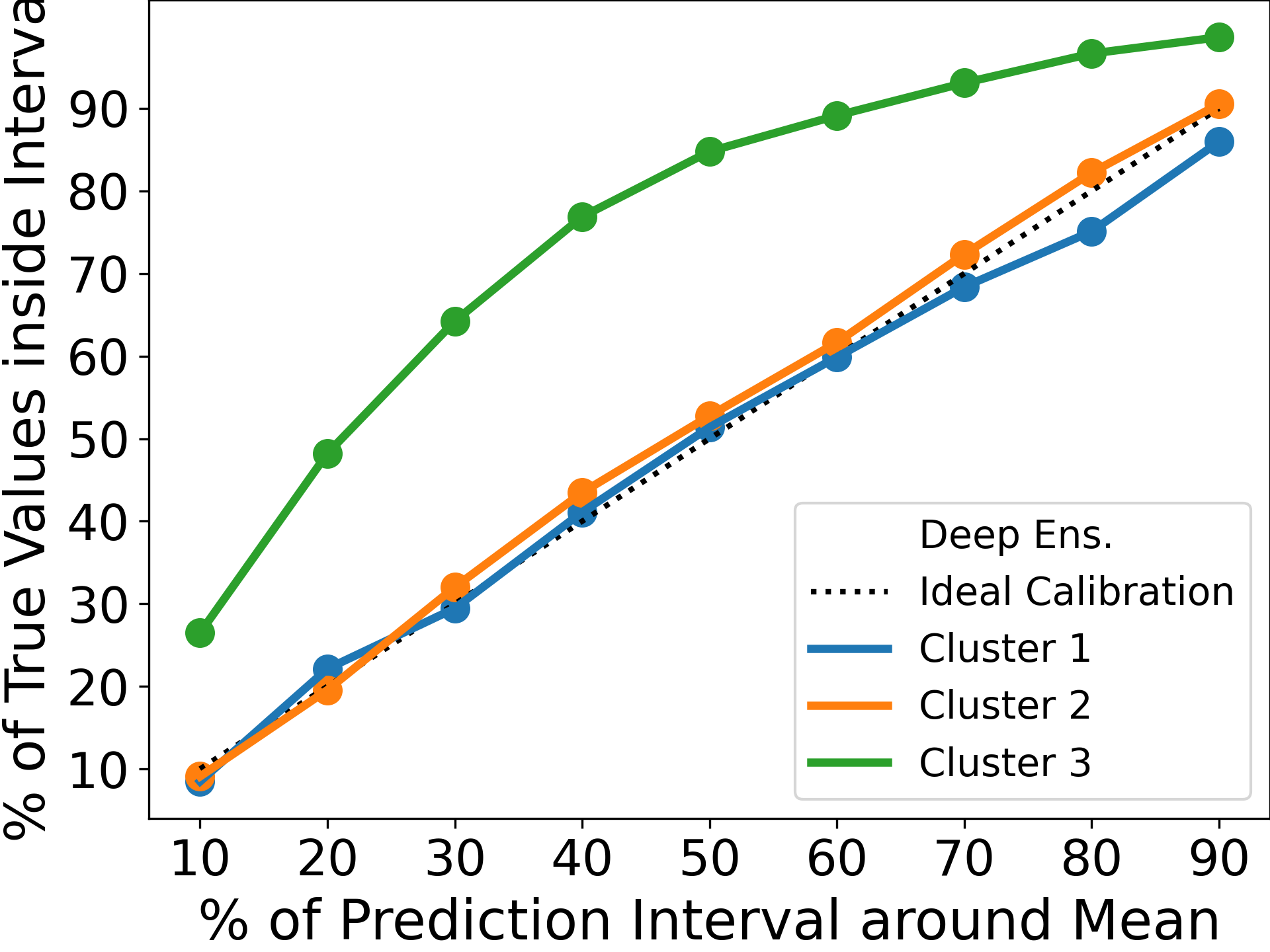}}
    ~
\subfloat{%
    \includegraphics[width=0.23\linewidth, valign=t]{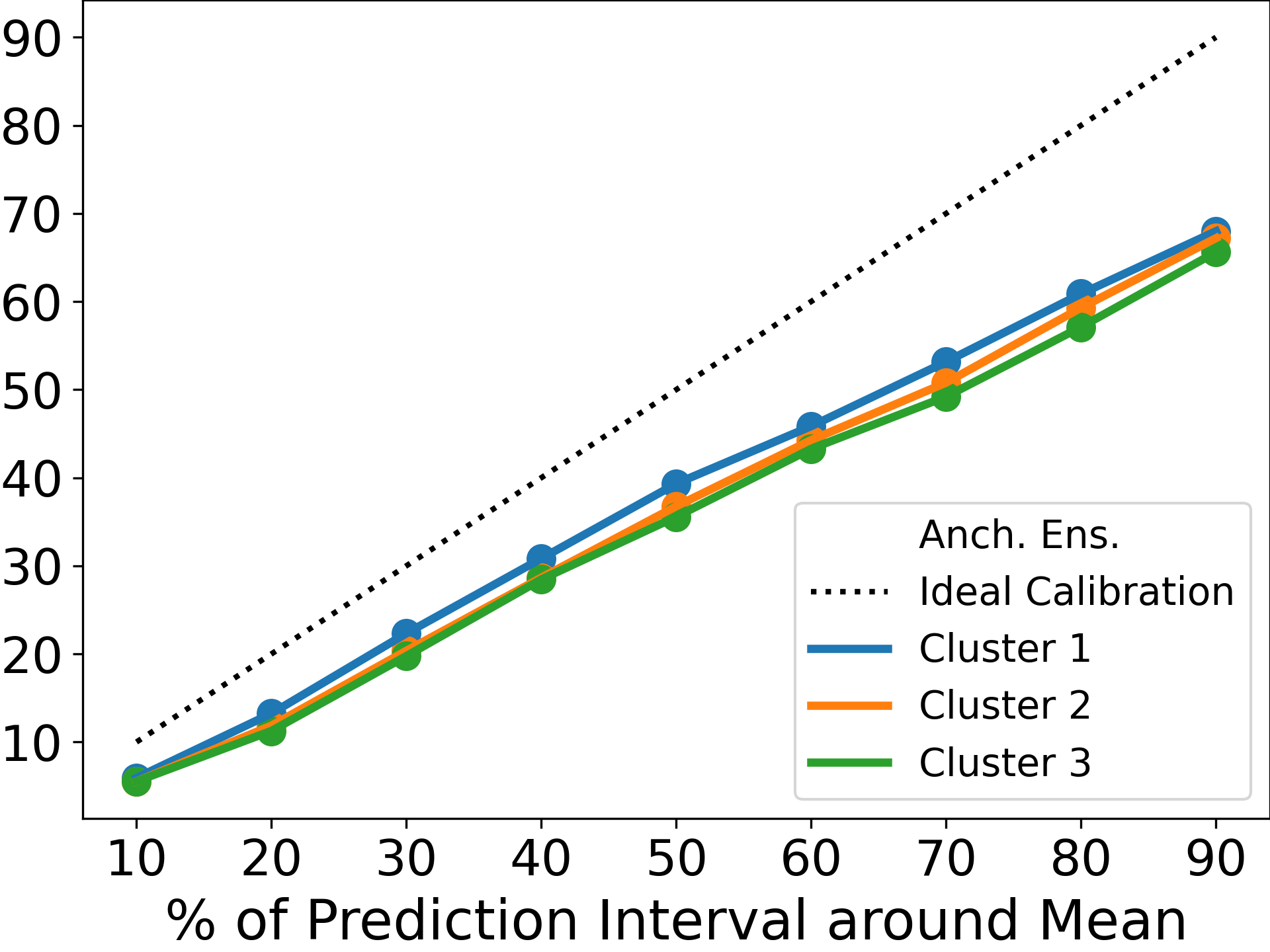}}
    ~
\subfloat{%
    \includegraphics[width=0.23\linewidth, valign=t]{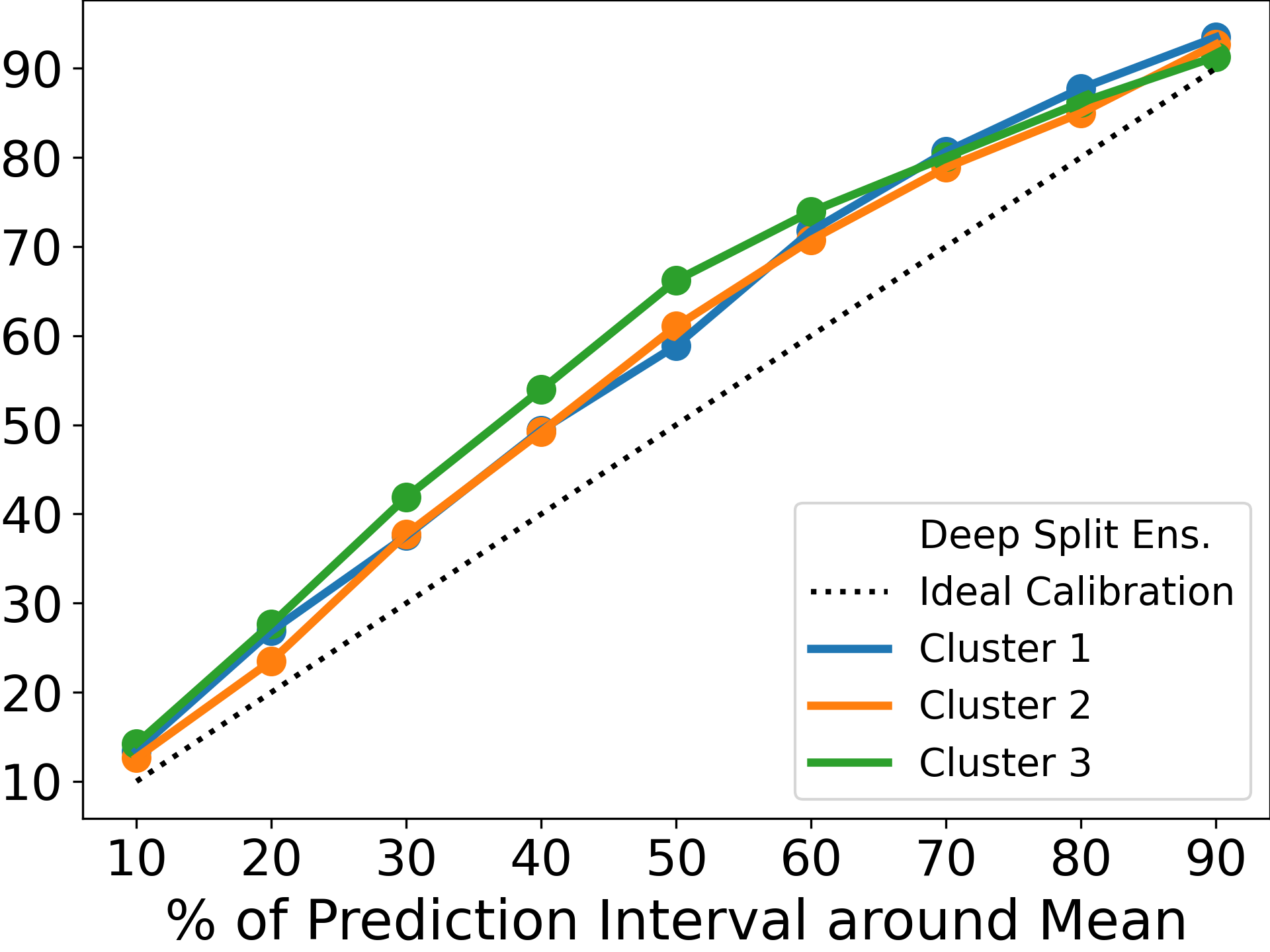}}

\subfloat{%
    \includegraphics[width=0.01\linewidth, height=2.7cm, valign=t]{figures/entropy_plots/concrete_t.png}}\hspace{0.005mm}
    ~   
\subfloat{%
    \includegraphics[width=0.23\linewidth, valign=t]{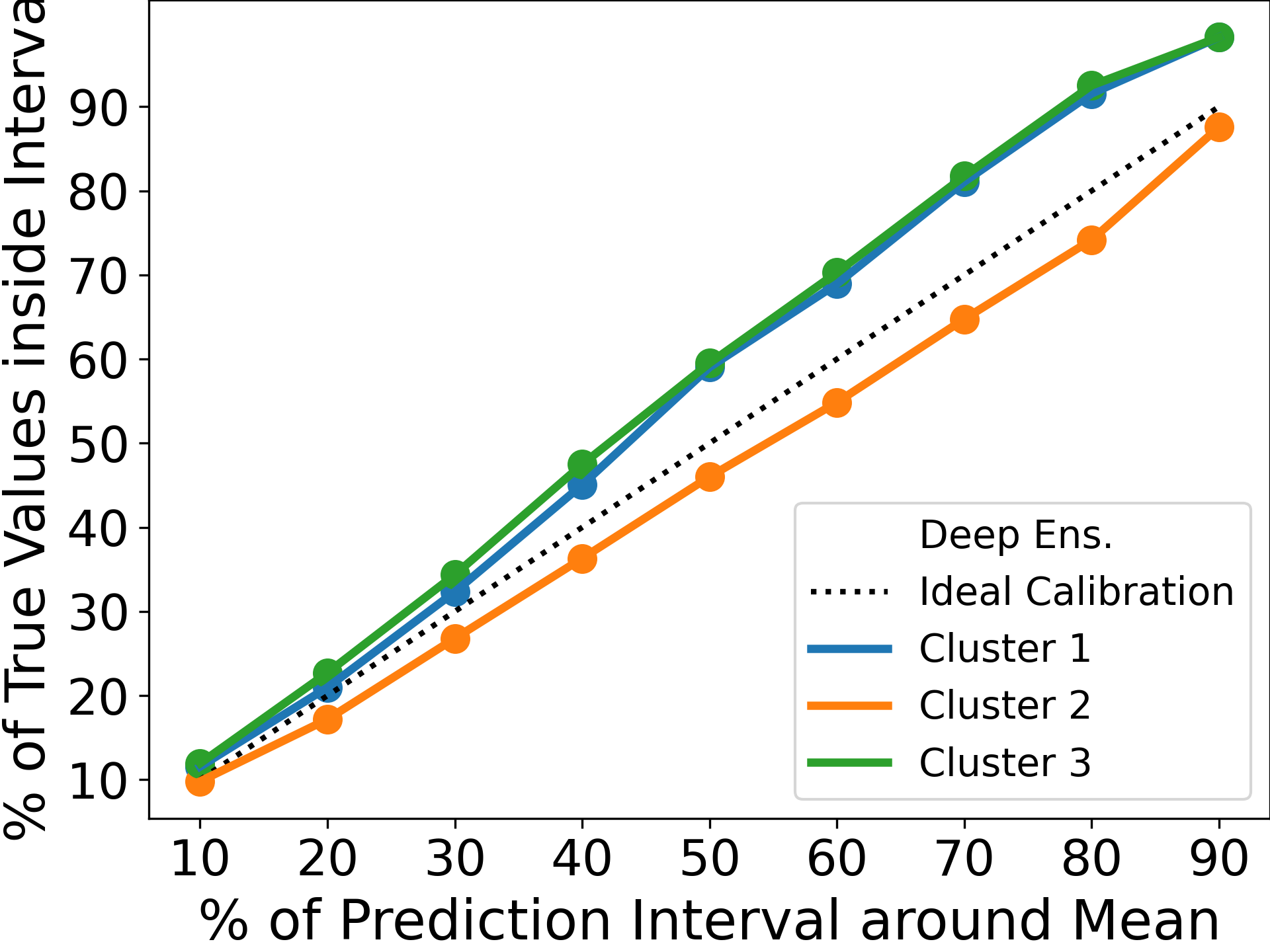}}
    ~
\subfloat{%
    \includegraphics[width=0.23\linewidth, valign=t]{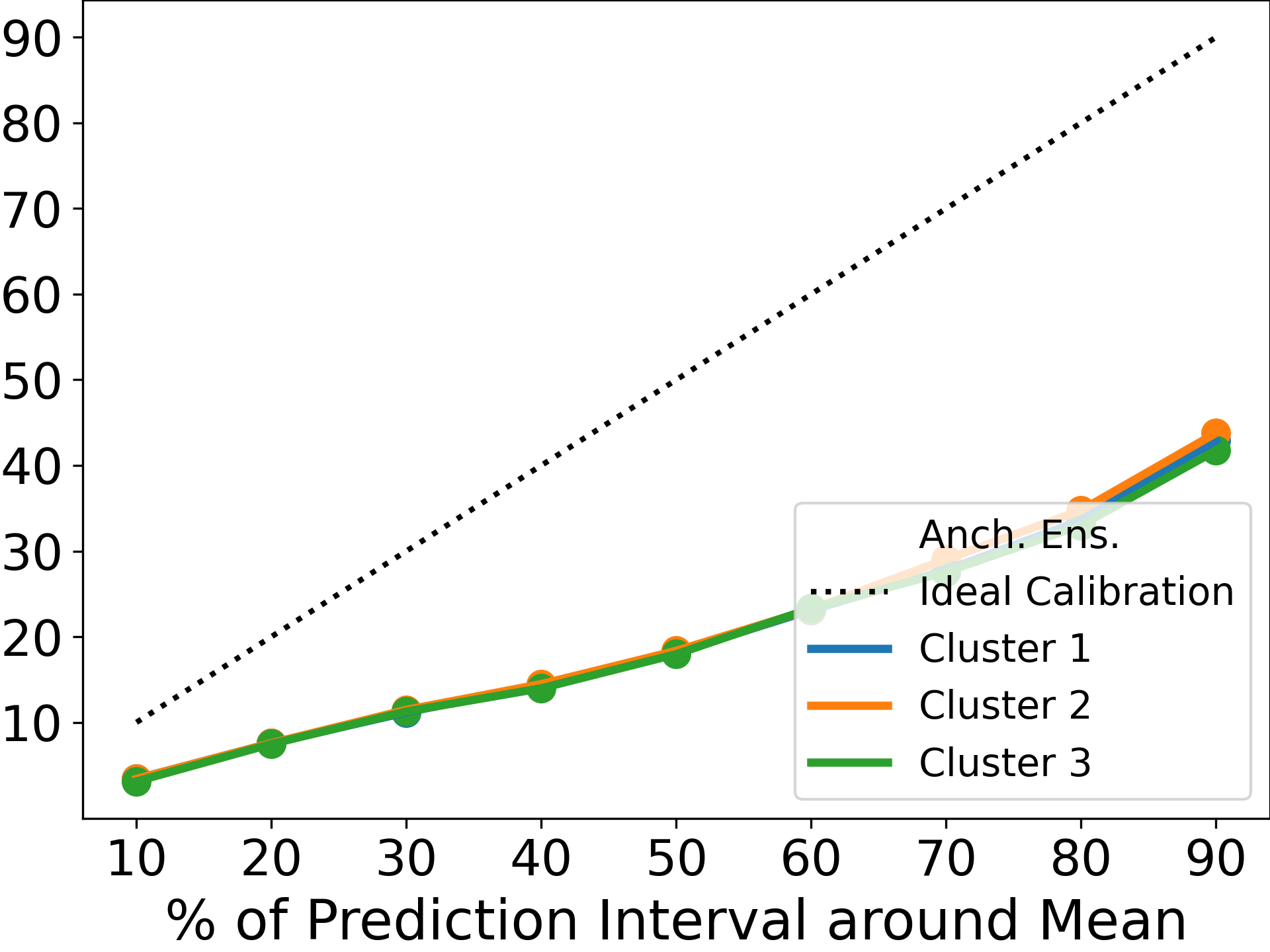}}
    ~
\subfloat{%
    \includegraphics[width=0.23\linewidth, valign=t]{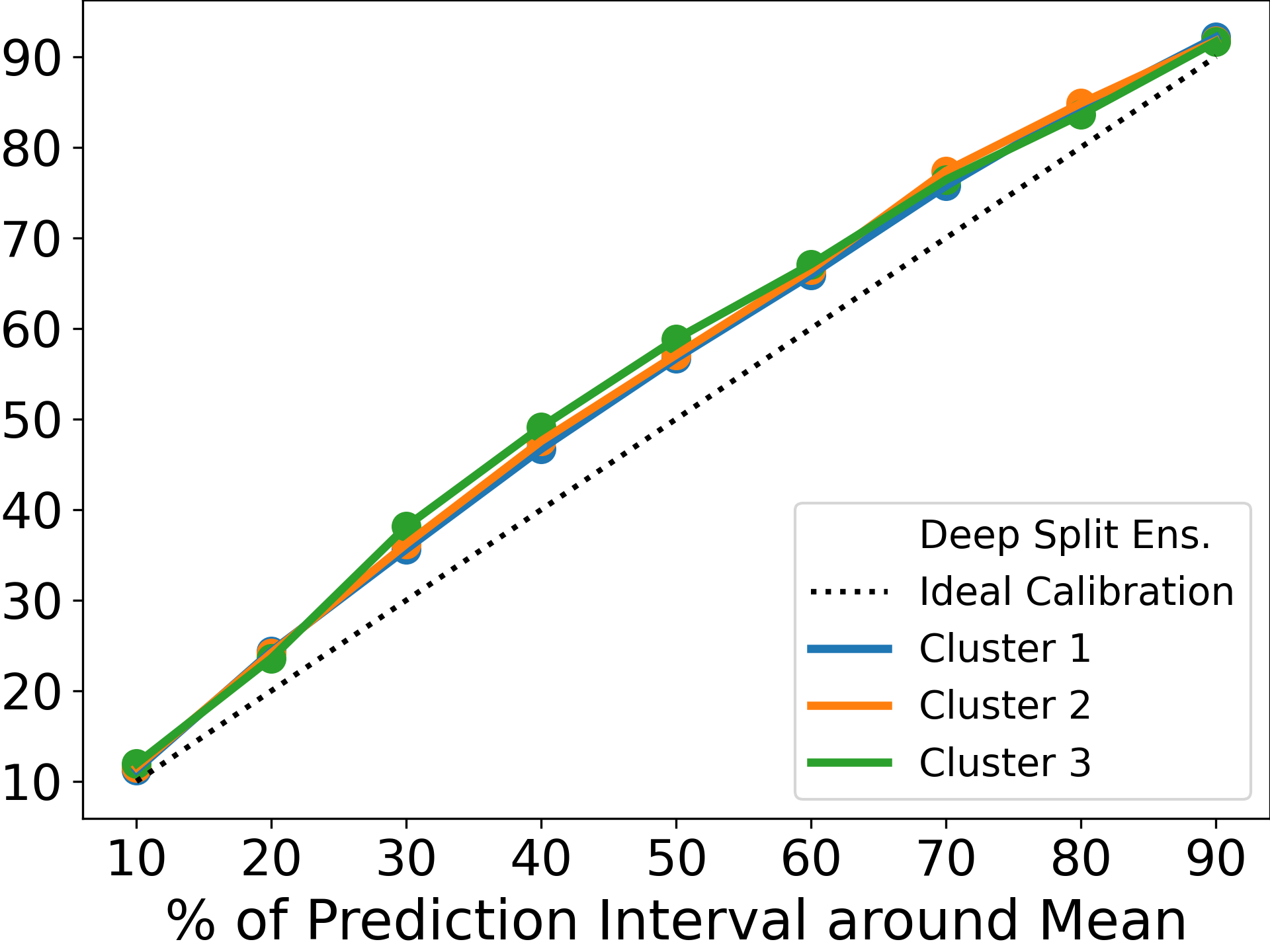}}
\caption{`Cluster-wise' calibration curves using empirical rule for `Boston' and `Concrete' datasets using hierarchical clustering (Section \ref{training}). The columns contain experiments using deep ensemble per input cluster (DEPC), anchored ensembling per input cluster (AEPC) and deep split ensembles respectively. See Appendix \ref{appendix_empirical} for similar analyses on all other datasets}
\label{empirical_plots}
\end{figure*}
\begin{table*}[htb!]
\centering
\begin{tabular}{lccccccc}
\toprule
Datasets & \multicolumn{3}{c}{RMSE} & Clusters ($C$) & \multicolumn{3}{c}{Cluster-wise NLL} \\
\cmidrule(r){2-4}
\cmidrule(r){6-8}
 & DEPC & AEPC & Deep Split Ens. & & DEPC & AEPC & Deep Split Ens. \\
 \midrule
\multirow{2}{*}{Power} 
 & \multirow{2}{*}{4.90 $\pm$ 0.23} & \multirow{2}{*}{4.91 $\pm$ 0.22} & \multirow{2}{*}{\textbf{4.07 $\pm$ 0.04}} & $C_{p1}$ & 2.99 $\pm$ 0.04 & 3.08 $\pm$ 0.07 & \textbf{2.81 $\pm $ 0.05} \\
 & & & & $C_{p2}$ & 3.13 $\pm$ 0.03 & 3.08 $\pm$ 0.07 & \textbf{2.83 $\pm$ 0.05}\\
 \midrule
\multirow{3}{*}{Wine} 
 & \multirow{3}{*}{0.64 $\pm$ 0.04} 
 & \multirow{3}{*}{0.66 $\pm$ 0.05} 
 & \multirow{3}{*}{\textbf{0.59 $\pm$ 0.02}} 
 & $C_{w1}$ & 0.94 $\pm$ 0.06 & 1.02 $\pm$ 0.09 & \textbf{0.88 $\pm $ 0.02} \\
 & & & & $C_{w3}$ & 0.96 $\pm$ 0.06 & 1.03 $\pm$ 0.09 & \textbf{0.89 $\pm $ 0.03} \\
 & & & & $C_{w3}$ & 0.94 $\pm$ 0.07 & 1.03 $\pm$ 0.09 & \textbf{0.90 $\pm $ 0.04} \\ 
 \bottomrule
\end{tabular}
  \caption{Results with clusters from human experts for the two datasets below. See Appendix \ref{appendix_human} for list of clusters ($C$).}
  \label{human_table}
\end{table*}
\footnotetext{Refer to Appendix \ref{appendix_nlls} for results on all other datasets.}
\paragraph{`Cluster-wise' calibration curves using empirical rule:} It is crucial to have good and stable calibration for reliable uncertainty estimates. To highlight our inherently well-calibrated models, we further evaluate to obtain calibration curves using the 68–95–99.7 rule (also called empirical rule). We first compute the $x\%$ prediction interval for each test datapoint based on Gaussian quantiles using the predicted mean and variance. We then calculate the fraction of test observations (true values) that fall within this prediction interval. For a well-calibrated model, the observed fraction should be close to $x\%$ calculated earlier. To see how our models perform in this setting, we sweep from $x = 10\%$ to $x = 90\%$ in steps of 10, and consequently a line lying very close to the line $y=x$ would indicate a well-calibrated model. Here, we further define stability of `cluster-wise' calibration as having similar calibration curves across clusters. As this experiment aims to test the calibration of the model with respect to each of the clusters individually, we use DEPC and AEPC to produce more suitable baselines to compare our method more rigorously. Figure \ref{empirical_plots} shows the calibration curves for each feature cluster for the different methods. We notice that deep split ensembles have a more uniform and stable calibration across clusters.

\subsection{Deep split ensembles based on domain knowledge and/or user needs}\label{human}
We illustrate how deep split ensembles allow for modelling predictive uncertainties using domain knowledge and/or user needs by taking in such clusters of input feature space. This is important as it brings the human in the loop and helps better define the task qualitatively. We consulted human experts, for the `Power' and `Wine' datasets, to qualitatively cluster the input features based on the uncertainties they would desire from a machine learning system trained on those datasets (details in Appendix \ref{appendix_human}). We then trained a deep split ensemble, DEPC and AEPC using the same experimental setup as above. Table \ref{human_table} shows that the results of deep split ensembles here are comparable to results in Table \ref{base_comp_table}, while outperforming DEPC and AEPC. The consistent improved performance upon changing the constituents of the clusters of input features demonstrates the inherent flexibility available while training deep split ensembles.

\begin{figure*}[ht!]
\subfloat{%
    \includegraphics[width=0.23\linewidth, valign=t]{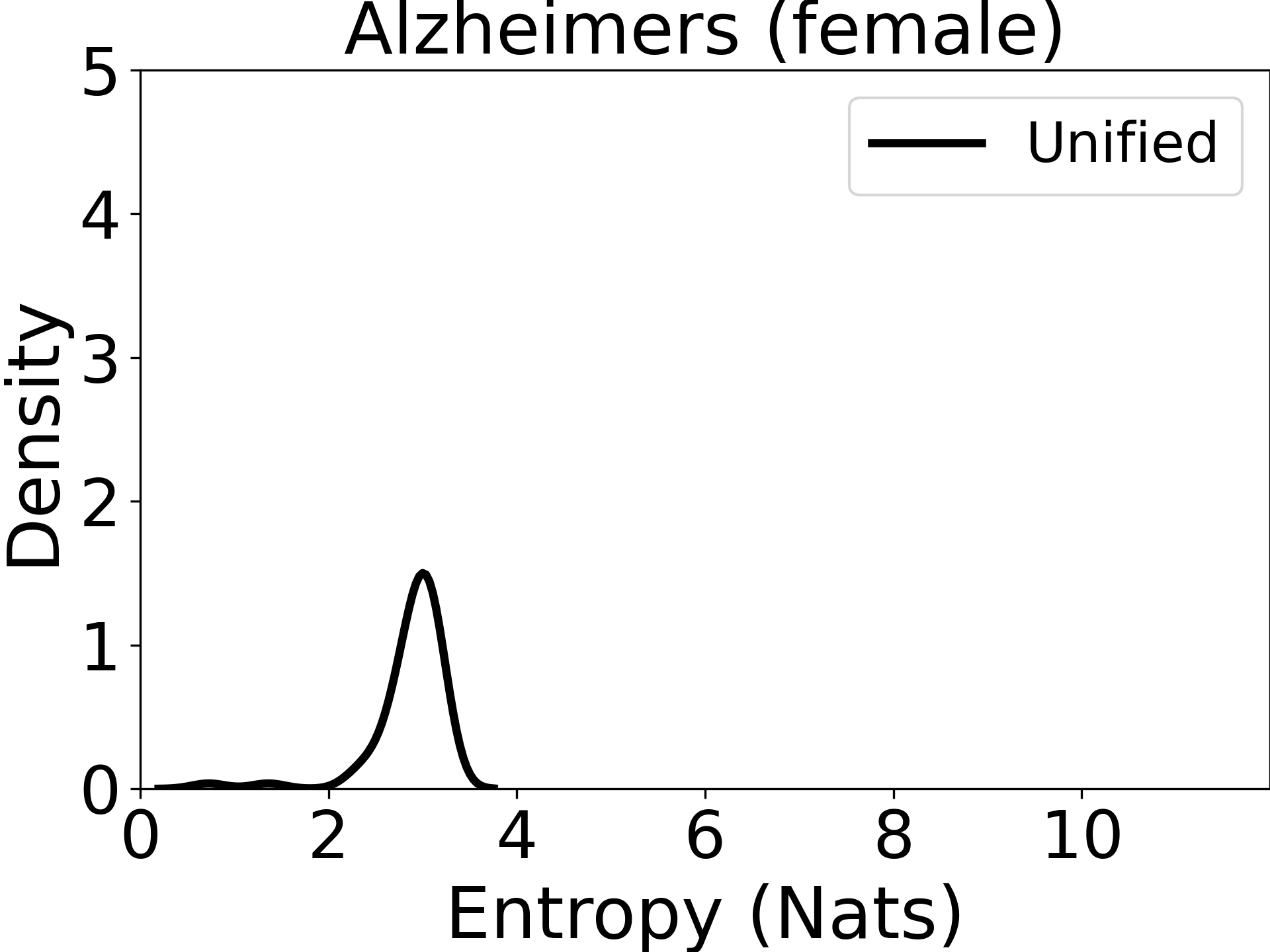}}
    ~~
\subfloat{%
    \includegraphics[width=0.23\linewidth, valign=t]{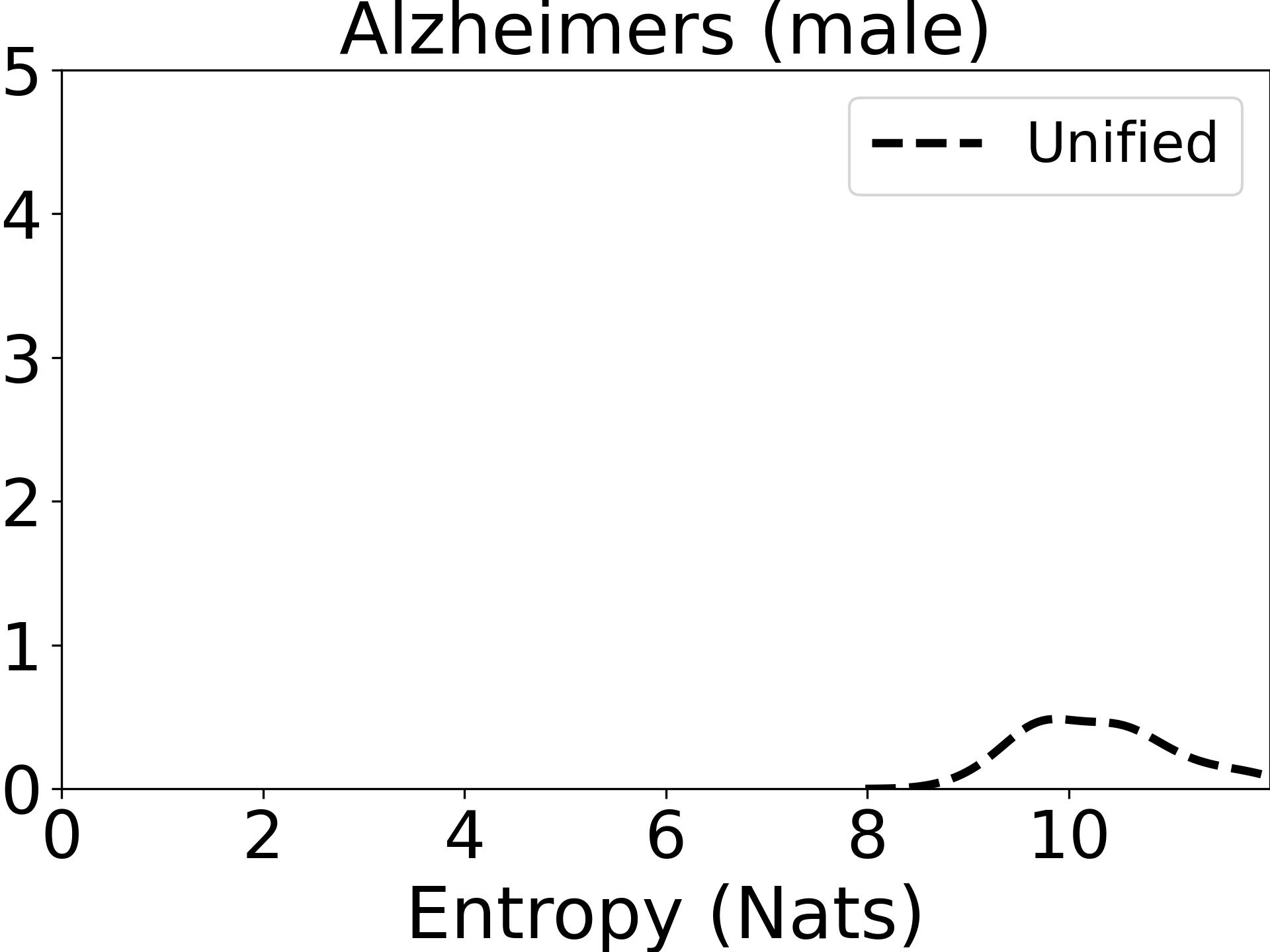}}
    ~~
\subfloat{%
    \includegraphics[width=0.23\linewidth, valign=t]{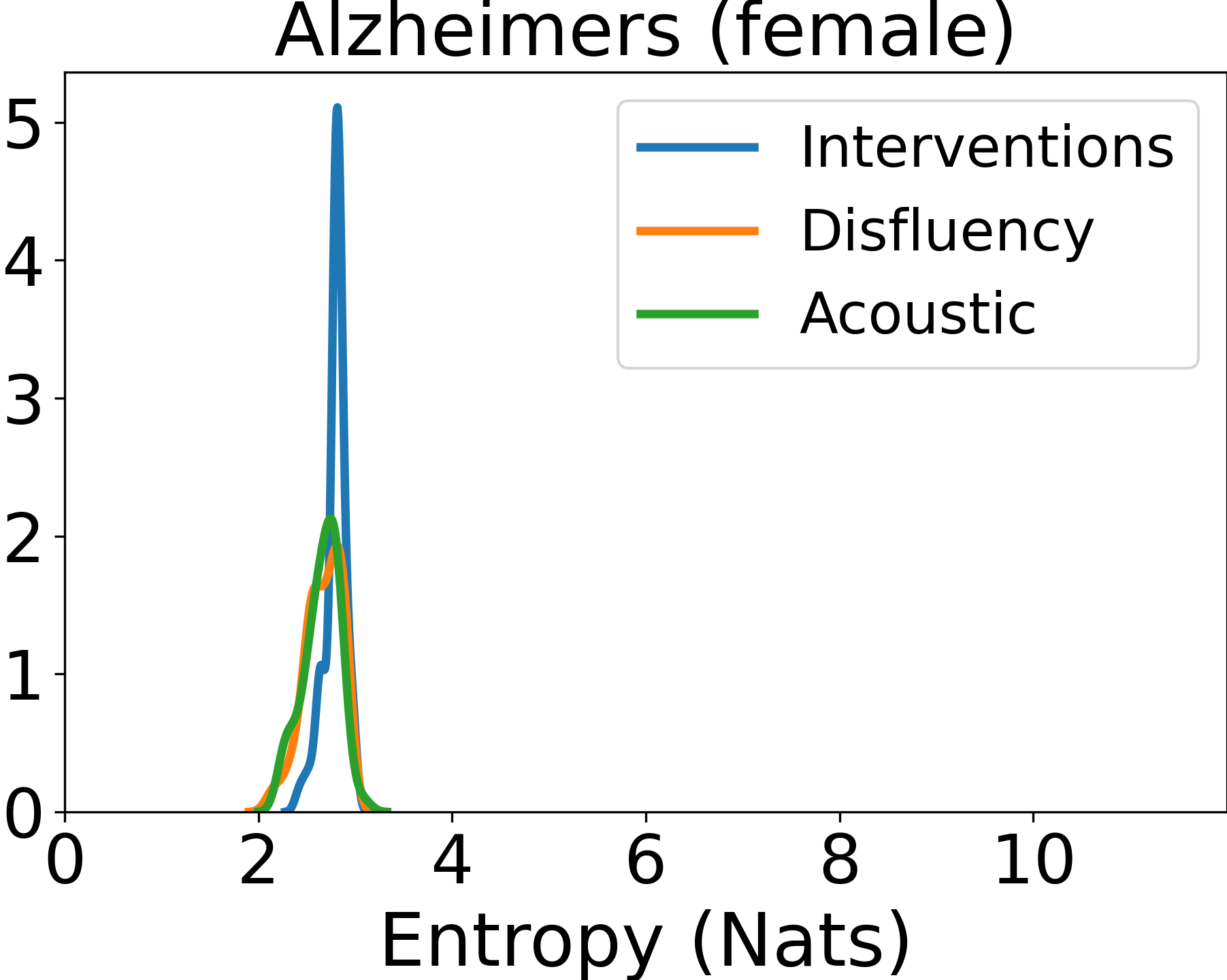}}
    ~~
\subfloat{%
    \includegraphics[width=0.23\linewidth, valign=t]{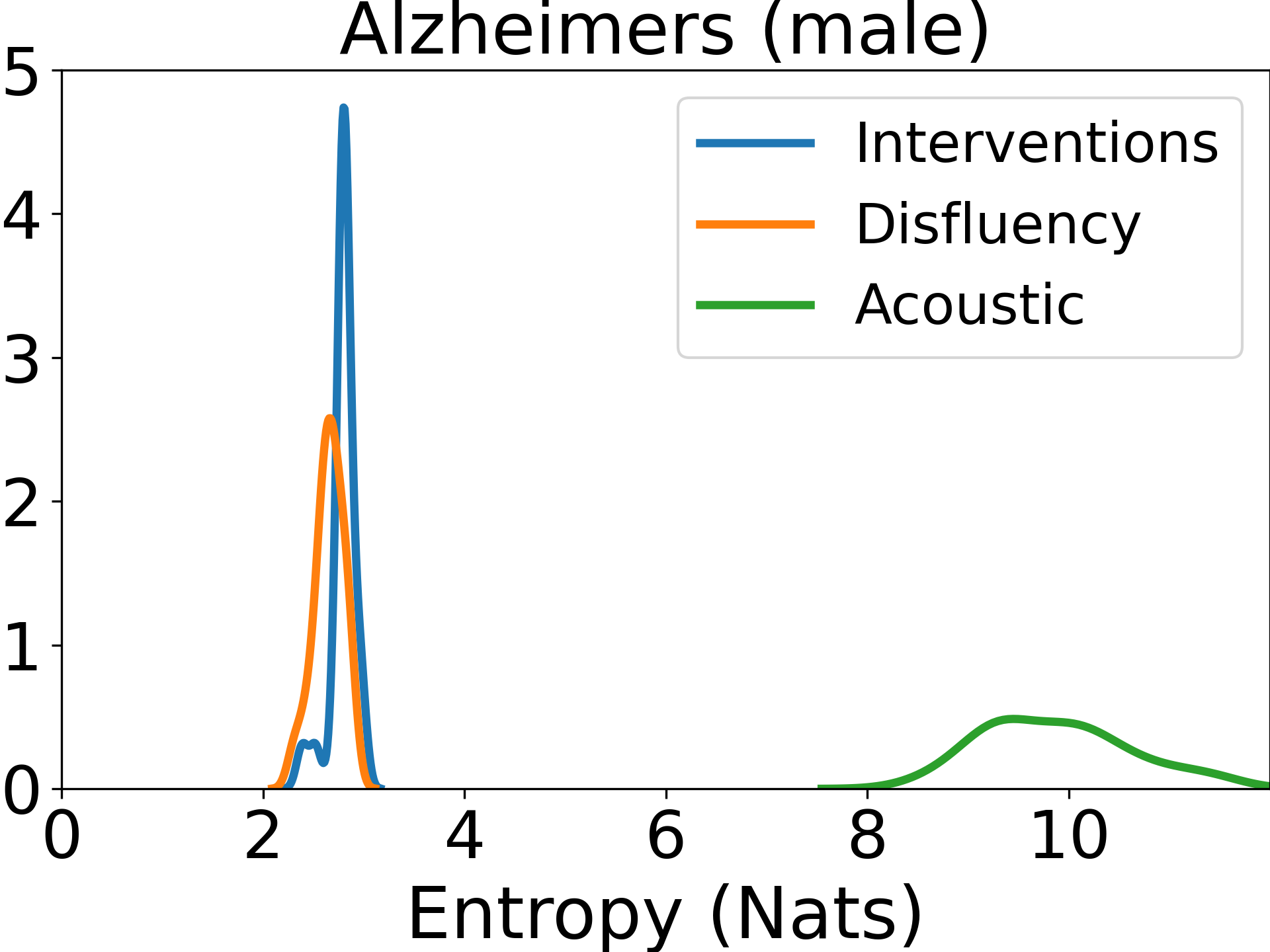}}
\caption{Entropy plots for Alzheimer's dataset showing the kernel density estimation (KDE) for male and female subjects specifically. The first two plots indicate unified uncertainty estimation using deep ensembles while the last two plots show disentangled uncertainty estimation using deep split ensembles. Both the methods are trained on female subjects and evaluated on male subjects to highlight hidden `modality-specific' gender bias.}
\label{male_female}
\end{figure*}
\subsection{Deep split ensembles in multi-modal settings}\label{multimodal}
The split nature of deep split ensembles makes them suitably applicable in multi-modal settings as heteroscedasticity in data can be highly decoupled due to the individual nature of the modalities. Each cluster of input features can be used to represent a particular modality of the input feature space to obtain a predictive uncertainty per modality.

To demonstrate this in a safety-critical application, we use a multi-modal Alzheimer's dementia (AD) dataset, `ADReSS', consisting of speech samples (audio) and their transcriptions (text), to regress MMSE\footnote{Mini-Mental State Examination (MMSE) scores, ranging from 0 to 30, offers a way to quantify cognitive function and screen for cognitive loss by testing the individuals’ attention, recall, language and motor skills \cite{tombaugh1992mini}.} scores. The standardized dataset contains 108 train and 48 heldout-test subjects. The train set is further split into 80\%-20\% train-validation sets. We first devise a feature engineering pipeline that extracts several multi-modal cognitive and acoustic feature sets - interventions, disfluency, and acoustic - based on domain knowledge and context.  We then train a deep split NN with those feature sets. The deep splits corresponding to disfluency and acoustic feature sets are fully connected and that corresponding to interventions feature set is an LSTM. See Appendix \ref{appendix_dementia} for details about the dataset, feature sets, model architecture and hyperparameters. Our method shows improved performance over state-of-the-art RMSE results, while also estimating a predictive uncertainty for each modality (Table \ref{AD_table}). We attribute this observation to the variance-affected learning rate \cite{nix1994estimating} which would help stabilize multi-modal training. We further train a deep ensemble for comparison on predictive uncertainty estimates, and observe that deep split ensemble achieves a better RMSE as well as a better NLL in each of the modalities.

To assess the calibration of our model and highlight potential hidden bias in real-world settings, we then train on only female subjects and evaluate it on only male subjects from the dataset. This would induce a bias in the model, which can be illustrated with the predicted uncertainties, as two modes (disfluency and interventions) are gender neutral by nature, whereas acoustic features can significantly vary. Consequently, upon experimentation (Figure \ref{male_female}), we observe that the entropy corresponding to all modalities together increases on the male inputs in case of unified uncertainty estimation using deep ensemble. However, in case of disentangled uncertainty estimation using deep split ensemble, only the entropy of the acoustic feature set on the male inputs significantly increases. The high predicted uncertainty corresponding to acoustic features for only male subjects highlights the hidden `modality-specific' gender bias.
\begin{table}[ht!]
\small
  \centering
  \begin{tabular}{lcc}
    \toprule
    Model & RMSE & NLL \\
    \midrule
    \citet{pappagariusing} & 5.37 & -- \\
    \citet{luz2020alzheimer}   & 5.20  &  -- \\
    \citet{sarawgi2020multimodal} & 4.60 & -- \\
    \citet{balagopalan2020bert} & 4.56 & -- \\
    \citeauthor{rohanianmulti} & 4.54 & -- \\
    Deep Ensemble & 4.90 & 3.08\\
    \textbf{Deep Split Ensemble}   & \textbf{4.37}   &  \textbf{2.94, 2.98, 2.94} \\
    \bottomrule
  \end{tabular}
    \caption{Test results on multi-modal ADReSS dataset for MMSE score regression.}
    \label{AD_table}
  
\end{table}

\section{Discussion and future work}
We have proposed a conceptually simple yet effective non-Bayesian method, \textit{deep split ensembles}, to estimate disentangled predictive uncertainties using NNs for input feature clusters. Disentangling a unified uncertainty allows for granular information about plausible sources of heteroscedasticity in the data. This is important in safety-critical settings as it enables improved risk assessment and decision-making. One can further form clusters containing one feature each to estimate feature-wise uncertainties. Using thus produced entropy values, noisy features or clusters can be suppressed while training a more reliable model for the same dataset with potentially improved performance. This encourages interoperability between humans and models in a unique way. Our method also reduces computational costs through sparser clusterwise connections, requires few changes in the NN, and can be readily implemented and trained. The nature of the split NN structure facilitates intuitive model-parallelism training for large models in multi-GPU systems where each cluster can be placed in separate GPUs. Using domain knowledge from human experts, deep split ensembles can help satisfy user needs by generating different combinations of uncertainty estimates desired from a machine learning system, thus providing a more controllable form of reliability and awareness with the model. The potential to highlight hidden biases, such as shown in multi-modal settings, has immediate and apparent real world applications to mitigate unseen biases in deployed models. This serves as a motivation for fair and aware systems supporting human-assisted AI. 

A direct extension of our work would be to use complex distributions such as mixture density networks \cite{bishop1994mixture} for modelling output distributions. There are many exciting future directions, such as unsupervised learning to form deeper representation for the clusters of features \cite{xie2016unsupervised}, using uncertainty attention \cite{heo2018uncertainty, lee2018dropmax} to aid in training of the ensemble classifiers, exploring adaptive defer systems \cite{madras2018predict} along with partial deferring based on clusters for better calibration of uncertainties coupled with human experts, and considering uncertainty of human feedback \cite{he2020learning}.

\newpage
\bibliography{aaai21}
\section*{Broader and ethical impact}
Uncertainty estimations are crucial when employing black box neural networks (NNs) in sensitive and critical applications such as healthcare and self-driving cars. These NNs often tend to be over confident of their predictions; the confidence measures are not a true estimate of the model's performance. Predicting the uncertainties can help in better understanding how confident the model is in its predictions and reflect upon the noise introduced by the stochastic data generation processes. It is important to know what the model is unsure about. Although current NNs can demonstrate high performance on test datasets, they sometimes tend to fail when deployed in real world settings due to noisy real-world data and dataset shifts.

A unified uncertainty estimation helps in providing confidence estimates. Further disentangling the unified predictive uncertainties give deeper insights into the various feature clusters and their associated heteroscedasticity. We believe that anyone who is involved in synergy with a machine learning system in decision making will benefit from such systems. For example, when such a system is deployed in a hospital setting to stratify the risk of a disease/condition, the doctor can understand the uncertainties associated with each input feature modality, and is able to better interpret the model's belief. In such high-risk and safety-critical settings, deploying a black box NN could be sub-optimal. Additionally, multidisciplinary machine learning researchers will benefit from this as they will have a tool to better incorporate domain knowledge and user demands/needs.


\newpage
\clearpage
\setcounter{page}{1}
\appendix
\begin{center}
    \huge \textbf{Appendix}
\end{center}
    

\section{Feature Clusters}\label{appendix_clusters}
\subsection{Feature clusters from hierarchical clustering}\label{appendix_hc}
\begin{figure*}[ht!]
\centering
\subfloat[Boston Housing]{%
    \includegraphics[width=0.31\linewidth, valign=t]{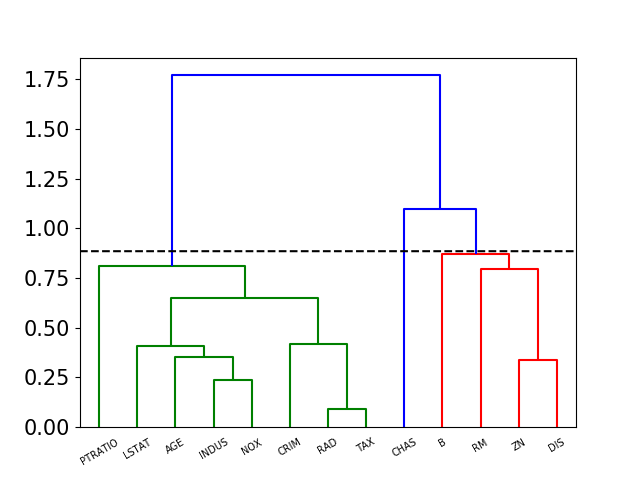}}
    ~~~
\subfloat[Concrete]{%
    \includegraphics[width=0.31\linewidth, valign=t]{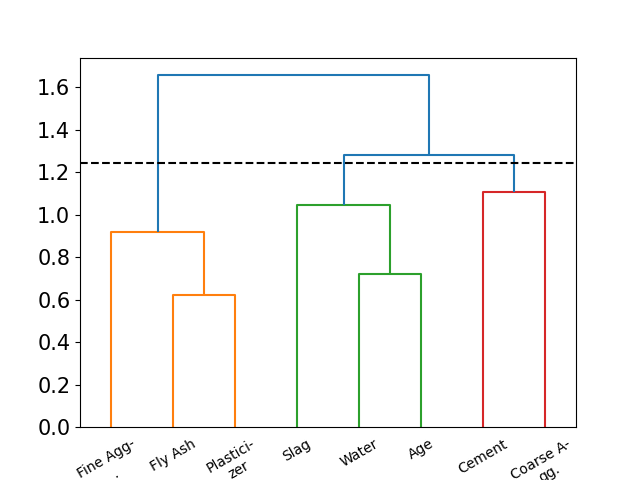}}
    ~~~
\subfloat[Energy Efficiency]{%
    \includegraphics[width=0.31\linewidth, valign=t]{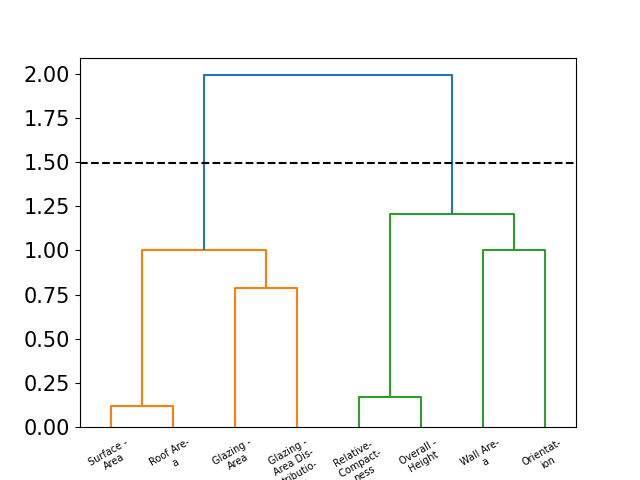}}
    \\
\subfloat[Kin8nm]{%
    \includegraphics[width=0.31\linewidth, valign=t]{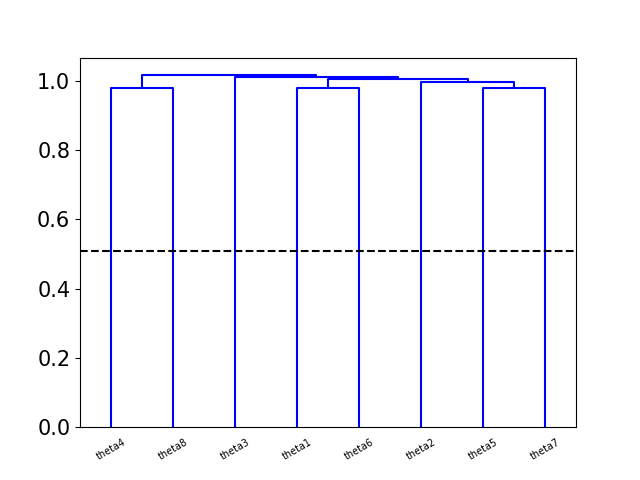}}
    ~~~
\subfloat[Naval propulsion plant]{%
    \includegraphics[width=0.31\linewidth, valign=t]{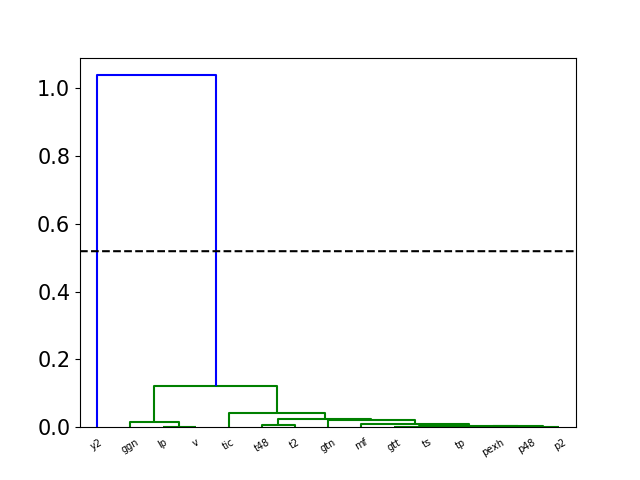}}
    ~~~
\subfloat[Power Plant Output]{%
    \includegraphics[width=0.31\linewidth, valign=t]{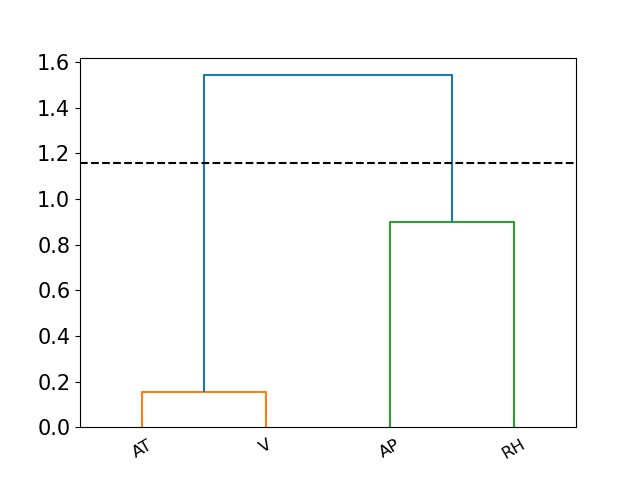}}
    \\
\subfloat[Protein Structure]{%
    \includegraphics[width=0.31\linewidth, valign=t]{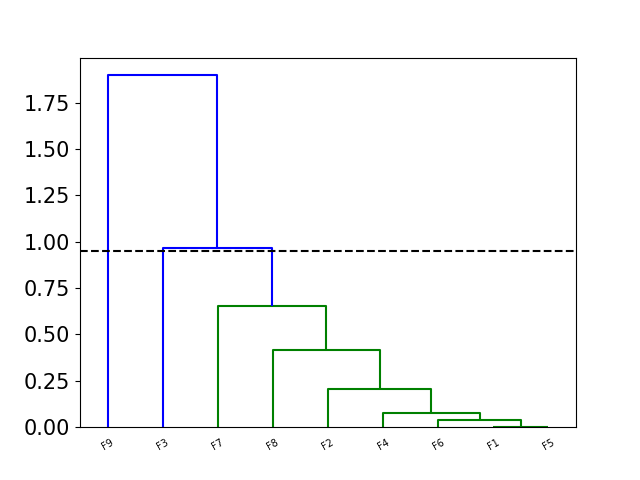}}
    ~~~
\subfloat[Red Wine Quality]{%
    \includegraphics[width=0.31\linewidth, valign=t]{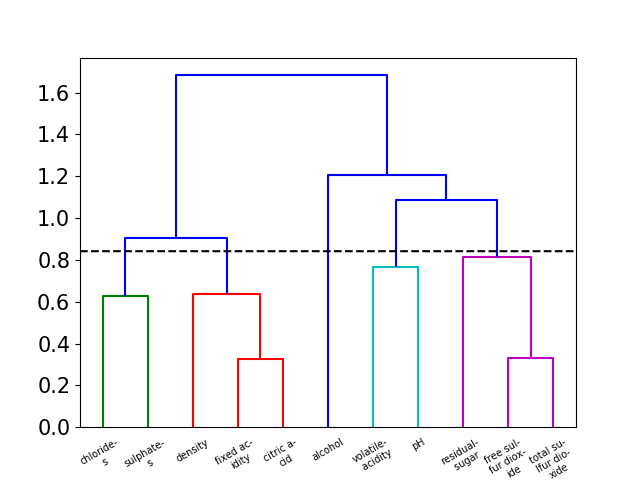}}
    ~~~
\subfloat[Yacht Hydrodynamics]{%
    \includegraphics[width=0.31\linewidth, valign=t]{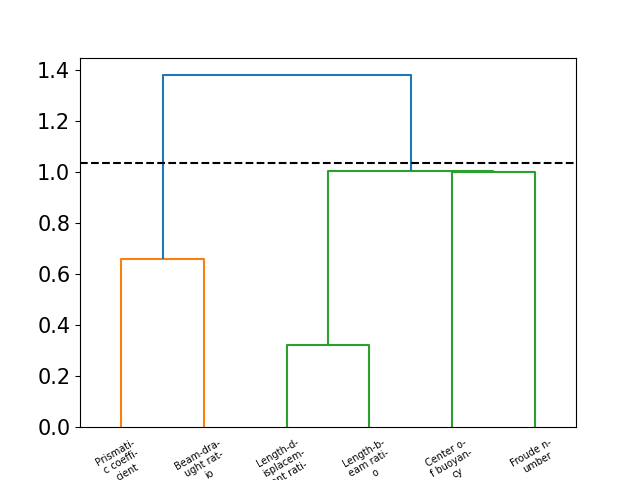}}
\caption{Dendograms obtained for UCI datasets upon hierarchical clustering of features based on Pearson correlation distance. The dotted line shows the threshold value used to extract clusters.}
\label{appendix_dendograms}
\end{figure*}

As discussed in Section \ref{training}, the input feature space is split into $k$ exhaustive clusters using hierarchical clustering based on Pearson correlation distance.  The dendograms thus obtained upon hierarchical clustering with complete linkage are thresholded relative to the maximum distance to obtain feature clusters (Figure \ref{appendix_dendograms}); we chose 0.5 and 0.75 to span a variety of number of clusters and features per cluster. One can change this threshold value to obtain different sets of feature clusters. Table \ref{appendix_cluster_table} enlists the clusters thus obtained for each dataset.

\subsection{Feature clusters from human experts}\label{appendix_human}
As discussed in Section \ref{human}, we consulted human experts, for the `Power Plant Output' and `Red Wine Quality' datasets, to qualitatively cluster the input features based on the uncertainties desired from a machine learning system trained on those datasets. Table \ref{human_appendix_table} shows the features cluster thus obtained, and the reasons as mentioned by the human experts are summarized below:
\begin{itemize}
    \item `Power Plant Output': While the Vacuum is collected from and has effect on the Steam Turbine, the three other ambient variables effect the GT performance.
    \item `Red Wine Quality': Alcohol, pH, fixed acidity, density, and residual sugar are resultant characteristics of the wine. Volatile acidity and citric acid are added acidity in the wine-making. Chlorides, free sulphur dioxide, total sulphur dioxide, and sulphates are preservatives and antibacterials.
\end{itemize}

\begin{table*}[ht!]
\centering
  \setlength\tabcolsep{4.7pt}
  \begin{tabular}{lcl}
    \toprule
        Dataset     & Cluster     & Features \\
    \toprule
\multirow{3}{*}{Boston Housing}   
& 1 & CRIM, INDUS, NOX, AGE, RAD, TAX, PTRATIO, LSTAT \\
& 2 & ZN, RM, DIS, B \\
& 3 & CHAS \\
 \midrule
\multirow{3}{*}{Concrete}   
& 1 & Fly Ash, Superplasticizer, Fine Aggregate \\
& 2 & Water, Age, Blast Furnace Slag \\
& 3 & Cement, Coarse Aggregate \\
 \midrule
\multirow{2}{*}{Energy Efficiency }   
& 1 & Surface Area, Roof Area, Glazing Area, Glazing Area Distribution \\
& 2 & Relative Compactness, Overall Height, Wall Area, Orientation \\
\midrule
\multirow{8}{*}{Kin8nm} 
& 1 & theta1 \\
& 2 & theta2 \\
& 3 & theta3 \\
& 4 & theta4 \\
& 5 & theta5 \\
& 6 & theta6 \\
& 7 & theta7 \\
& 8 & theta8 \\
\midrule
\multirow{2}{*}{Naval Propulsion Plant}   
& 1 & lp, v, ggn \\
& 2 & gtt, gtn, ts, tp, t48, t2, p48, p2, pexh, tic, mf \\
\midrule
\multirow{2}{*}{Power Plant Output}   
& 1 & AT, V \\
& 2 & AP, RH \\
\midrule
\multirow{3}{*}{Protein Structure}   
& 1 & F1, F2, F4, F5, F6, F7, F8 \\
& 2 & F3 \\
& 3 & F9 \\
\midrule
\multirow{5}{*}{Red Wine Quality }   
& 1 & chlorides, sulphates \\
& 2 & fixed acidity, citric acid, density \\
& 3 & volatile acidity, pH \\
& 4 & residual sugar, free sulfur dioxide, total sulfur dioxide \\
& 5 & alcohol \\
\midrule
\multirow{2}{*}{Yacht Hydrodynamics}   
& 1 & Prismatic coefficient, Beam-draught ratio \\
& 2 & Length-displacement ratio,  Length-beam ratio, Longitudinal position, Froude number \\
    \bottomrule
  \end{tabular}
    \caption{List of feature clusters obtained for UCI datasets using hierarchical clustering.}
  \label{appendix_cluster_table}
  
\end{table*}

\begin{table*}[ht!]
\centering
  \begin{tabular}{lcl}
    \toprule
        Dataset     & Cluster     & Features \\
    \toprule
\multirow{2}{*}{Power Plant Output}   
& $C_{p1}$ & AT, AP, RH \\
& $C_{p2}$ & V \\
\midrule
\multirow{3}{*}{Red Wine Quality }   
& $C_{w1}$ & alcohol, pH, fixed acidity, density, residual sugar \\
& $C_{w2}$ & volatile acidity, citric acid \\
& $C_{w3}$ & chlorides, free sulphur dioxide, total sulphur dioxide, sulphates \\
    \bottomrule
  \end{tabular}
    \caption{List of feature clusters obtained for `Power' and `Wine' UCI datasets from human experts.}
  \label{human_appendix_table}
\end{table*}

\section{Derivations and proofs}\label{appendix_proofa}
\subsection{Derivations of $\mu$ and $\sigma$ of a Gaussian mixture}\label{appendix_musigma}
Given a Gaussian mixture $p_E(\mathbf{y}|\mathbf{x})$, where \\ $p_E(\mathbf{y}|\mathbf{x}) = E^{-1}\sum_e \mathcal{N}\left({\boldsymbol\mu_\theta}_e(\mathbf{x}), {\Sigma_\theta}_e(\mathbf{x})\right)$, let the mean and the variance of the mixture be $\boldsymbol\mu_E(\mathbf{x})$ and $\Sigma_E(\mathbf{x})$ respectively. Let $t \sim E^{-1}\sum_e \mathcal{N}\left({\boldsymbol\mu_\theta}_e(\mathbf{x}), {\Sigma_\theta}_e(\mathbf{x})\right) $.
\subsection*{Derivation of the mean of a Gaussian mixture}
$t = z + \epsilon$, where $z={\boldsymbol\mu_\theta}_e(\mathbf{x})$ with equal probability $E^{-1}$ for $e=1, 2, \dots, E$ and the conditional probability distribution of $\epsilon$ given $z$ will be $N(0,  {\Sigma_\theta}_e(\mathbf{x}))$.  

\[
    \boldsymbol\mu_E(\mathbf{x}) = E[t]
\]

\[
\implies \boldsymbol\mu_E(\mathbf{x}) = E[E[t|z]]
\]

\[
\implies\boldsymbol\mu_E(\mathbf{x}) = E\left.\begin{cases} \vdots \\ {\boldsymbol\mu_\theta}_e( \mathbf{x}) & \text{with probability }E^{-1} \\  \vdots \end{cases}\right\} 
\]

\begin{equation}
\label{mu_mix_proof}
\implies\boldsymbol\mu_E(\mathbf{x}) = E^{-1} \sum_{e=1}^E{\boldsymbol\mu_\theta}_e(\mathbf{x})
\end{equation}

\subsection*{Derivation of the variance of a Gaussian mixture}










We have,
\[
\text{var}(t) = E[\text{var}(t|z)] + \text{var}(E[t|z])
\]

\begin{multline*}
\implies \text{var}(t) = 
E \left.\begin{cases} \vdots \\ {\Sigma_\theta}_e( \mathbf{x}) & \text{with probability }E^{-1} \\  \vdots \end{cases}\right\} \\+
\text{var} \left.\begin{cases} \vdots \\ {\boldsymbol\mu_\theta}_e( \mathbf{x}) & \text{with probability }E^{-1} \\  \vdots \end{cases}\right\}
\end{multline*}

\begin{multline*}
\implies \text{var}(t) = 
E^{-1} \sum_{e=1}^E{\Sigma_\theta}_e(\mathbf{x}) \\+
E^{-1} \sum_{e=1}^E({\boldsymbol\mu_\theta}_e(\mathbf{x}) - \boldsymbol\mu_E(\mathbf{x}))({\boldsymbol\mu_\theta}_e(\mathbf{x}) - \boldsymbol\mu_E(\mathbf{x}))^T
\end{multline*}


Using Equation (\ref{mu_mix_proof}) and given our assumption that the outputs of deep splits (in a deep split NN) are linearly uncorrelated, we have,





\begin{equation*}
\Sigma_E(\mathbf{x}) =\text{diag}(\sigma^2_E(\mathbf{c}^1),\dots,\sigma^2_E(\mathbf{c}^k))
\end{equation*}

\begin{equation*}
\sigma_E^2(\mathbf{c}^i) = 
E^{-1} \sum_{e=1}^E({\sigma_\theta}_e^2(\mathbf{c}^i) +
{\boldsymbol\mu_\theta}_e^2(\mathbf{c}^i)) - \boldsymbol\mu_E^2(\mathbf{c}^i)
\end{equation*}
\section{Details of datasets, model and hyperparameters}\label{appendix_dataset}

\begin{table}[t]
  \small
  \centering
  \begin{tabular}{lrr}
    \toprule
    Dataset     & No. of datapoints & No. of features\\
    \midrule
 Boston Housing & 506 & 13 \\
 Concrete & 1,030 & 8 \\
 Energy Efficiency & 768 & 8 \\
 Kin8nm & 8,192 & 8 \\
 Naval propulsion plant & 11,934 & 16 \\
 Power Plant Output & 9,568 & 4 \\
 Protein Structure & 45,730 & 9 \\
 Red Wine Quality & 1,599 & 11 \\
 Yacht Hydrodynamics & 308 & 6 \\

    \bottomrule
  \end{tabular}
  \caption{UCI Benchmark regression dataset details}
  \label{datasetdetails_title}
\end{table}

\subsection{Benchmark regression datasets}\label{appendix_regdata}

Table \ref{datasetdetails_title} shows some statistics of the 9 benchmark regression datasets used in our experiments (Section \ref{exp_and_results}). We have included all the datasets in the Supplementary Material provided. The hyperparameters used for training the deep split ensembles are enlisted in Table \ref{dse_hyperparameters}. For anchored ensembling per feature cluster (AEPC) and deep ensembles per feature cluster (DEPC), we use the hyperparameters mentioned in the anchored ensembling and deep ensembles papers respectively.

  


\begin{table}[ht!]
\small
  \centering
  \begin{tabular}{lrrr}
    \toprule
    \multirow{2}{*}{Dataset} & Learning & \multirow{2}{*}{Epochs} & Batch\\  
     & Rate & & Size\\  
    \midrule
Boston Housing & 0.1 & 1000 & 100 \\
Concrete & 0.01 & 1500 & 32 \\
Energy Efficiency & 0.01 & 1500 & 16 \\
Kin8nm & 0.1 & 1000 & 100 \\
Naval Propulsion Plant & 0.01 & 1500 & 32 \\
Power Plant Output & 0.01 & 2500 & 256 \\
Protein Structure & 0.01 & 4000 & 1024 \\
Red Wine Quality & 0.1 & 1000 & 100 \\
Yacht Hydrodynamics & 0.01 & 1500 & 8 \\

    \bottomrule
  \end{tabular}
    \caption{Hyperparameters used to train deep split ensembles}\label{dse_hyperparameters}
\end{table}

\subsection{Alzheimer's dementia (AD) - dataset, model and hyperparameters}\label{appendix_dementia}

\subsection*{Dataset}
The ADReSS (Alzheimer's Dementia Recognition through Spontaneous Speech) dataset, available through the benchmark DementiaBank database upon access request, is a standardized and balanced dataset of 156 speech samples, each from a unique subject, matched for age and gender. The dataset consists of speech recordings and transcripts of spoken picture descriptions elicited from participants through the Cookie Theft picture from the Boston Diagnostic Aphasia Exam. The dataset also provides corresponding Mini-Mental Status Examination (MMSE) scores, ranging from 0 to 30, of the subjects, which offers a way to quantify cognitive function and screen for cognitive loss by testing the individuals’ orientation, attention, calculation, recall, language and motor skills. These scores are used as labels for the regression task. A standardized train-test split of around 70\%-30\% (108 and 48 subjects) is provided by this dataset. The test set is held-out until final evaluation, and the train set is split into train and validation sets for training models. This dataset was used for evaluating deep split ensembles in multi-modal settings (Section \ref{multimodal}).

\subsection*{Multi-modal feature engineering pipeline}
People with dementia show symptoms of cognitive decline, impairment in memory, communication and thinking. To include such domain knowledge and context, we devised an automated feature engineering pipeline that extracts several multi-modal cognitive and acoustic feature sets - interventions, disfluency, and acoustic. These three feature sets are then fed to a deep split NN, model architecture of which is shown in Figure \ref{alz_gauss_model}. Similar extracted features have been repeatedly used to propose speech recognition based solutions for automated detection of mild cognitive impairment from spontaneous speech. The three extracted feature sets are as follows:
\begin{itemize}
    \item \textit{Interventions features:} Cognitive features reflect upon potential loss of train of thoughts and context. Our system extracts the sequence of speakers from the transcripts, categorizing it as the subject or the interviewer. To accommodate for the variable length of these sequences, they are padded or truncated to length of 32 steps, found upon analyses and tuning of sequence lengths. These (subject, interviewer, or padding) are then one-hot encoded resulting in 32x3 input size for this feature set corresponding to every datapoint.
    \item \textit{Disfluency features:} A set of 11 distinct and carefully curated features from the transcripts; word rate, intervention rate, and 9 different kinds of pause rates, reflecting upon speech impediments like slurring and stuttering. These are normalized by the respective audio lengths and scaled thereafter.
    \item \textit{Acoustic features:} The ComParE 2013 feature set\footnote{\url{https://dl.acm.org/doi/abs/10.1145/2502081.2502224}, {https://dl.acm.org/doi/abs/10.1145/2502081.2502224}} was extracted from the audio samples using the open-sourced openSMILE v2.1 toolkit, widely used for affect analyses in speech. This provides a total of 6,373 features that include energy, MFCC, and voicing related low-level descriptors (LLDs), and other statistical functionals. This feature set encodes changes in speech of a person and has been used as an important noninvasive marker for AD detection. Our system standardizes this set of features using z-score normalization, and uses principal component analysis (PCA) to project the 6,373 features onto a low-dimensional space of 21 orthogonal features with highest variance. The number of orthogonal features was selected by analyzing the percentage of variance explained by each of the components.
\end{itemize}
\subsection*{Model architecture}
\begin{figure*}[ht!]
  \centering
  \includegraphics[width=0.7\linewidth]{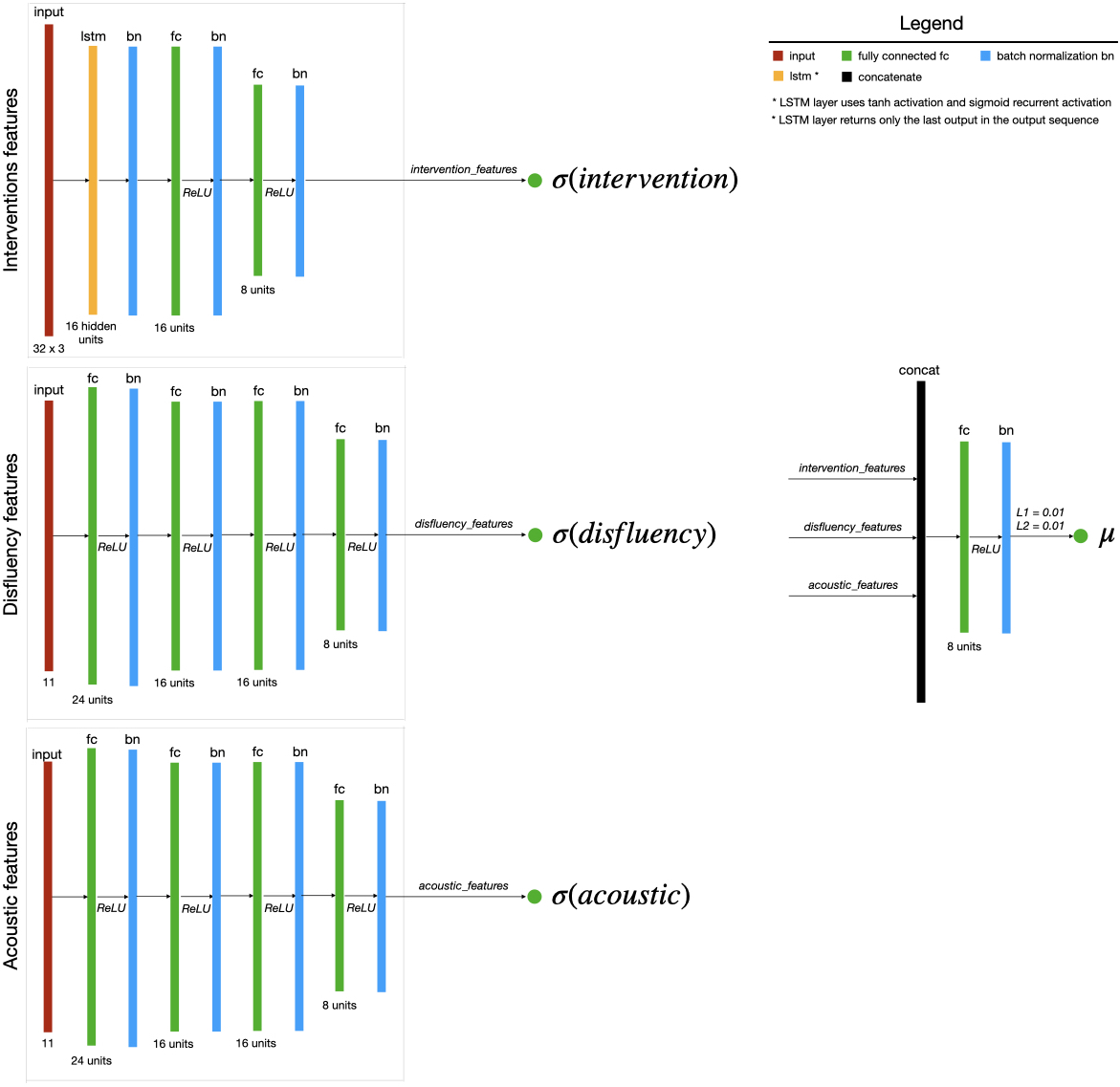}
  \caption{Model architecture of deep split NN for the ADReSS dataset.}
  \label{alz_gauss_model}
\end{figure*}

The three feature sets are then fed to a deep split NN, the model architecture of which is shown in Figure \ref{alz_gauss_model}.
\subsection*{Setup and hyperparameters}
Table \ref{alz_table_setup} shows the setup and hyperparameters for the ADReSS dataset. Best model was saved upon monitoring negative log-likelihood of the validation set (val NLL).
\begin{table}[ht!]
\setlength\tabcolsep{10pt}
  \centering
  \begin{tabular}{ll}
    \toprule
    Train-val split & 80\%-20\% (86-22 datapoints)  \\
    Heldout-test set & 48 datapoints \\
    Optimizer     & Adam    \\
    Learning rate & 0.001 \\
    Batch size & 8 \\
    \bottomrule
  \end{tabular}
    \caption{Setup and hyperparameters for the ADReSS dataset}
  \label{alz_table_setup}
\end{table}

\section{Results on benchmark regression datasets}\label{appendix_baseline}
\subsection{Comparison with state-of-the-art methods}\label{appendix_vipbp}
Tables \ref{full_comp_rmse} and \ref{full_comp_nll} compare the RMSE and NLL\footnote{NLLs of Deep Split Ensembles in Tables \ref{full_comp_nll} and \ref{var_table_res_rmse} are averaged over feature clusters of corresponding datasets. See Appendix \ref{appendix_nlls} for an exhaustive list of cluster-wise predictive uncertainty estimates for all datasets.\label{tab_fn_exhaustive}} of our method on benchmark regression datasets with other state-of-the-art methods - particularly VI \cite{graves2011practical}, PBP \cite{hernandez2015probabilistic}, MC-dropout \cite{gal2016dropout}, deterministic VI (DVI) \cite{wu2018deterministic} and subspace inference (SI) \cite{izmailov2020subspace}.

\begin{table*}[ht!]
  \small
  \centering
  \begin{tabular}{lcccccccc}
    \toprule
\multirow{2}{*}{Datasets} & \multirow{2}{*}{VI} & \multirow{2}{*}{PBP} & MC & \multirow{2}{*}{SI} & \multirow{2}{*}{RIO} & Deep & Anchored & Deep Split \\
 &  &  & Dropout & & & Ensembles & Ensembling & Ensembles \\ 
\toprule
Boston & 4.32 $\pm$ 0.29 & 3.01 $\pm$ 0.18 & 2.97 $\pm$ 0.85 & 3.45$\pm$0.95 & -- & 3.28 $\pm$ 1.00 & 3.09 $\pm$ 0.17 & \textbf{2.53 $\pm$ 0.15} \\
Concrete & 7.13 $\pm$ 0.12 & 5.67 $\pm$ 0.09 & 5.23 $\pm$ 0.53 & 5.19 $\pm$ 0.44 & 5.97 $\pm$ 0.48 & 6.03 $\pm$ 0.58 & 4.87 $\pm$ 0.11 & \textbf{4.40 $\pm$ 0.10} \\
Energy & 2.65 $\pm$ 0.08 & 1.80 $\pm$ 0.05 & 1.66 $\pm$ 0.19 & 1.59 $\pm$ 0.27 & 0.70 $\pm$ 0.38 & 2.09 $\pm$ 0.29 & \textbf{0.35 $\pm$ 0.01} & 0.41 $\pm$ 0.02 \\
Kin8nm & 0.10 $\pm$ 0.00 & 0.10 $\pm$ 0.00 & 0.10 $\pm$ 0.00 & -- & -- & 0.09 $\pm$ 0.00 & \textbf{0.07 $\pm$ 0.00} & 0.19 $\pm$ 0.00 \\
Naval & 0.01 $\pm$ 0.00 & 0.01 $\pm$ 0.00 & 0.01 $\pm$ 0.00 & \textbf{0.00 $\pm$ 0.00} & -- & \textbf{0.00 $\pm$ 0.00} & \textbf{0.00 $\pm$ 0.00} & \textbf{0.00 $\pm$ 0.00} \\
Power & 4.32 $\pm$ 0.04 & 4.12 $\pm$ 0.03 & \textbf{4.02 $\pm$ 0.18} & -- & 4.05 $\pm$ 0.12 & 4.11 $\pm$ 0.17 & 4.07 $\pm$ 0.04 & 4.04 $\pm$ 0.05 \\
Protein & 4.84 $\pm$ 0.03 & 4.73 $\pm$ 0.01 & 4.36 $\pm$ 0.04 & -- & 4.08 $\pm$ 0.06 & 4.71 $\pm$ 0.06 & 4.36 $\pm$ 0.02 & \textbf{4.05 $\pm$ 0.03} \\
Wine & 0.65 $\pm$ 0.01 & 0.64 $\pm$ 0.01 & 0.62 $\pm$ 0.04 & -- & 0.67 $\pm$ 0.03 & 0.64 $\pm$ 0.04 & 0.63 $\pm$ 0.01 & \textbf{0.60 $\pm$ 0.02} \\
Yacht & 6.89 $\pm$ 0.67 & 1.02 $\pm$ 0.05 & 1.11 $\pm$ 0.38 & 0.97 $\pm$ 0.37 & 1.46 $\pm$ 0.49 & 1.58 $\pm$ 0.48 & \textbf{0.57 $\pm$ 0.05} & 0.86 $\pm$ 0.07 \\
    \bottomrule
  \end{tabular}
  \caption{Results on UCI regression datasets comparing RMSE\footnotemark}\label{full_comp_rmse}
\end{table*}

\begin{table*}[ht!]
  \small
  \centering
  \begin{tabular}{lccccccccc}
    \toprule
\multirow{2}{*}{Datasets} & \multirow{2}{*}{VI} & \multirow{2}{*}{PBP} & MC & \multirow{2}{*}{dVI} & \multirow{2}{*}{SI} & \multirow{2}{*}{RIO} & Deep & Anchored & Deep Split \\
 &  &  & Dropout & & & & Ensembles & Ensembling & Ensembles \\ 
\toprule
Boston & 2.90$\pm$ 0.07 & 2.57$\pm$0.09 & 2.46$\pm$0.25 & 2.41$\pm$0.02 & 2.71$\pm$0.13 & -- & 2.41$\pm$0.25 & 2.52$\pm$0.05 & \textbf{2.23$\pm$0.04} \\
Concrete & 3.39$\pm$0.02 & 3.16$\pm$0.02 & 3.04$\pm$0.09 & 3.06$\pm$0.01 & 3.00$\pm$0.08 & 3.24$\pm$0.10 & 3.06$\pm$0.18 & 2.97$\pm$0.02 & \textbf{2.85$\pm$0.02} \\
Energy & 2.39$\pm$0.03 & 2.04$\pm$0.02 & 1.99$\pm$0.09 & 1.01$\pm$0.06 & 1.56$\pm$1.24 & 1.03$\pm$0.35 & 1.38$\pm$0.22 & 0.96$\pm$0.13 & \textbf{0.28$\pm$0.11} \\
Kin8nm & -0.90$\pm$0.01 & -0.90$\pm$0.01 & -0.95$\pm$0.03 & -1.13$\pm$0.00 & -- & -- & \textbf{-1.20$\pm$0.02} & -1.09$\pm$0.01 & -0.20$\pm$0.02 \\
Naval & -3.73$\pm$0.12 & -3.73$\pm$0.01 & -3.80$\pm$0.05 & -6.29$\pm$0.04 & -6.54$\pm$0.09 & -- & -5.63$\pm$0.05 & \textbf{-7.17$\pm$0.03} & -5.28$\pm$0.02 \\
Power & 2.89$\pm$0.01 & 2.84$\pm$0.01 & 2.80$\pm$0.05 & 2.80$\pm$0.00 & -- & 2.81$\pm$0.03 & 2.79$\pm$0.04 & 2.83$\pm$0.01 & \textbf{2.78$\pm$0.01} \\
Protein & 2.99$\pm$0.01 & 2.97$\pm$0.00 & 2.89$\pm$0.01 & 2.85$\pm$0.01 & -- & 2.82$\pm$0.01 & 2.83$\pm$0.02 & 2.89$\pm$0.01 & \textbf{2.76$\pm$0.00} \\
Wine & 0.98$\pm$0.01 & 0.97$\pm$0.01 & 0.93$\pm$0.06 & 0.90$\pm$0.01 & -- & 1.09$\pm$0.10 & 0.94$\pm$0.12 & 0.95$\pm$0.01 & \textbf{0.89$\pm$0.02} \\
Yacht & 3.44$\pm$0.16 & 1.63$\pm$0.02 & 1.55$\pm$0.12 & 0.47$\pm$0.03 & 0.225$\pm$0.40 & 1.79$\pm$0.88 & 1.18$\pm$0.21 & \textbf{0.37$\pm$0.08} & 0.90$\pm$0.09 \\
    \bottomrule
  \end{tabular} 
  \caption{Results on UCI regression datasets comparing NLL}\label{full_comp_nll}
\end{table*}

\subsection{Results of variants of deep split ensembles}
\label{appendix_variants_results}
Table \ref{var_table_res_rmse} shows the RMSE and NLL\footref{tab_fn_exhaustive} results of variants of deep split ensemble, with different number of models $E = 1, 5, 10$ in the parallel ensemble. We observe that $E = 5$ is the best of all considering the performance results and the computational overhead.

\begin{table*}[ht!]
\small
  
  \centering
  \setlength\tabcolsep{4.3pt}
  \begin{tabular}{lcccccc}
    \toprule
     Datasets & \multicolumn{3}{c}{RMSE} & \multicolumn{3}{c}{NLL} \\
    \cmidrule(r){2-4}
    \cmidrule(r){5-7}
     & $E = 1$ & $E = 5$ & $E = 10$ & $E = 1$ & $E = 5$ & $E = 10$ \\  
    \midrule
Boston & 2.76 $\pm$ 1.16 & \textbf{2.53 $\pm$ 0.15} & 2.60 $\pm$ 1.31 & 2.33 $\pm$ 0.26 & \textbf{2.23 $\pm$ 0.04} & 2.25 $\pm$ 0.28 \\
Concrete & 4.52 $\pm$ 0.55 & \textbf{4.40 $\pm$ 0.10} & 4.63 $\pm$ 0.57 & 2.89 $\pm$ 0.12 & \textbf{2.85 $\pm$ 0.02} & 2.87 $\pm$ 0.13 \\
Energy & 0.43 $\pm$ 0.06 & \textbf{0.41 $\pm$ 0.02} & 0.44 $\pm$ 0.08 & 0.31 $\pm$ 0.19 & \textbf{0.28 $\pm$ 0.11} & 0.33 $\pm$ 0.25 \\
Kin8nm & 0.20 $\pm$ 0.00 & \textbf{0.19 $\pm$ 0.00} & \textbf{0.19 $\pm$ 0.00} & -0.18 $\pm$ 0.03 & \textbf{-0.20 $\pm$ 0.02} & \textbf{-0.20 $\pm$ 0.02} \\
Naval & \textbf{0.00 $\pm$ 0.00} & \textbf{0.00 $\pm$ 0.00} & \textbf{0.00 $\pm$ 0.00} & -5.21 $\pm$ 0.07 & \textbf{-5.28 $\pm$ 0.02} & -5.26 $\pm$ 0.03 \\
Power & 4.06 $\pm$ 0.25 & \textbf{4.04 $\pm$ 0.05} & 4.07 $\pm$ 0.26 & 2.80 $\pm$ 0.05 & \textbf{2.78 $\pm$ 0.01} & 2.83 $\pm$ 0.06 \\
Protein & 4.14 $\pm$ 0.03 & \textbf{4.05 $\pm$ 0.03} & 4.09 $\pm$ 0.04 & 2.79 $\pm$ 0.01 & \textbf{2.76 $\pm$ 0.00} & 2.80 $\pm$ 0.02 \\
Wine & 0.63 $\pm$ 0.12 & \textbf{0.60 $\pm$ 0.02} & \textbf{0.60 $\pm$ 0.05} & 0.95 $\pm$ 0.17 & \textbf{0.89 $\pm$ 0.02} & 0.92 $\pm$ 0.10 \\
Yacht & 0.89 $\pm$ 0.41 & \textbf{0.86 $\pm$ 0.07} & 0.90 $\pm$ 0.46 & 0.93 $\pm$ 0.25 & \textbf{0.90 $\pm$ 0.09} & 0.93 $\pm$ 0.20 \\
    \bottomrule
  \end{tabular}
  \caption{Results of deep split ensembles with different no. of models $E$ in the parallel ensemble.}\label{var_table_res_rmse}
\end{table*}

\subsection{Cluster-wise NLLs}\label{appendix_nlls}
Table \ref{clusterwise_nll_table} shows an exhaustive list of NLLs of all clusters for different number of models $E$ in the parallel ensemble. Table \ref{var_table_res_rmse} shows the average of these cluster-wise NLLs corresponding to each dataset. Table \ref{hc_perclusternll_appendix} shows NLL per cluster for DSE, AEPC, DEPC.
\begin{table*}[ht!]
  \small
  \centering
  \begin{tabular}{lcccc}
    \toprule
        \multirow{2}{*}{Dataset}     & \multirow{2}{*}{Cluster}     & \multicolumn{3}{c}{Cluster-wise NLL} \\
        \cmidrule(r){3-5}
         &   & $E = 1$ & $E = 5$ & $E = 10$\\
    \toprule
\multirow{3}{*}{Boston}   
& 1 & 2.29 $\pm$ 0.07 & 2.23 $\pm$ 0.04 & \textbf{2.21 $\pm$ 0.03} \\
& 2 & 2.33 $\pm$ 0.08 & \textbf{2.20 $\pm$ 0.05} & 2.24 $\pm$ 0.03 \\
& 3 & 2.37 $\pm$ 0.03 & \textbf{2.26 $\pm$ 0.04} & 2.30 $\pm$ 0.05 \\
 \midrule
\multirow{3}{*}{Concrete}   
& 1 & 2.87 $\pm$ 0.01 & \textbf{2.84 $\pm$ 0.02} & 2.87 $\pm$ 0.03\\
& 2 & 2.89 $\pm$ 0.02 & \textbf{2.85 $\pm$ 0.02} & 2.91 $\pm$ 0.04\\
& 3 & 2.91 $\pm$ 0.03 & \textbf{2.87 $\pm$ 0.01} & 2.93 $\pm$ 0.02\\
 \midrule
\multirow{2}{*}{Energy}   
& 1 & 0.28 $\pm$ 0.20 & \textbf{0.26 $\pm$ 0.11} & 0.29 $\pm$ 0.27\\
& 2 & 0.34 $\pm$ 0.18 & \textbf{0.30 $\pm$ 0.11} & 0.37 $\pm$ 0.24\\
\midrule
\multirow{8}{*}{Kin8nm} 
& 1 & -0.18 $\pm$ 0.03 & -0.19 $\pm$ 0.02 & \textbf{-0.20 $\pm$ 0.02}\\
& 2 & -0.18 $\pm$ 0.03 & -0.19 $\pm$ 0.03 & \textbf{-0.20 $\pm$ 0.02}\\
& 3 & -0.18 $\pm$ 0.03 & -0.19 $\pm$ 0.02 & \textbf{-0.20 $\pm$ 0.03}\\
& 4 & -0.18 $\pm$ 0.03 & -0.19 $\pm$ 0.03 & \textbf{-0.20 $\pm$ 0.02}\\
& 5 & -0.18 $\pm$ 0.03 & \textbf{-0.20 $\pm$ 0.03} & \textbf{-0.20 $\pm$ 0.02}\\
& 6 & -0.18 $\pm$ 0.03 & -0.19 $\pm$ 0.02 & \textbf{-0.20 $\pm$ 0.02}\\
& 7 & -0.18 $\pm$ 0.02 & \textbf{-0.20 $\pm$ 0.03} & \textbf{-0.20 $\pm$ 0.02}\\
& 8 & -0.18 $\pm$ 0.02 & \textbf{-0.21 $\pm$ 0.02} & \textbf{-0.21 $\pm$ 0.02}\\
\midrule
\multirow{2}{*}{Naval}   
& 1 & -5.19 $\pm$ 0.08 & \textbf{-5.25 $\pm$ 0.02} & -5.23 $\pm$ 0.04\\
& 2 & -5.24 $\pm$ 0.06 & \textbf{-5.31 $\pm$ 0.02} & -5.29 $\pm$ 0.02\\
\midrule
\multirow{2}{*}{Power}   
& 1 & 2.80 $\pm$ 0.06 & \textbf{2.79 $\pm$ 0.01} & 2.83 $\pm$ 0.07\\
& 2 & 2.81 $\pm$ 0.05 & \textbf{2.80 $\pm$ 0.01} & 2.84 $\pm$ 0.05\\
\midrule
\multirow{3}{*}{Protein}   
& 1 & 2.73 $\pm$ 0.01 & \textbf{2.68 $\pm$ 0.01} & 2.72 $\pm$ 0.02\\
& 2 & 2.83 $\pm$ 0.00 & \textbf{2.80 $\pm$ 0.00} & 2.84 $\pm$ 0.01\\
& 3 & 2.83 $\pm$ 0.00 & \textbf{2.81 $\pm$ 0.00} & 2.86 $\pm$ 0.02\\
\midrule
\multirow{5}{*}{Wine}   
& 1 & 0.95 $\pm$ 0.05 & \textbf{0.91 $\pm$ 0.03} & 0.94 $\pm$ 0.03\\
& 2 & 0.99 $\pm$ 0.08 & \textbf{0.92 $\pm$ 0.01} & \textbf{0.92 $\pm$ 0.08}\\
& 3 & 0.94 $\pm$ 0.04 & \textbf{0.91 $\pm$ 0.02} & \textbf{0.91 $\pm$ 0.08}\\
& 4 & 0.94 $\pm$ 0.05 & \textbf{0.90 $\pm$ 0.04} & 0.94 $\pm$ 0.03\\
& 5 & 0.95 $\pm$ 0.07 & \textbf{0.91 $\pm$ 0.01} & \textbf{0.91 $\pm$ 0.08}\\
\midrule
\multirow{2}{*}{Yacht}   
& 1 & 1.48 $\pm$ 0.20 & \textbf{1.45 $\pm$ 0.10} & 1.49 $\pm$ 0.12\\
& 2 & 0.39 $\pm$ 0.10 & \textbf{0.36 $\pm$ 0.09} & 0.38 $\pm$ 0.08\\
    \bottomrule
  \end{tabular}
  \caption{NLLs of all clusters for different number of models $E$ in the parallel ensemble.}
  \label{clusterwise_nll_table}
\end{table*}

\begin{table*}[t]
\small
\centering
\begin{tabular}{lccccccc}
\toprule
Datasets & \multicolumn{3}{c}{RMSE} & Clusters & \multicolumn{3}{c}{Cluster-wise NLL} \\
\cmidrule(r){2-4}
\cmidrule(r){6-8}
 & DEPC & AEPC & Deep Split Ens. & & DEPC & AEPC & Deep Split Ens. \\
 \midrule
\multirow{3}{*}{Boston} 
 & \multirow{3}{*}{5.11 $\pm$ 1.06} & \multirow{3}{*}{4.93 $\pm$ 1.03} & \multirow{3}{*}{\textbf{2.53 $\pm$ 0.15}} & 1 & 2.91 $\pm$ 0.16 & 3.87 $\pm$ 0.82 & 2.23 $\pm$ 0.04 \\
 & & & & 2 & 2.82 $\pm$ 0.16 & 3.99 $\pm$ 0.94 & 2.20 $\pm$ 0.05 \\
 & & & & 3 & 3.29 $\pm$ 0.10 & 4.23 $\pm$ 1.06 & 2.26 $\pm$ 0.04 \\ \midrule
\multirow{3}{*}{Concrete} 
 & \multirow{3}{*}{10.22 $\pm$ 0.82} 
 & \multirow{3}{*}{10.40 $\pm$ 0.93} 
 & \multirow{3}{*}{\textbf{4.40 $\pm$ 0.10}} 
 & 1 & 3.77 $\pm$ 0.05 & 5.75 $\pm$ 0.61 & 2.84 $\pm$ 0.02 \\
 & & & & 2 & 3.79 $\pm$ 0.10 & 5.68 $\pm$ 0.60 & 2.85 $\pm$ 0.02 \\
 & & & & 3 & 3.81 $\pm$ 0.05 & 5.83 $\pm$ 0.61 & 2.87 $\pm$ 0.01 \\ 
 \midrule
\multirow{2}{*}{Power} 
 & \multirow{2}{*}{7.51 $\pm$ 0.20} 
 & \multirow{2}{*}{7.53 $\pm$ 0.18} 
 & \multirow{2}{*}{\textbf{4.04 $\pm$ 0.05}} 
 & 1 & 3.90 $\pm$ 0.09 & 4.20 $\pm$ 0.09 & 2.79 $\pm$ 0.01 \\
 & & & & 2 & 3.61 $\pm$ 0.02 & 4.18 $\pm$ 0.09 & 2.80 $\pm$ 0.01 \\
 \midrule
 \multirow{3}{*}{Protein} 
 & \multirow{3}{*}{5.04 $\pm$ 0.01} 
 & \multirow{3}{*}{5.00 $\pm$ 0.02} 
 & \multirow{3}{*}{\textbf{4.05 $\pm$ 0.03}} 
 & 1 & 3.83 $\pm$ 0.10 & 3.04 $\pm$ 0.00 & 2.68 $\pm$ 0.01 \\
 & & & & 2 & 3.04 $\pm$ 0.00 & 3.05 $\pm$ 0.00 & 2.80 $\pm$ 0.00 \\
 & & & & 3 & 3.06 $\pm$ 0.00 & 3.05 $\pm$ 0.00 & 2.81 $\pm$ 0.00 \\ 
 \midrule
\multirow{5}{*}{Wine} 
 & \multirow{5}{*}{0.67 $\pm$ 0.05} 
 & \multirow{5}{*}{0.69 $\pm$ 0.04} 
 & \multirow{5}{*}{\textbf{0.60 $\pm$ 0.02}} 
 & 1 & 1.02 $\pm$ 0.07 & 1.08 $\pm$ 0.10 & 0.91 $\pm$ 0.03 \\
 & & & & 2 & 1.01 $\pm$ 0.06 & 1.08 $\pm$ 0.09 & 0.92 $\pm$ 0.01 \\
 & & & & 3 & 1.01 $\pm$ 0.06 & 1.08 $\pm$ 0.09 & 0.91 $\pm$ 0.02 \\
 & & & & 4 & 1.01 $\pm$ 0.06 & 1.09 $\pm$ 0.09 & 0.90 $\pm$ 0.04 \\
 & & & & 5 & 0.99 $\pm$ 0.07 & 1.09 $\pm$ 0.09 & 0.91 $\pm$ 0.01 \\

  \bottomrule
\end{tabular}
\caption{RMSE and Cluster-wise NLL values for DEPC, AEPC, DSE on UCI datasets}
\label{hc_perclusternll_appendix}
\end{table*}

\section{Experiments}
\subsection{Entropy analyses}\label{appendix_entropy}
\begin{figure*}[ht!]
\centering
\subfloat{%
    \includegraphics[width=0.01\linewidth, height=2.5cm, valign=t]{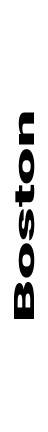}}\hspace{0.005mm}
    ~   
\subfloat{%
    \includegraphics[width=0.23\linewidth, valign=t]{figures/new_entropy_plots/de/boston_clusterwise_ood.png}}
    ~
\subfloat{%
    \includegraphics[width=0.23\linewidth, valign=t]{figures/new_entropy_plots/ae/boston_clusterwise_ood.png}}
    ~
\subfloat{%
    \includegraphics[width=0.23\linewidth, valign=t]{figures/new_entropy_plots/dse/boston/out_1.png}}
    ~
\subfloat{%
    \includegraphics[width=0.23\linewidth, valign=t]{figures/new_entropy_plots/dse/boston/out_2.png}}
    ~ \\
\subfloat{%
    \includegraphics[width=0.01\linewidth, height=2.8cm, valign=t]{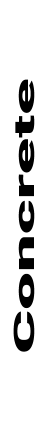}}\hspace{0.005mm}
    ~
\subfloat{%
    \includegraphics[width=0.23\linewidth, valign=t]{figures/new_entropy_plots/de/cement_clusterwise_ood.png}}
    ~
\subfloat{%
    \includegraphics[width=0.23\linewidth, valign=t]{figures/new_entropy_plots/ae/cement_clusterwise_ood.png}}
    ~
\subfloat{%
    \includegraphics[width=0.23\linewidth, valign=t]{figures/new_entropy_plots/dse/cement/out_1.png}}
    ~
\subfloat{%
    \includegraphics[width=0.23\linewidth, valign=t]{figures/new_entropy_plots/dse/cement/out_2.png}}
    ~\\
\subfloat{%
    \includegraphics[width=0.01\linewidth, height=2.8cm, valign=t]{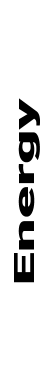}}\hspace{0.005mm}
    ~
\subfloat{%
    \includegraphics[width=0.23\linewidth, valign=t]{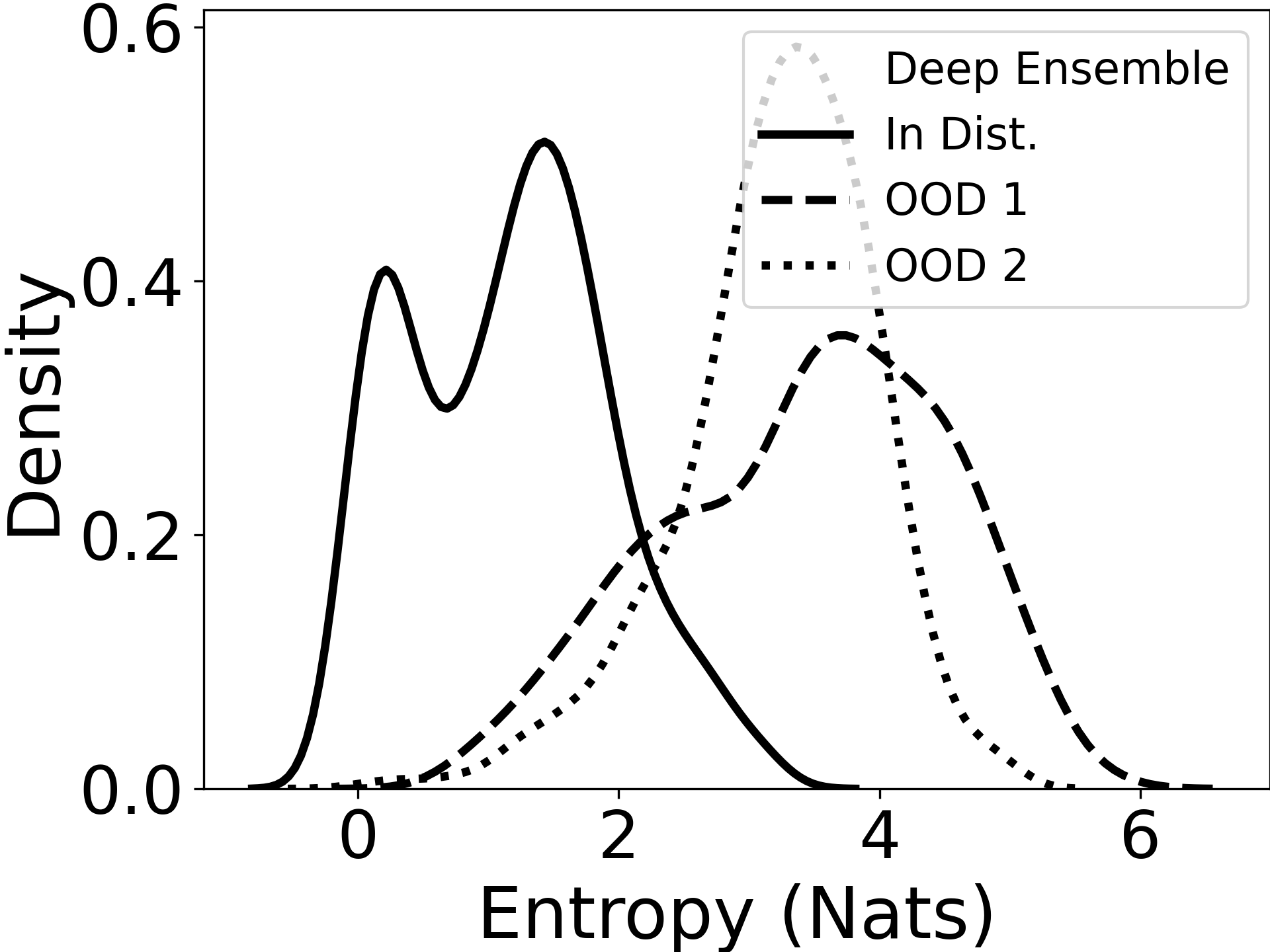}}
    ~
\subfloat{%
    \includegraphics[width=0.23\linewidth, valign=t]{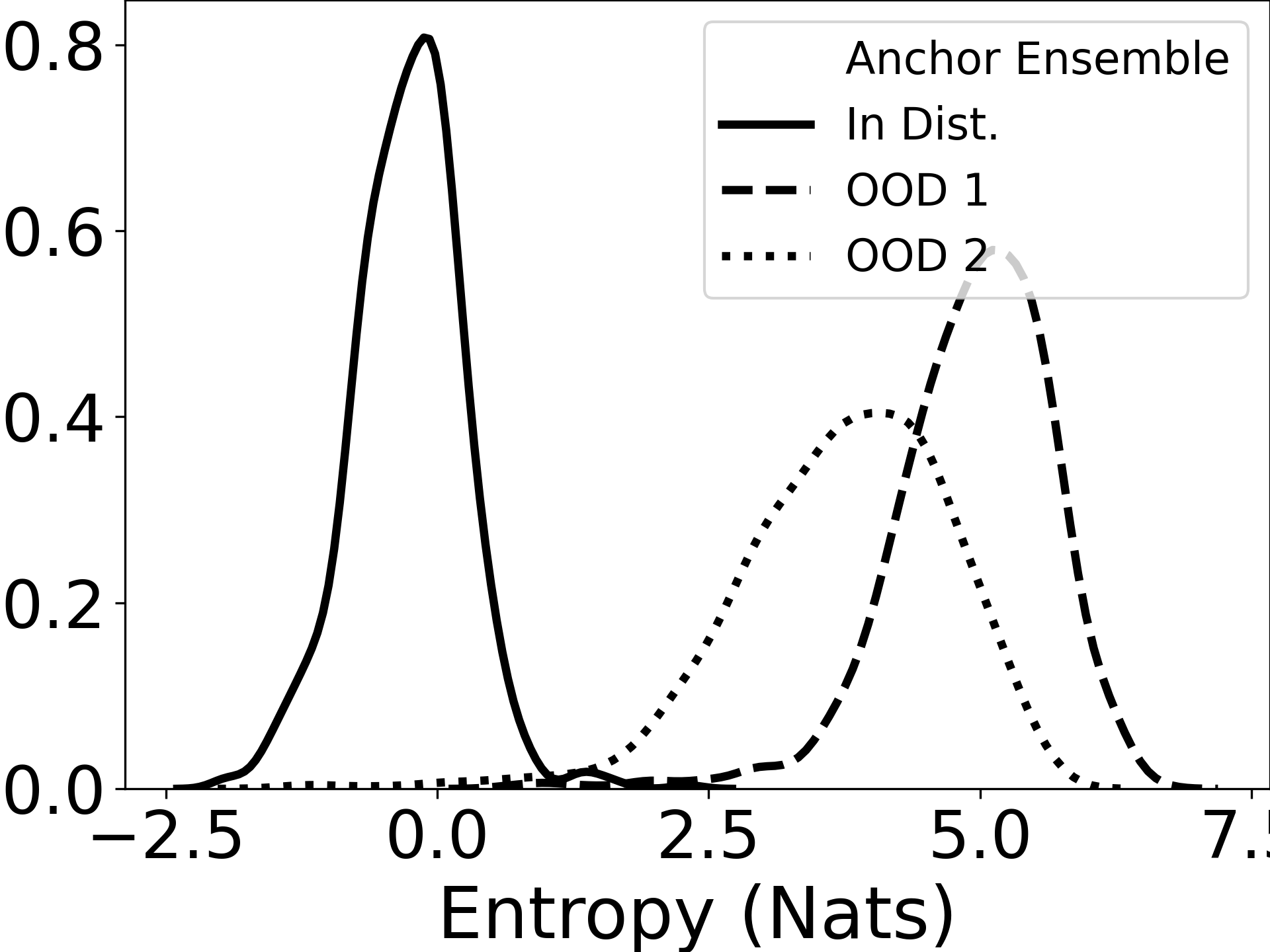}}
    ~
\subfloat{%
    \includegraphics[width=0.23\linewidth, valign=t]{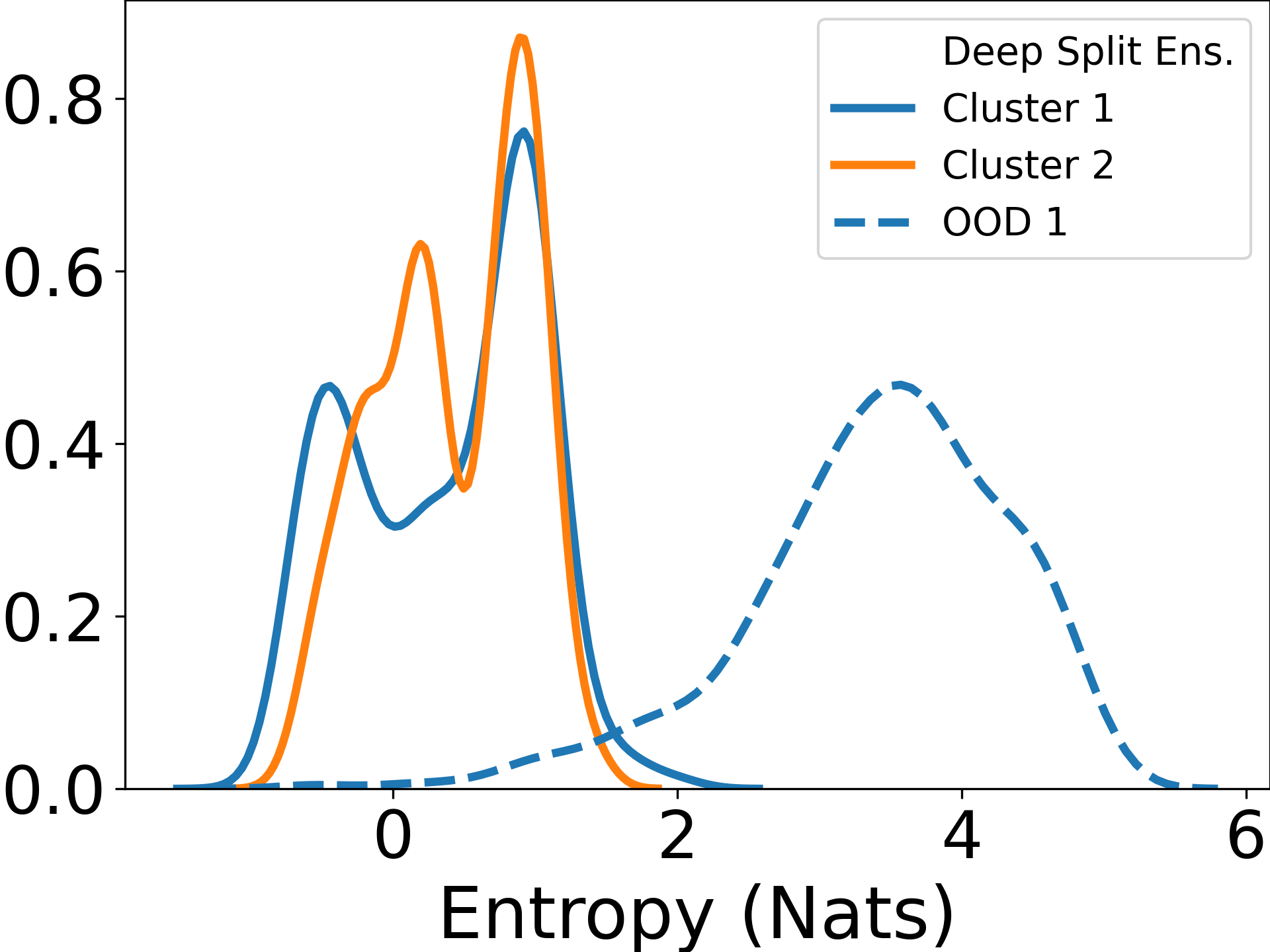}}
    ~
\subfloat{%
    \includegraphics[width=0.23\linewidth, valign=t]{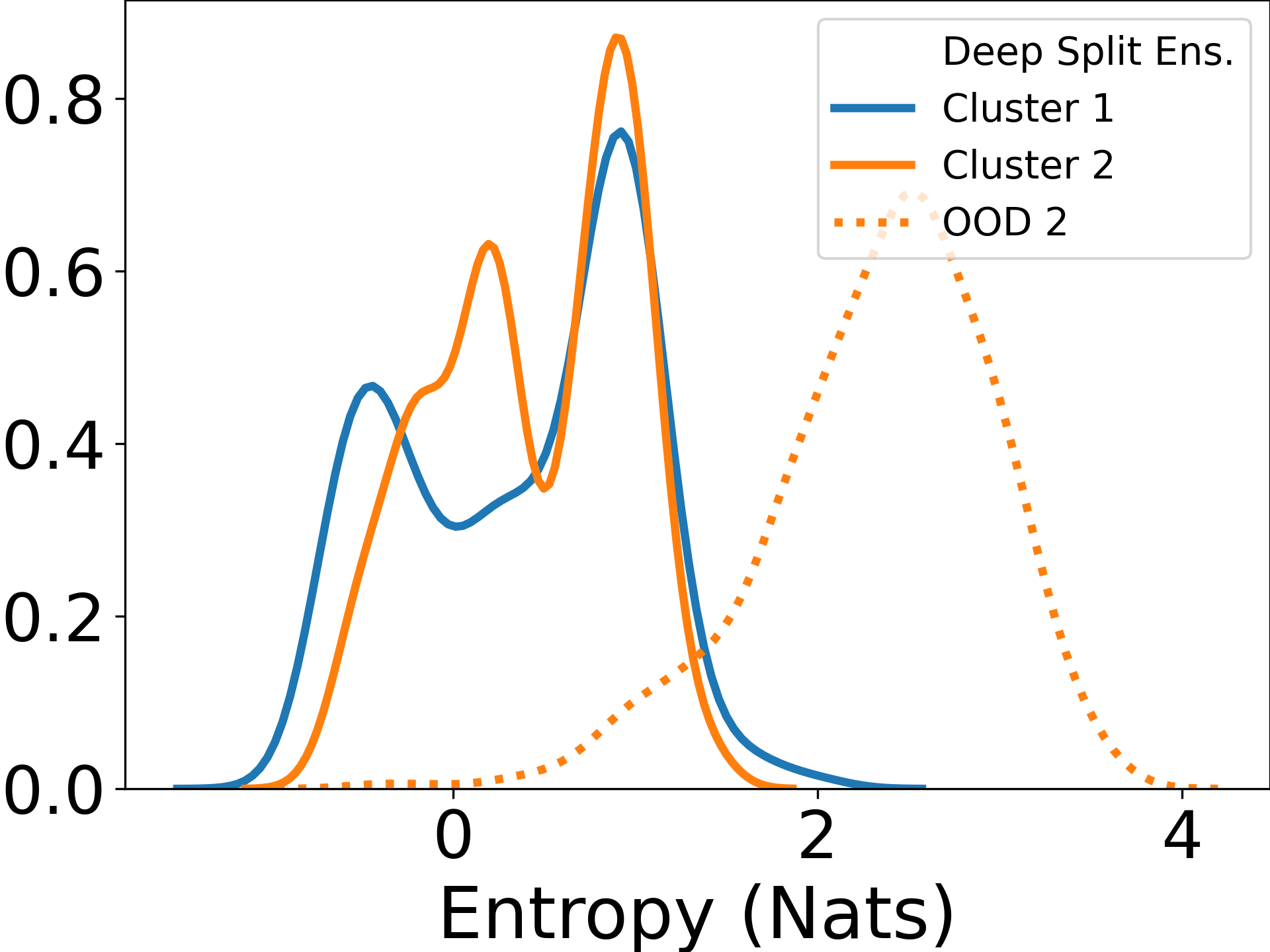}}
    ~ \\
\subfloat{%
    \includegraphics[width=0.01\linewidth, height=2.8cm, valign=t]{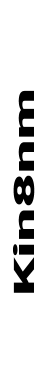}}\hspace{0.005mm}
    ~
\subfloat{%
    \includegraphics[width=0.23\linewidth, valign=t]{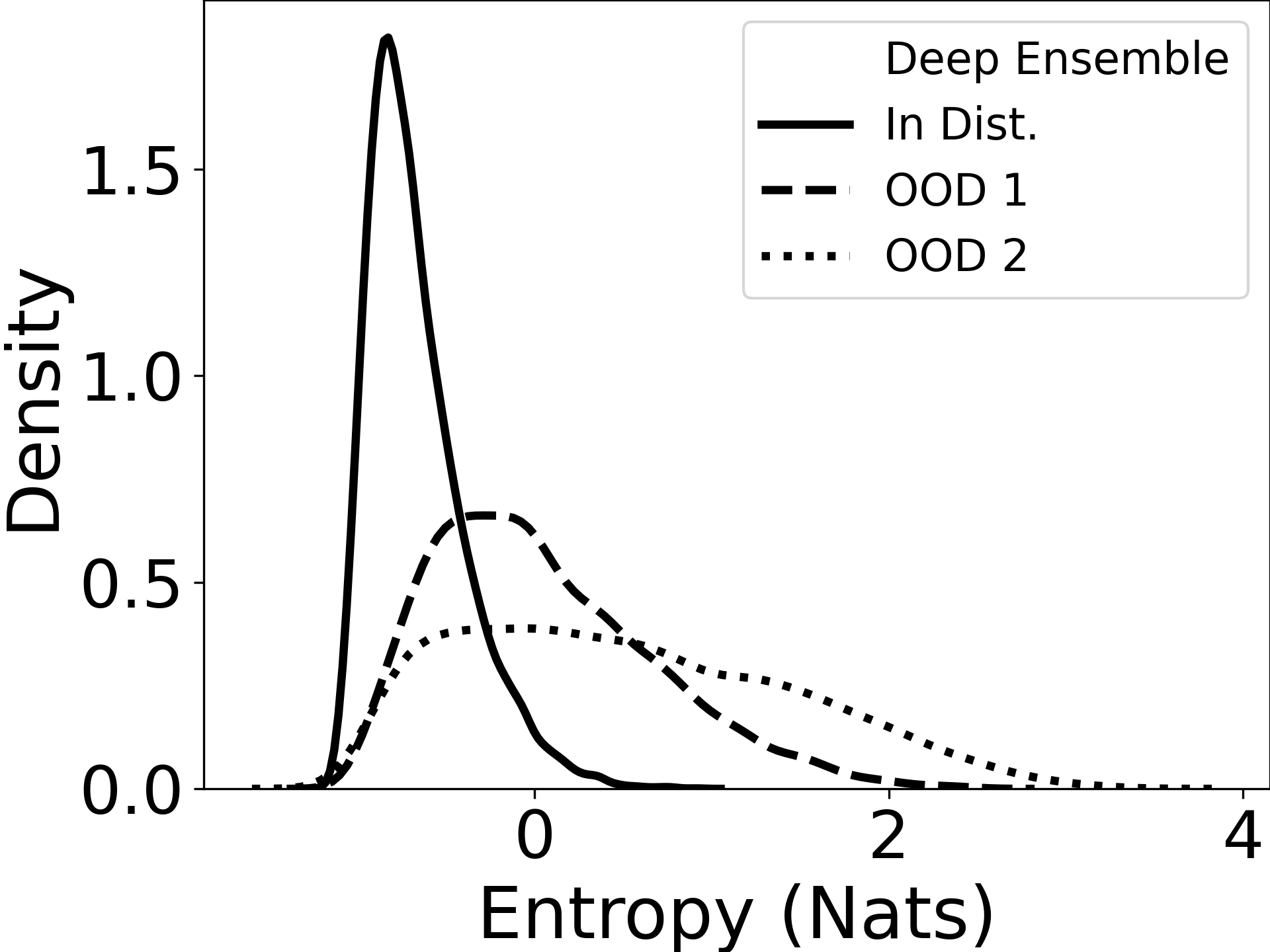}}
    ~
\subfloat{%
    \includegraphics[width=0.23\linewidth, valign=t]{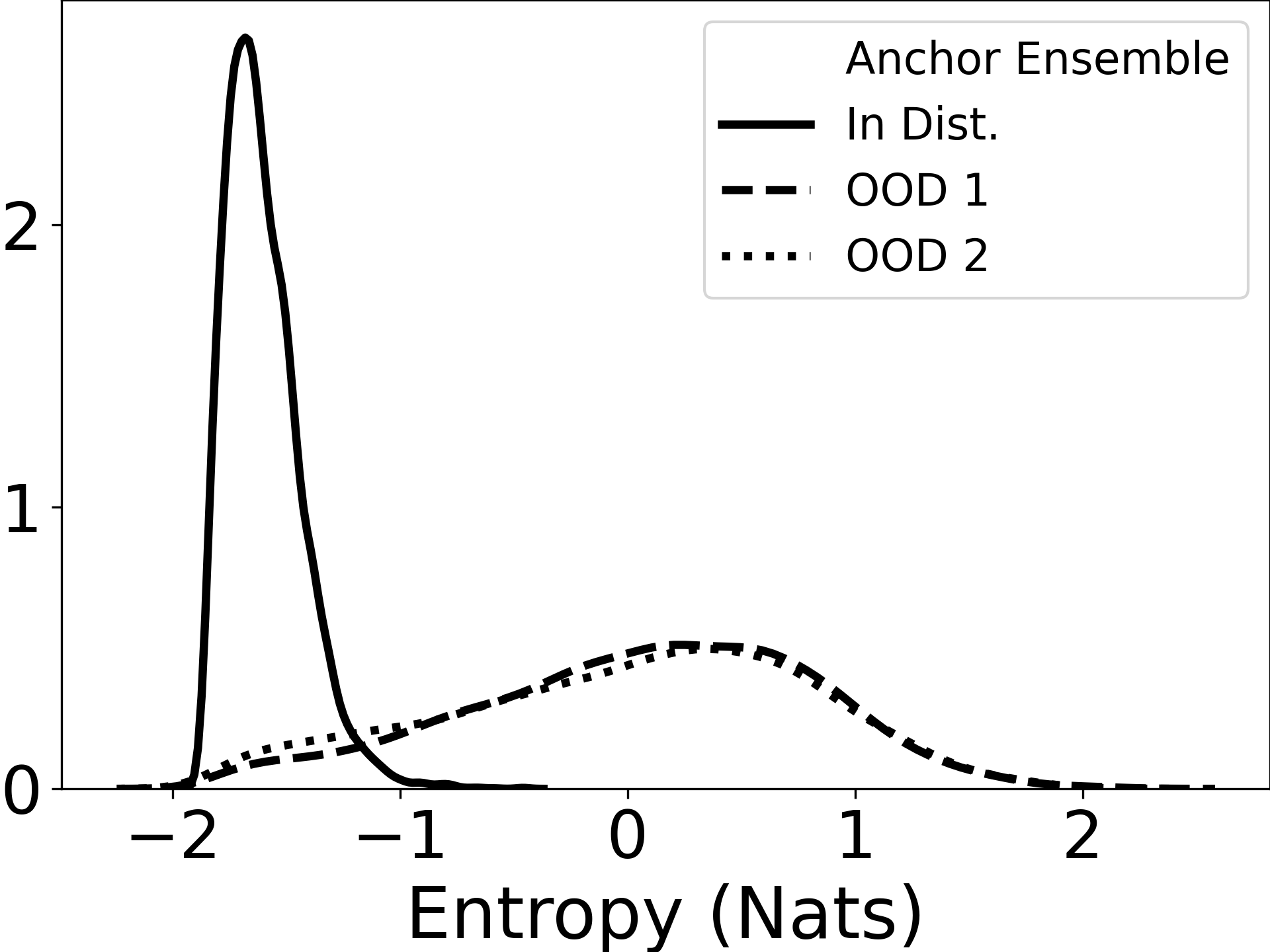}}
    ~
\subfloat{%
    \includegraphics[width=0.23\linewidth, valign=t]{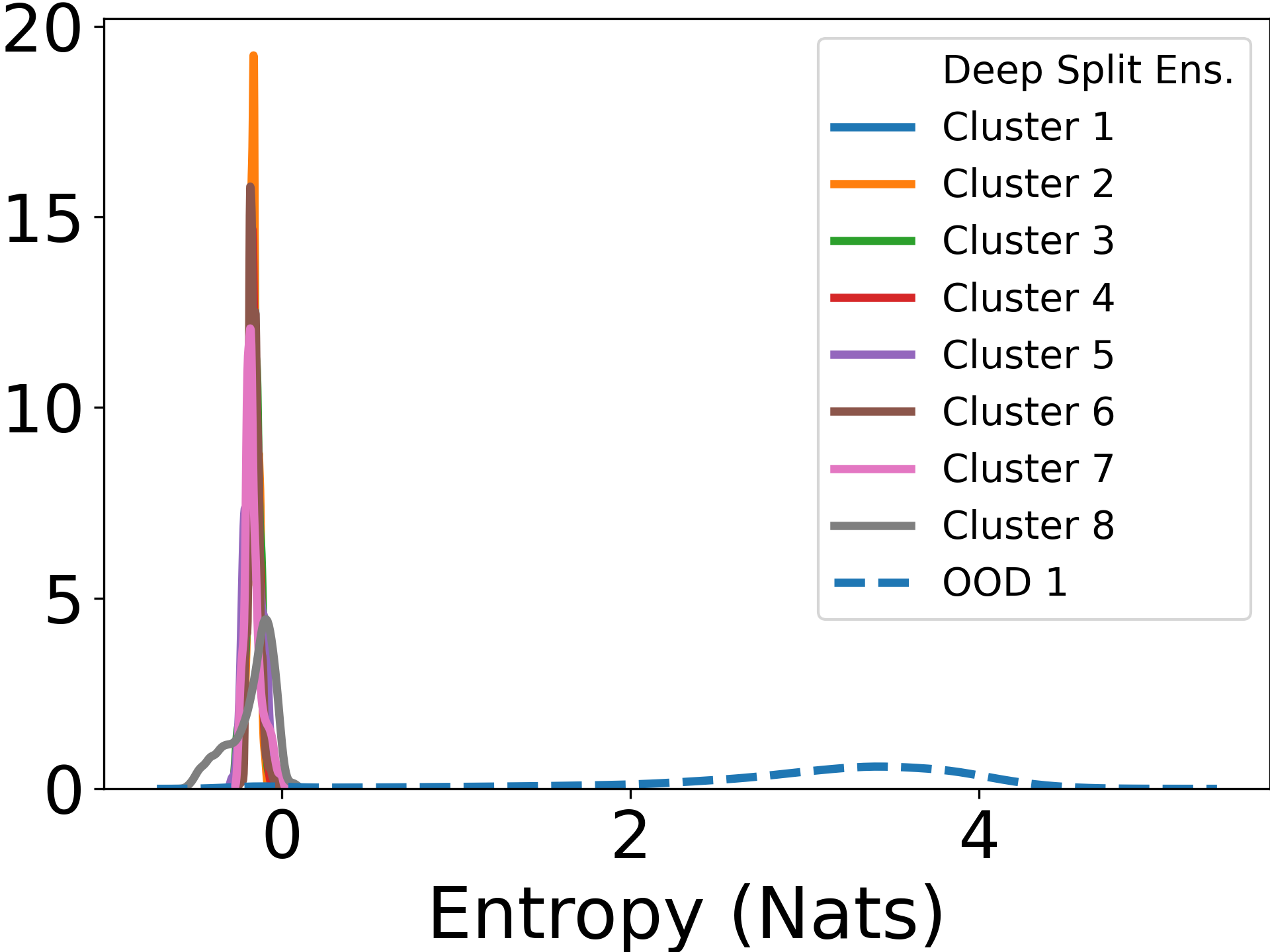}}
    ~
\subfloat{%
    \includegraphics[width=0.23\linewidth, valign=t]{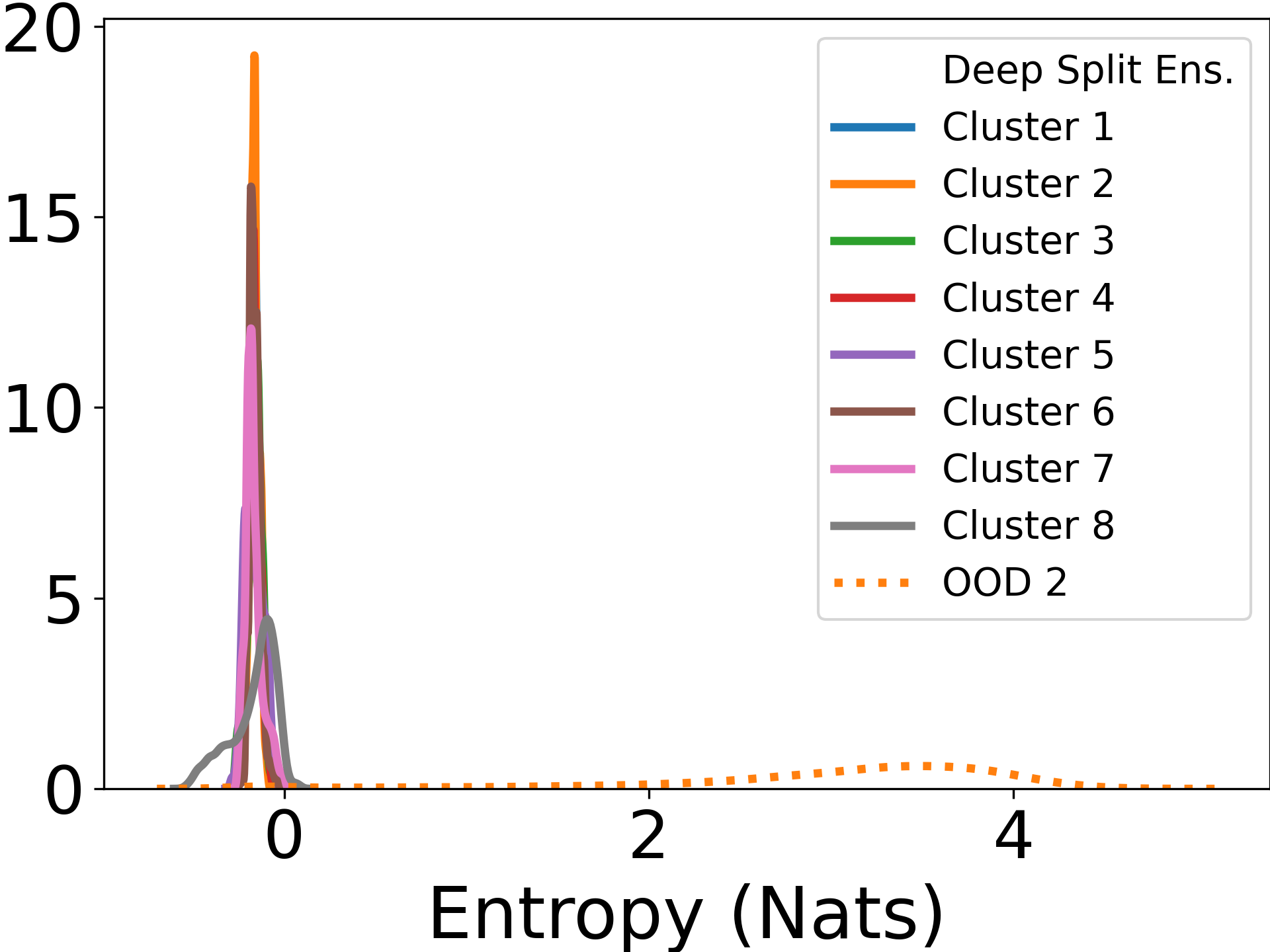}}
    ~  \\
\subfloat{%
    \includegraphics[width=0.01\linewidth, height=2.8cm, valign=t]{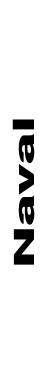}}\hspace{0.005mm}
    ~
\subfloat{%
    \includegraphics[width=0.23\linewidth, valign=t]{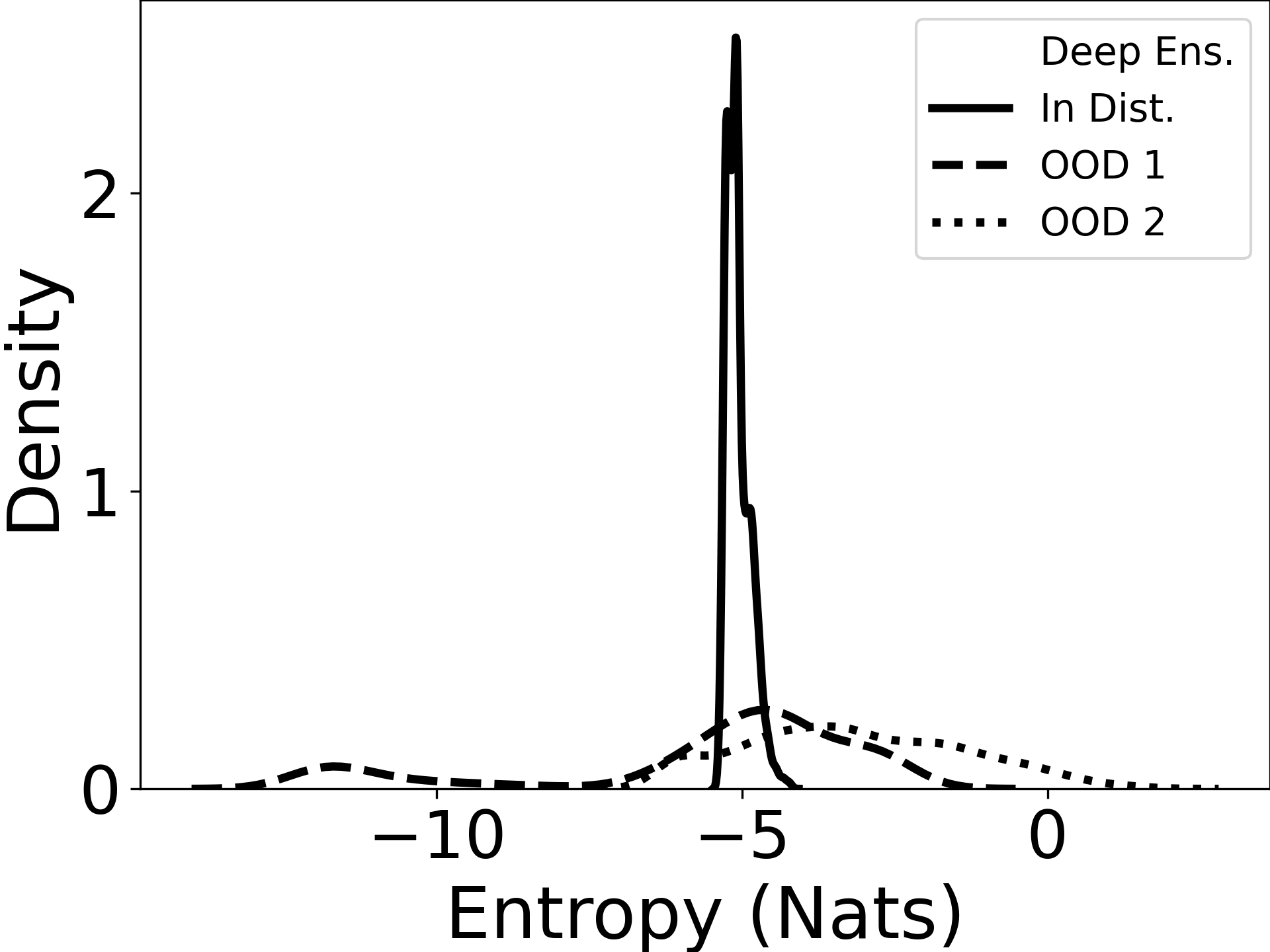}}
    ~
\subfloat{%
    \includegraphics[width=0.23\linewidth, valign=t]{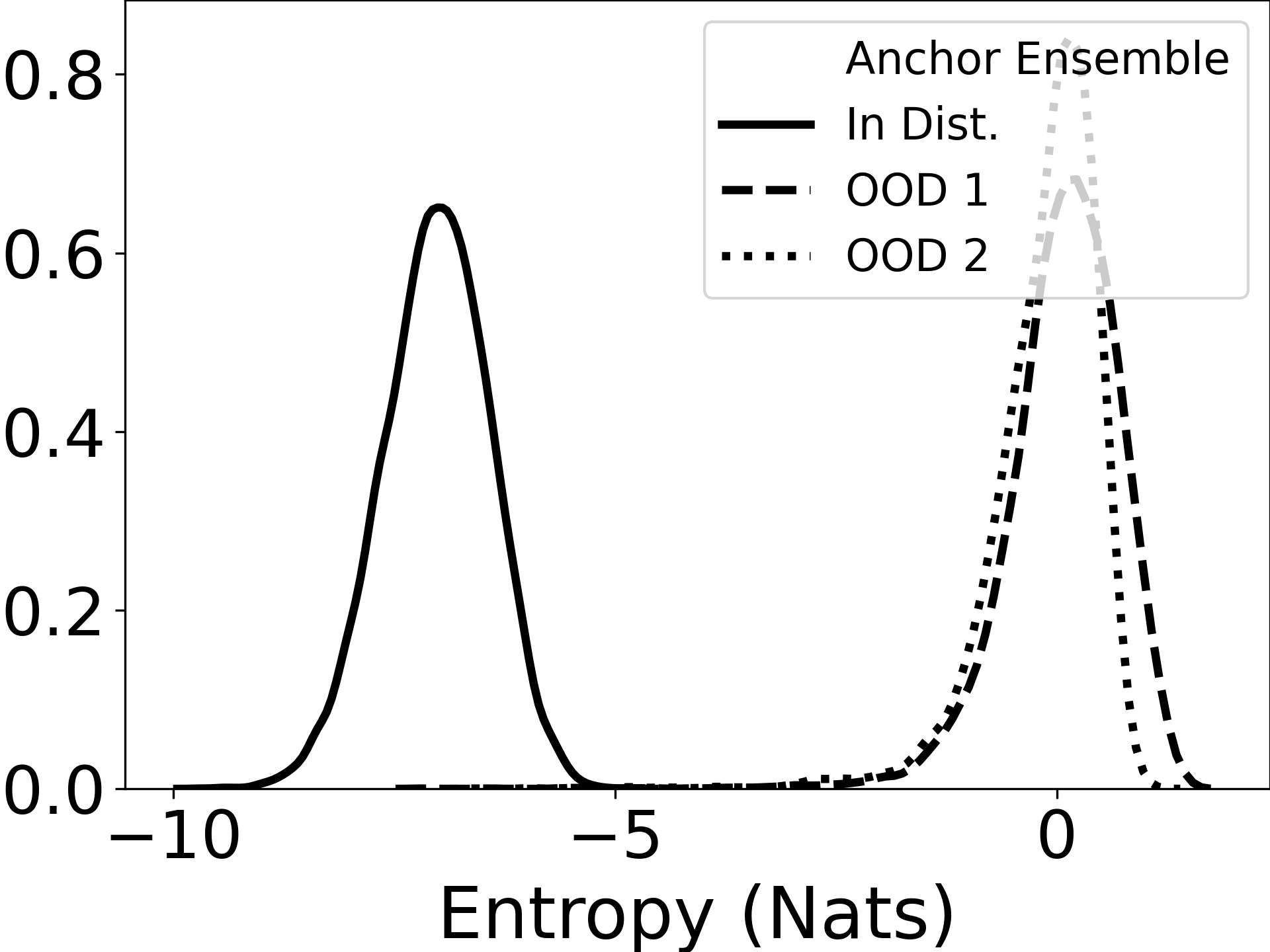}}
    ~
\subfloat{%
    \includegraphics[width=0.23\linewidth, valign=t]{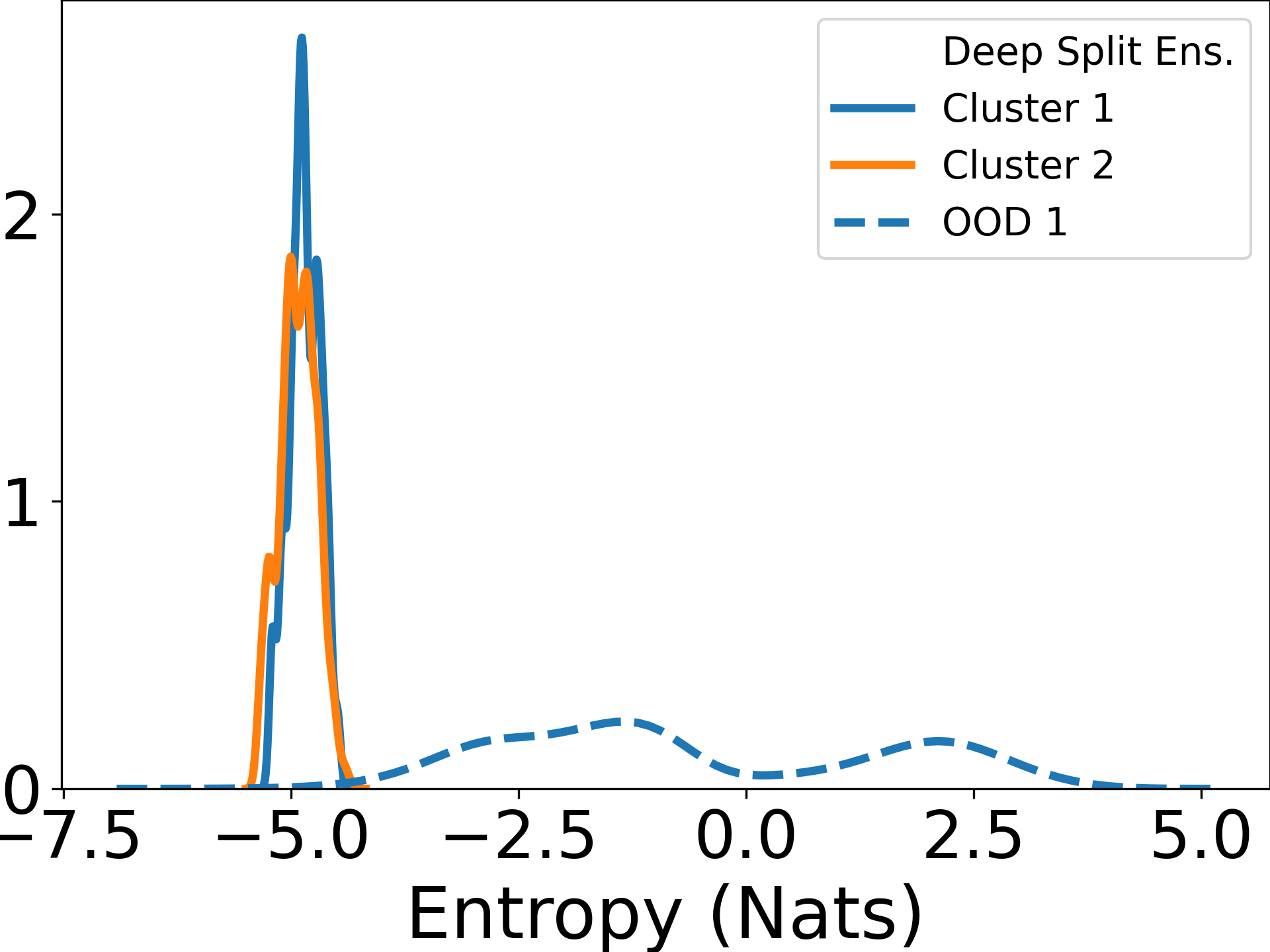}}
    ~
\subfloat{%
    \includegraphics[width=0.23\linewidth, valign=t]{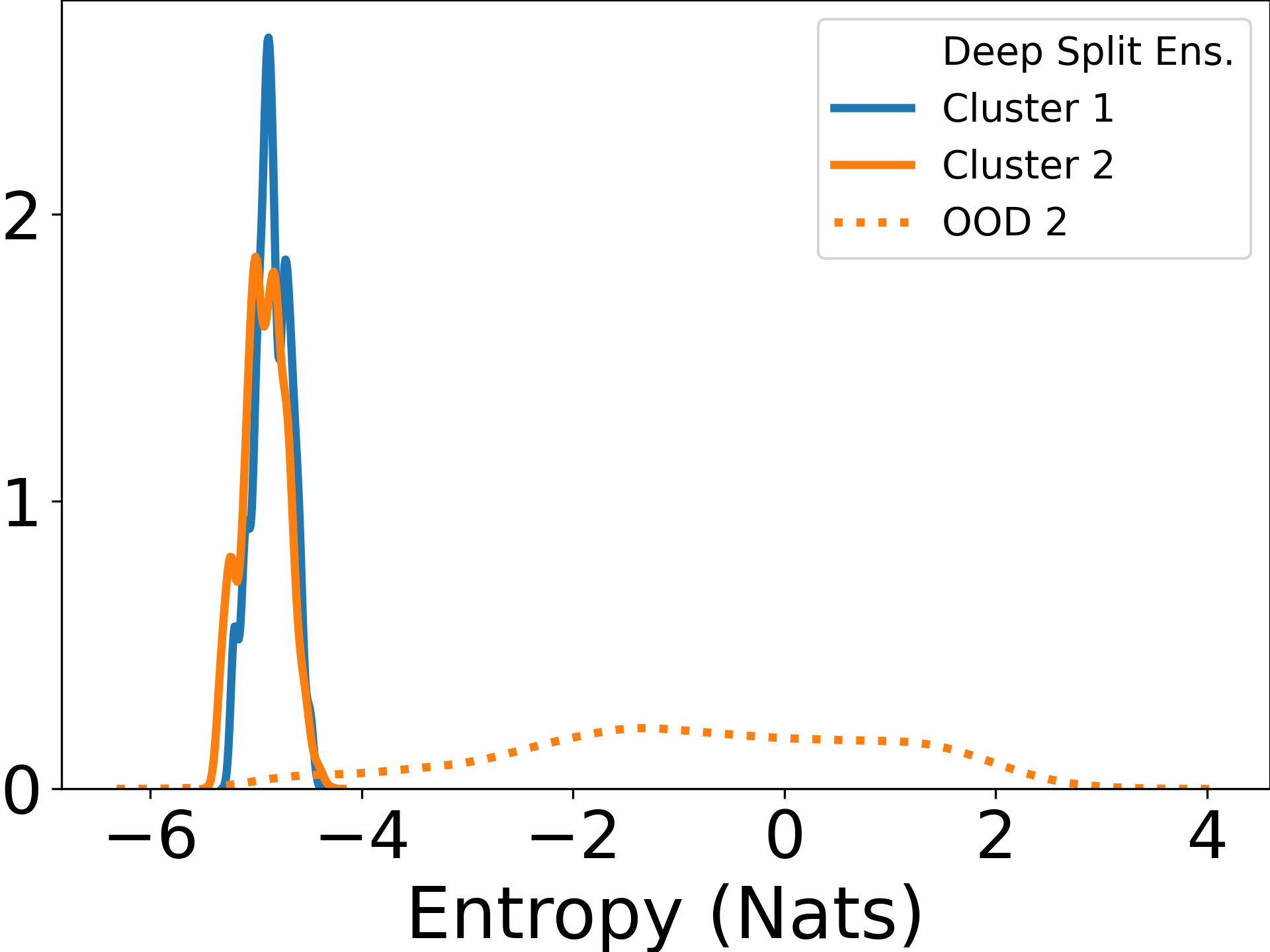}}

\caption{Entropy plots for `Boston', `Concrete', `Energy', `Kin8nm', and 'Naval' datasets using hierarchical clustering (Section \ref{training}). The first two columns show the kernel density estimation (KDE) of entropy for in- distribution i.e. $\mathcal{N}(0, 1)$ and out-of-distribution samples, obtained with unified uncertainty estimation using deep ensemble and anchored ensembling respetively. The last two columns show `cluster-wise' KDE of entropy for in-distribution and out-of-distribution samples, obtained with disentangled uncertainty estimation using deep split ensembles. OOD 1 and OOD 2 refer to introducing dataset shift by inducing noise sampled from $\mathcal{N}(6, 2^2)$ into 2 random input features; the features correspond to different clusters for deep split ensembles.}
\label{entropy_plots_appendix_hc}
\end{figure*}

\begin{figure*}[ht!]
\centering
\subfloat{%
    \includegraphics[width=0.01\linewidth, height=2.8cm, valign=t]{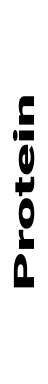}}\hspace{0.005mm}
    ~
\subfloat{%
    \includegraphics[width=0.23\linewidth, valign=t]{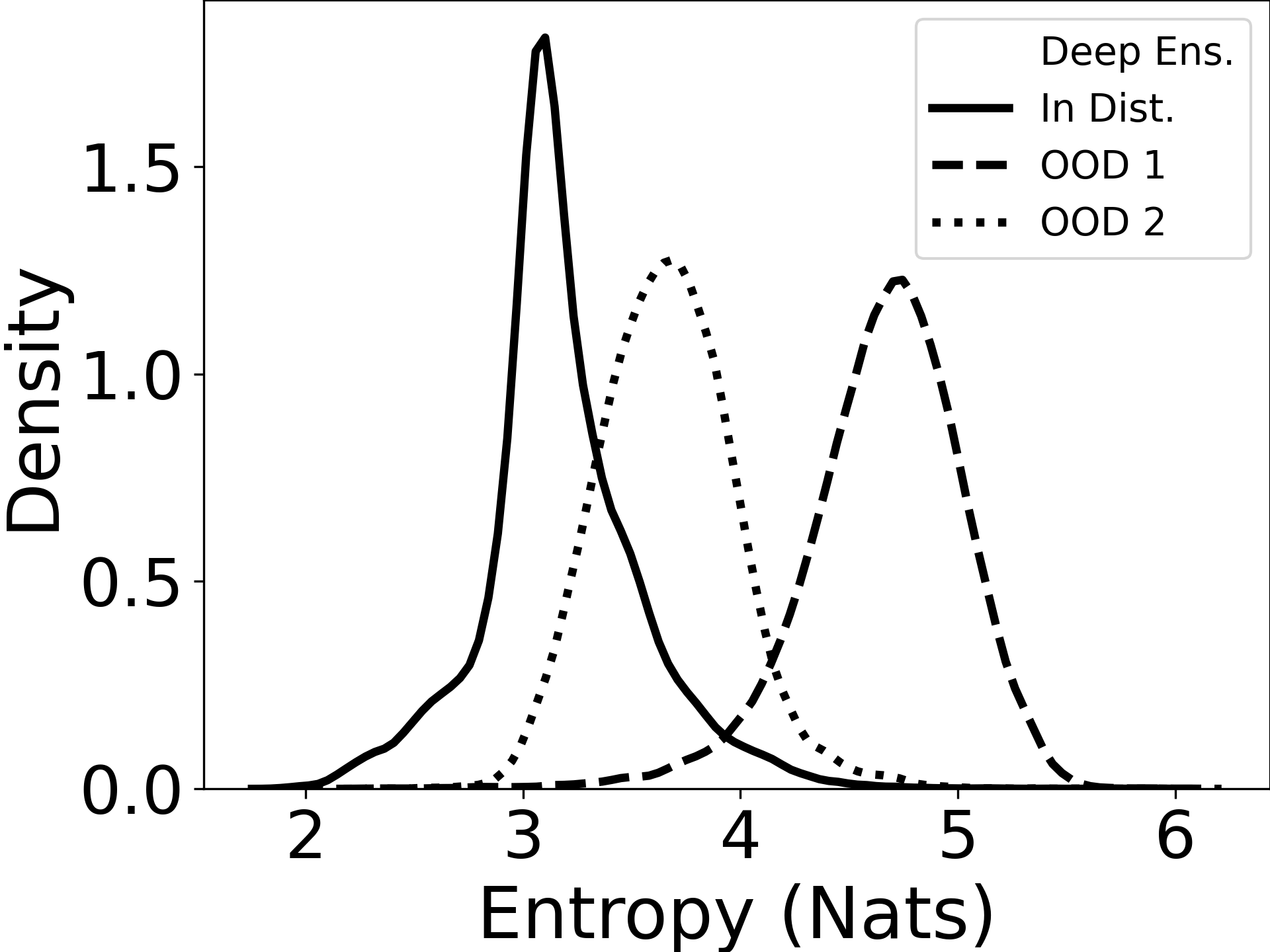}}
    ~
\subfloat{%
    \includegraphics[width=0.23\linewidth, valign=t]{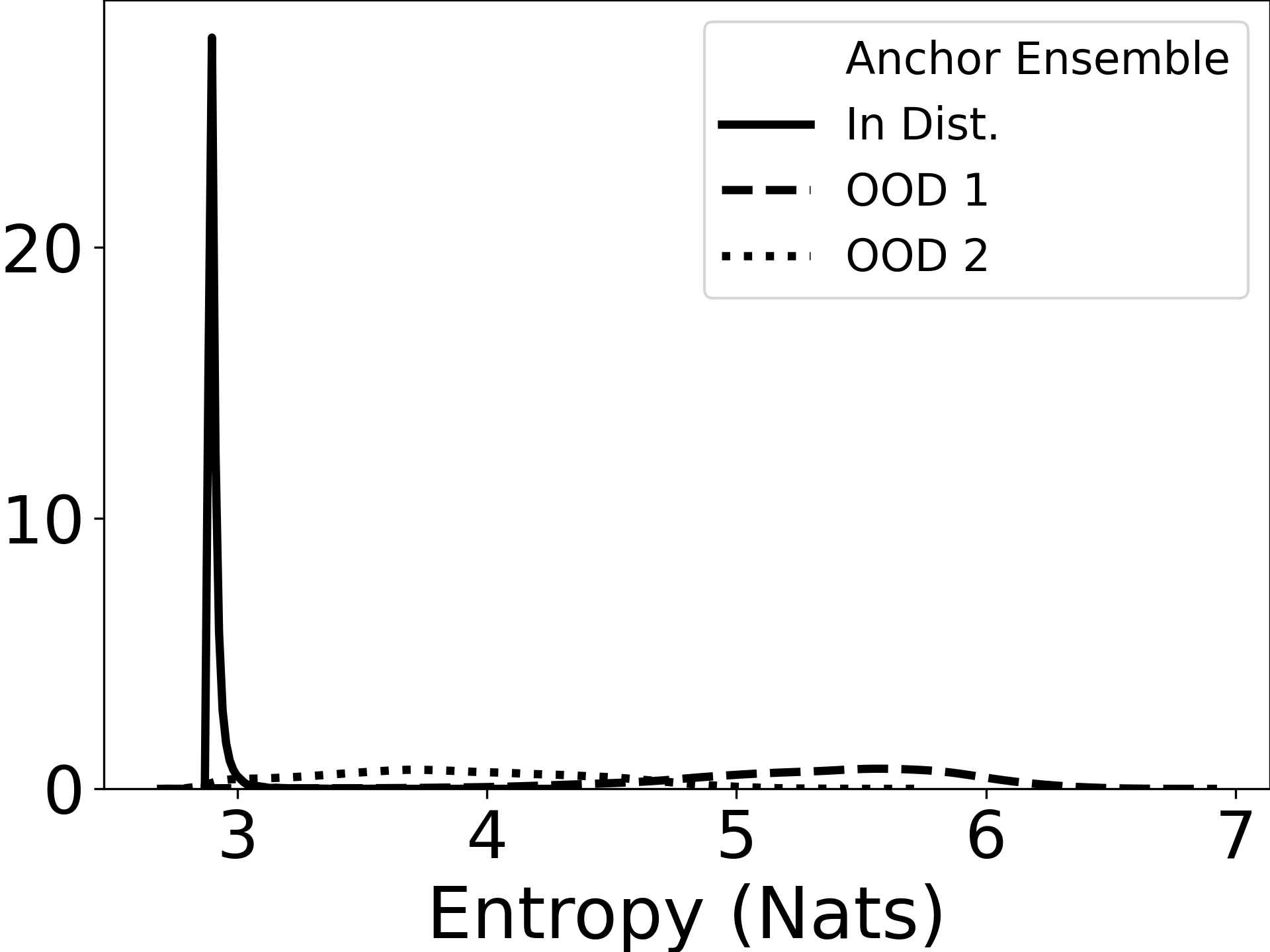}}
    ~
\subfloat{%
    \includegraphics[width=0.23\linewidth, valign=t]{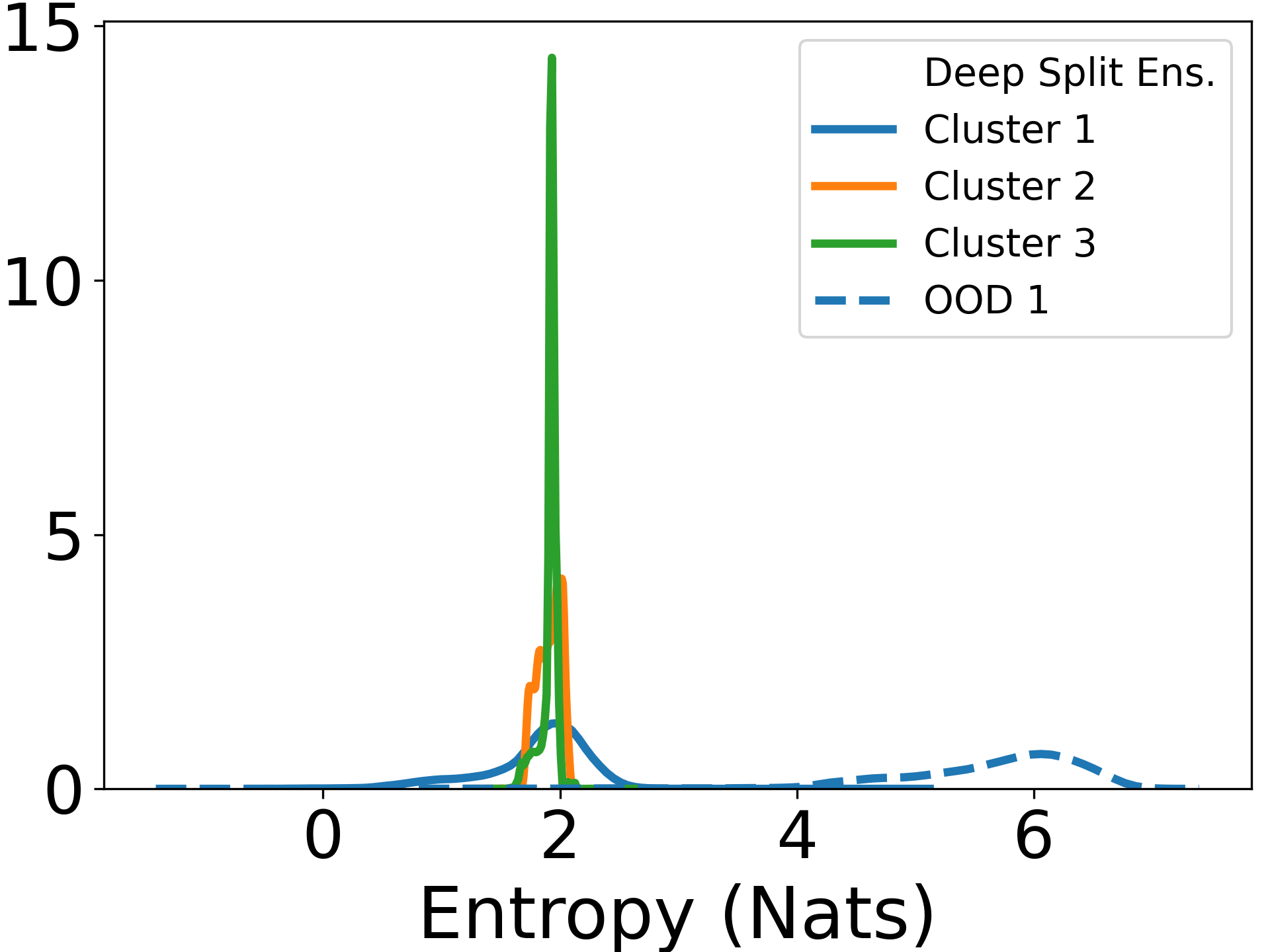}}
    ~
\subfloat{%
    \includegraphics[width=0.23\linewidth, valign=t]{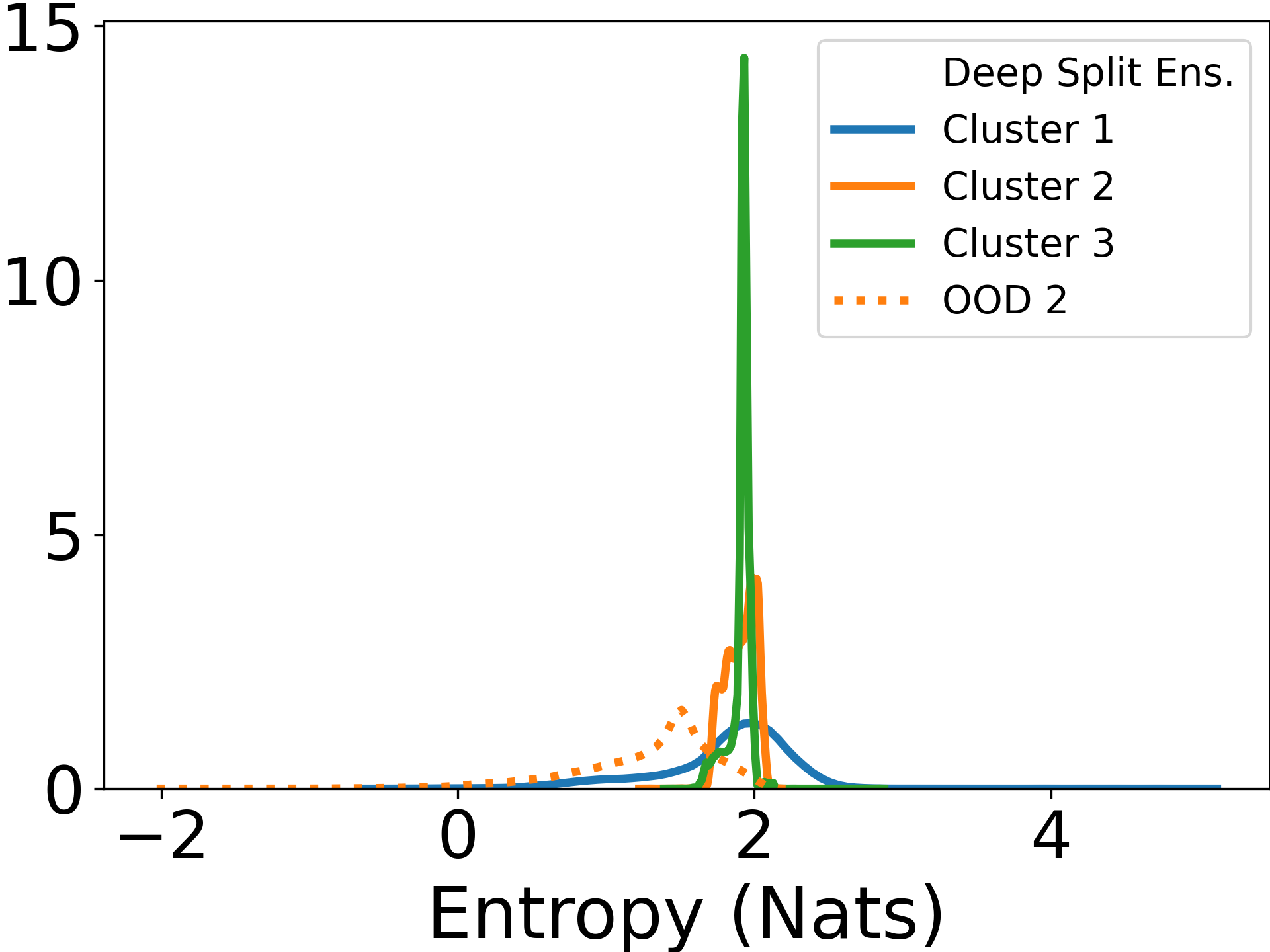}}
    ~ \\
    \subfloat{%
    \includegraphics[width=0.01\linewidth, height=2.8cm, valign=t]{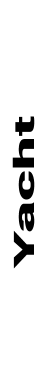}}\hspace{0.005mm}
    ~
\subfloat{%
    \includegraphics[width=0.23\linewidth, valign=t]{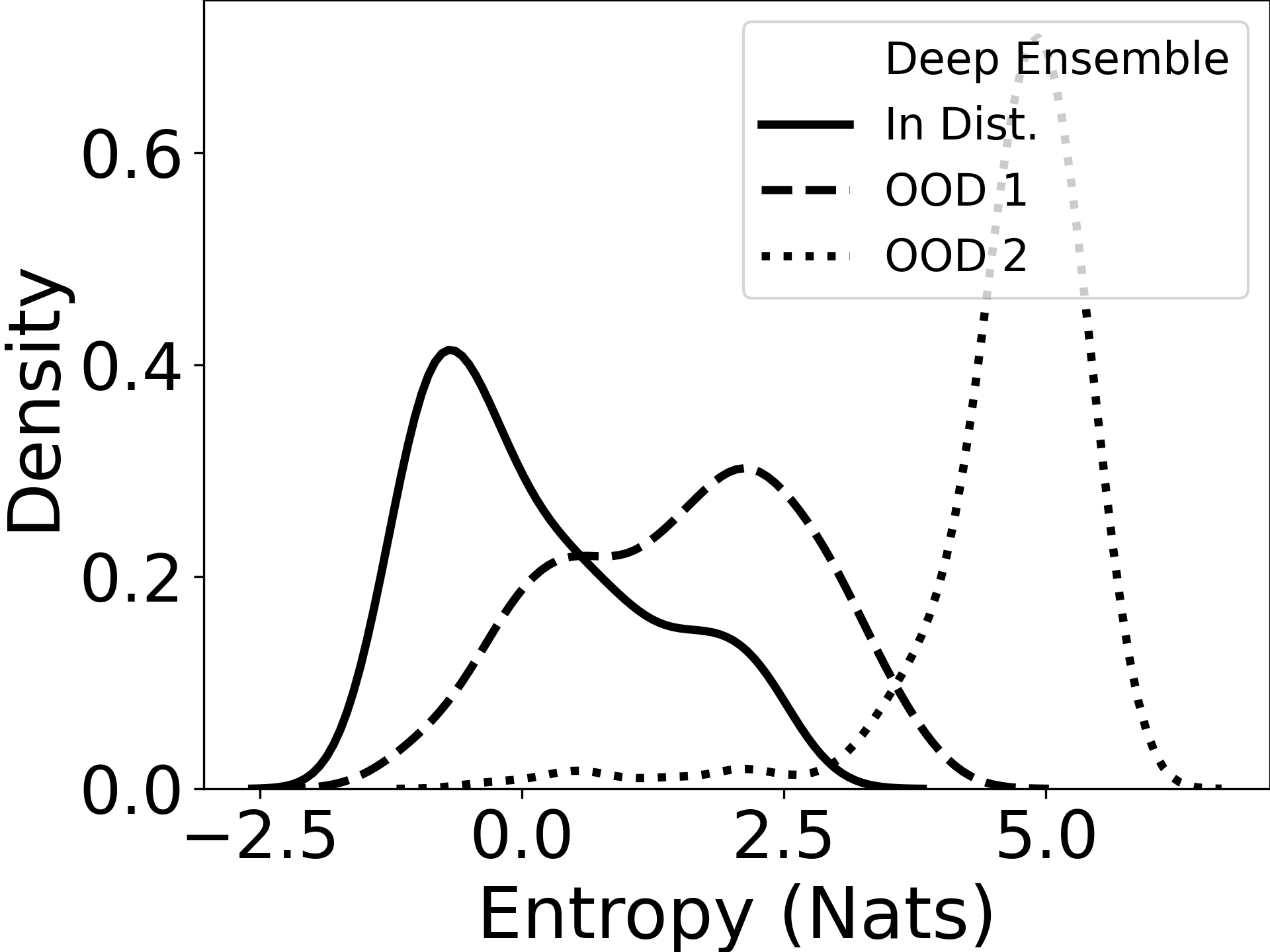}}
    ~
\subfloat{%
    \includegraphics[width=0.23\linewidth, valign=t]{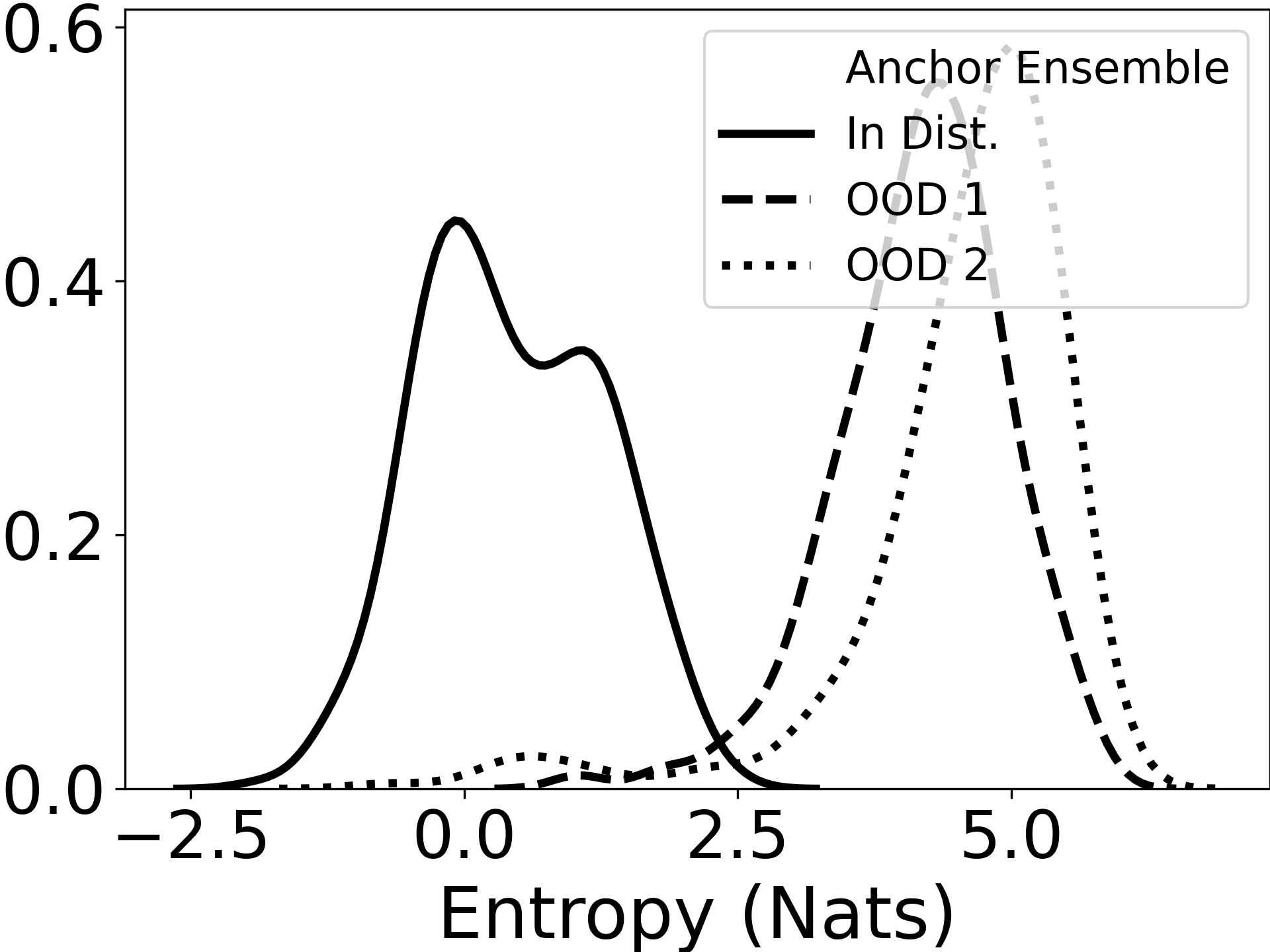}}
    ~
\subfloat{%
    \includegraphics[width=0.23\linewidth, valign=t]{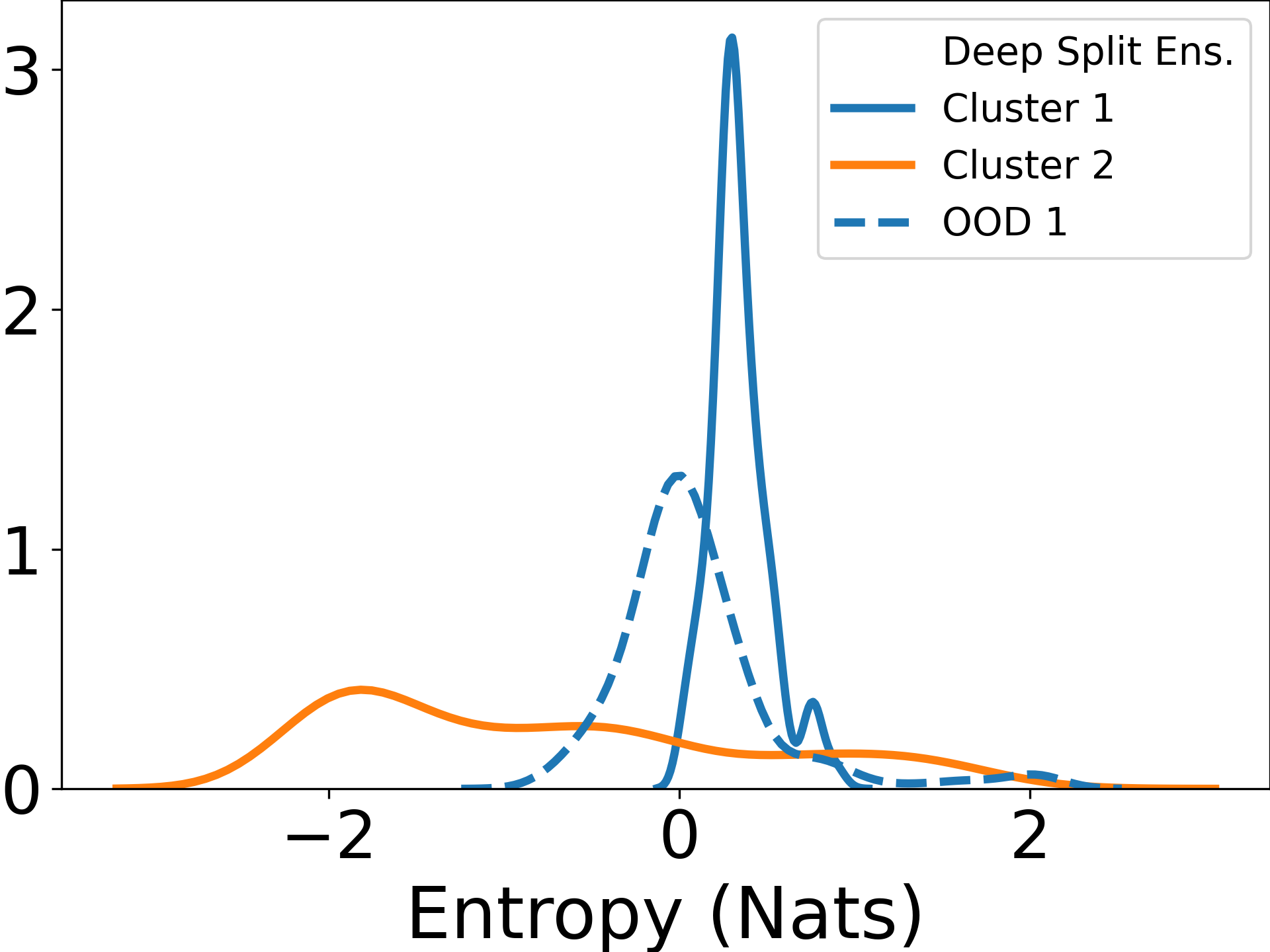}}
    ~
\subfloat{%
    \includegraphics[width=0.23\linewidth, valign=t]{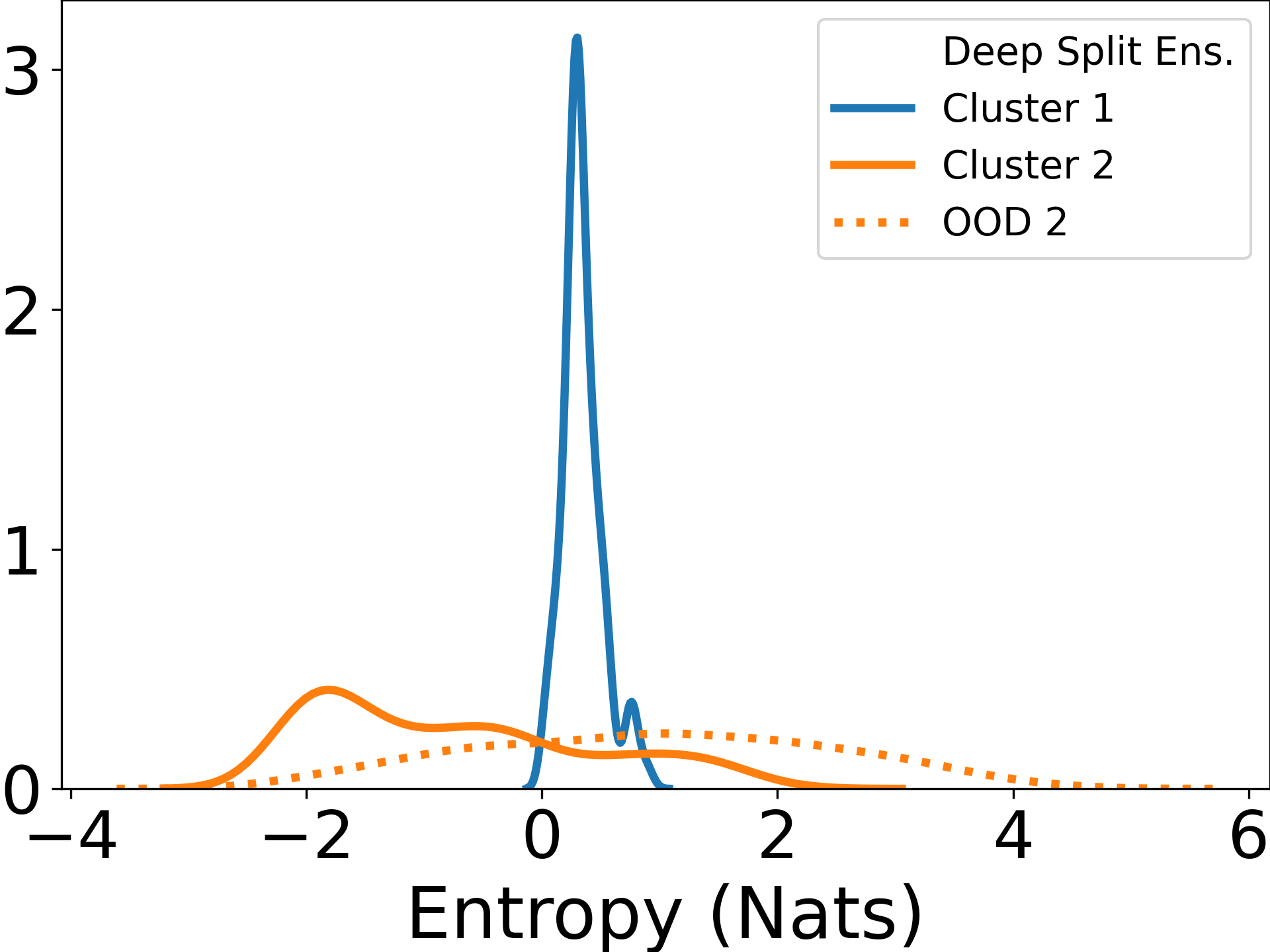}}
    ~ 
    \\
    \subfloat{%
    \includegraphics[width=0.01\linewidth, height=2.8cm, valign=t]{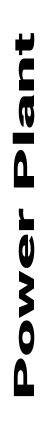}}\hspace{0.005mm}
    ~
\subfloat{%
    \includegraphics[width=0.23\linewidth, valign=t]{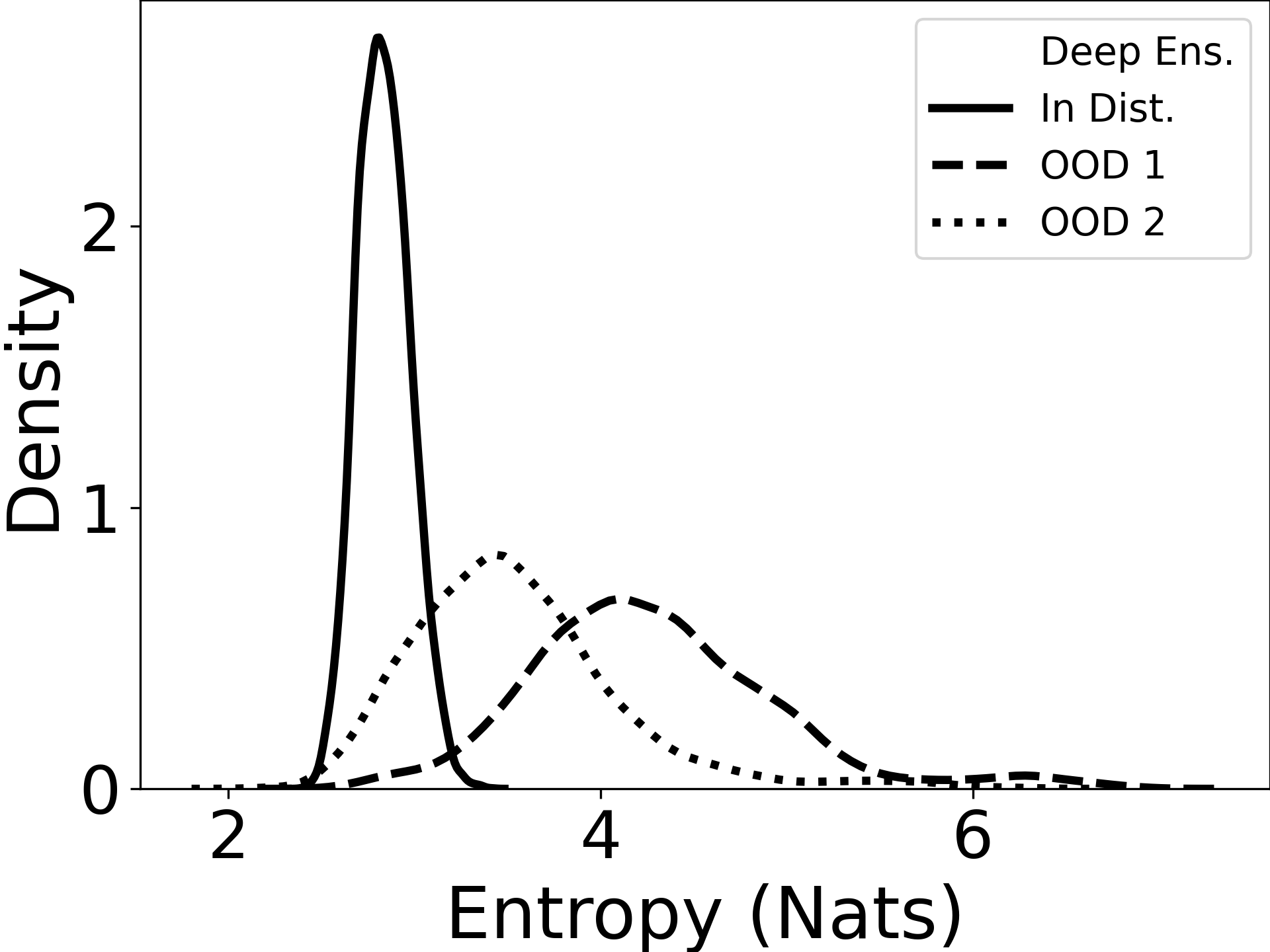}}
    ~
\subfloat{%
    \includegraphics[width=0.23\linewidth, valign=t]{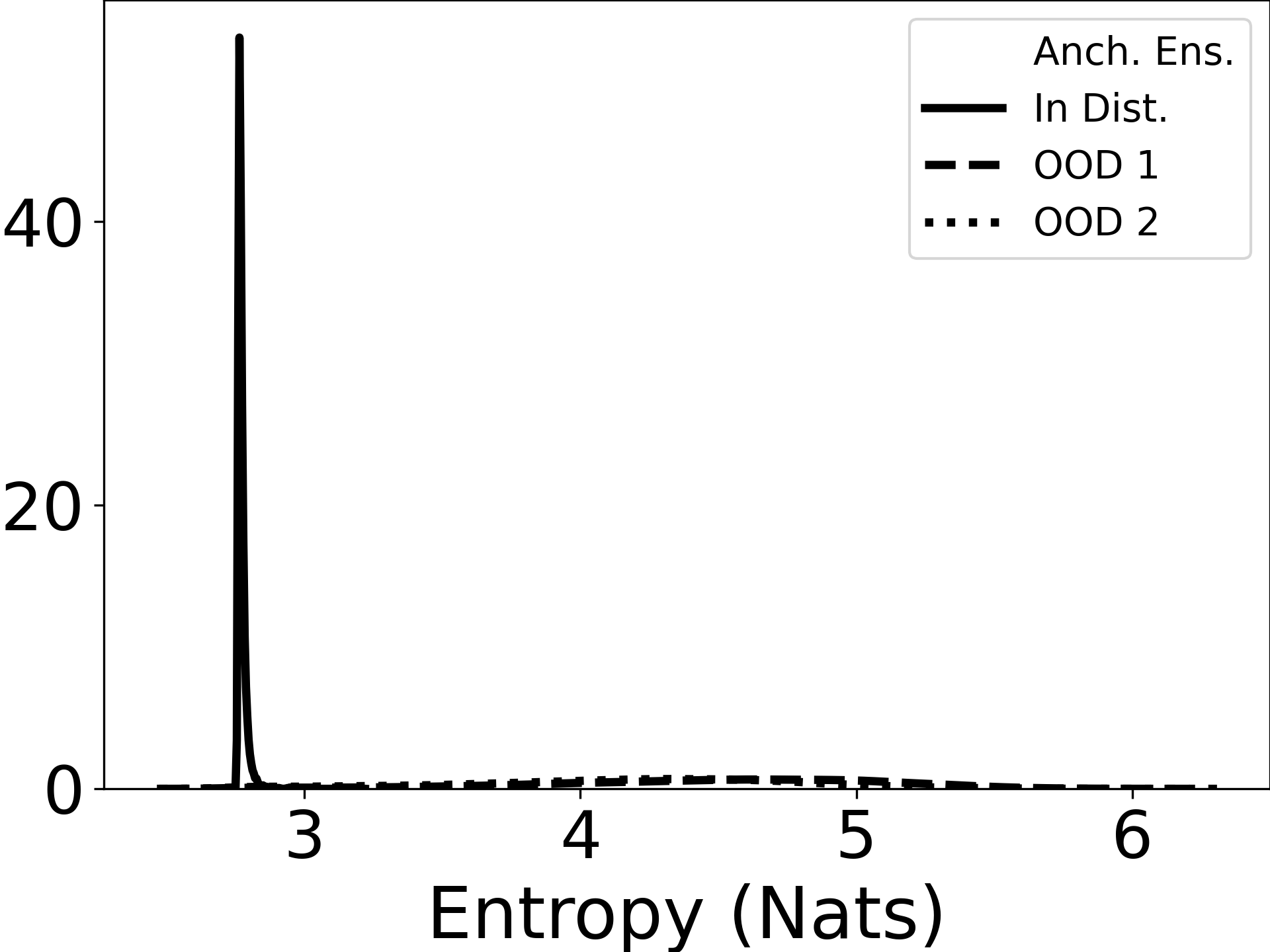}}
    ~
\subfloat{%
    \includegraphics[width=0.23\linewidth, valign=t]{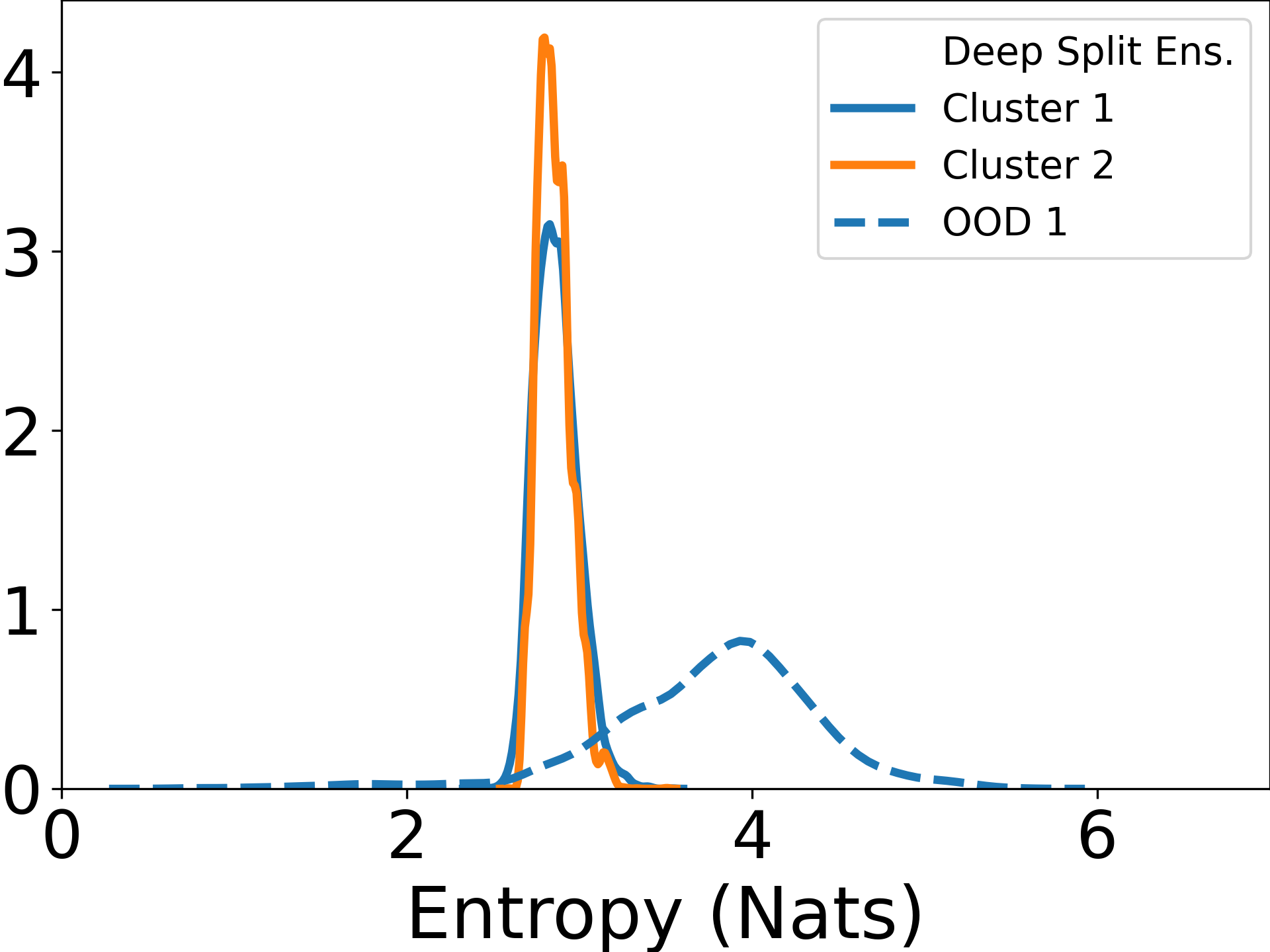}}
    ~
\subfloat{%
    \includegraphics[width=0.23\linewidth, valign=t]{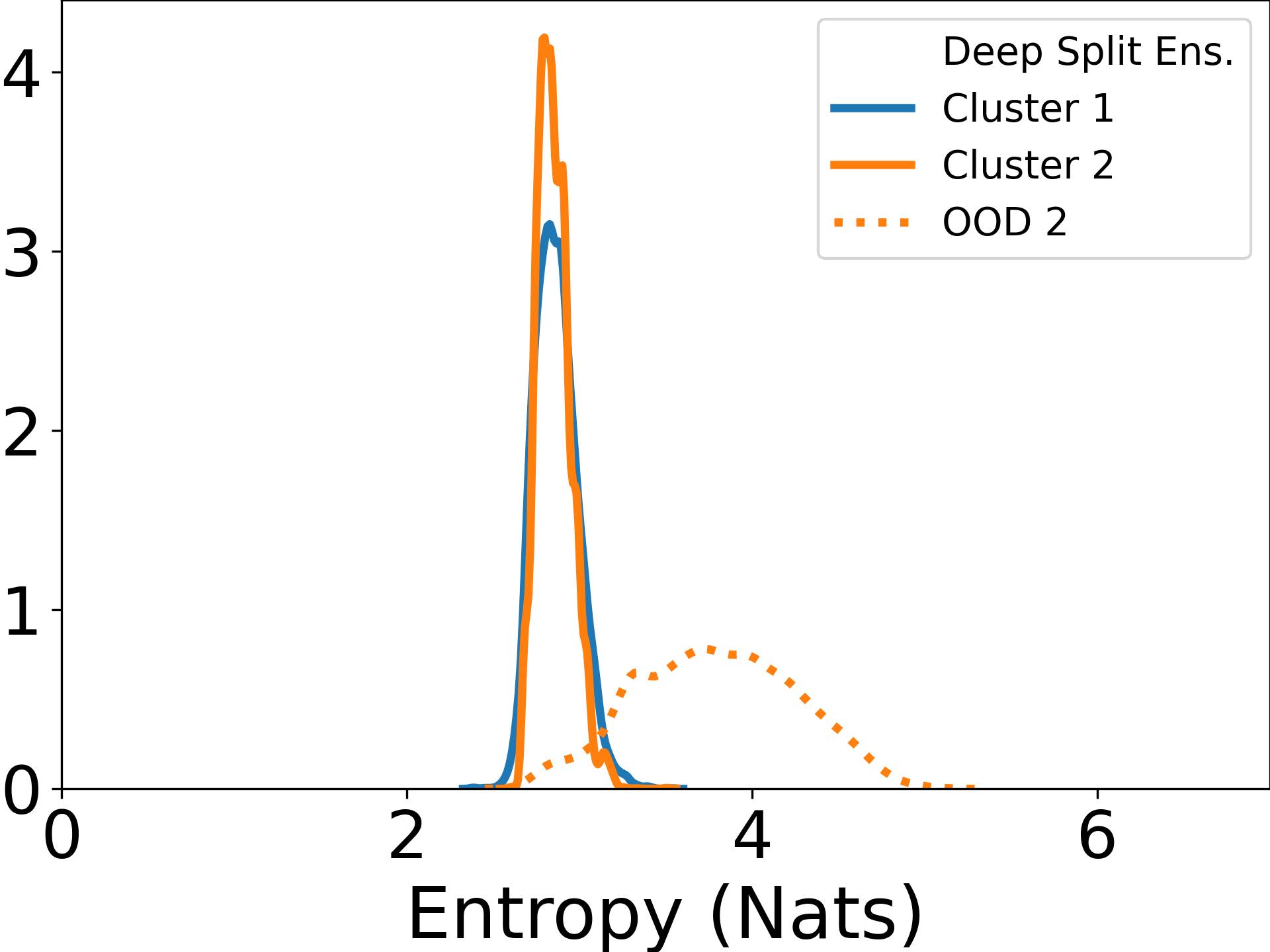}}
    ~ \\
\subfloat{%
    \includegraphics[width=0.01\linewidth, height=2.8cm, valign=t]{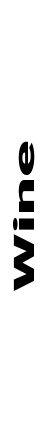}}\hspace{0.005mm}
    ~
\subfloat{%
    \includegraphics[width=0.23\linewidth, valign=t]{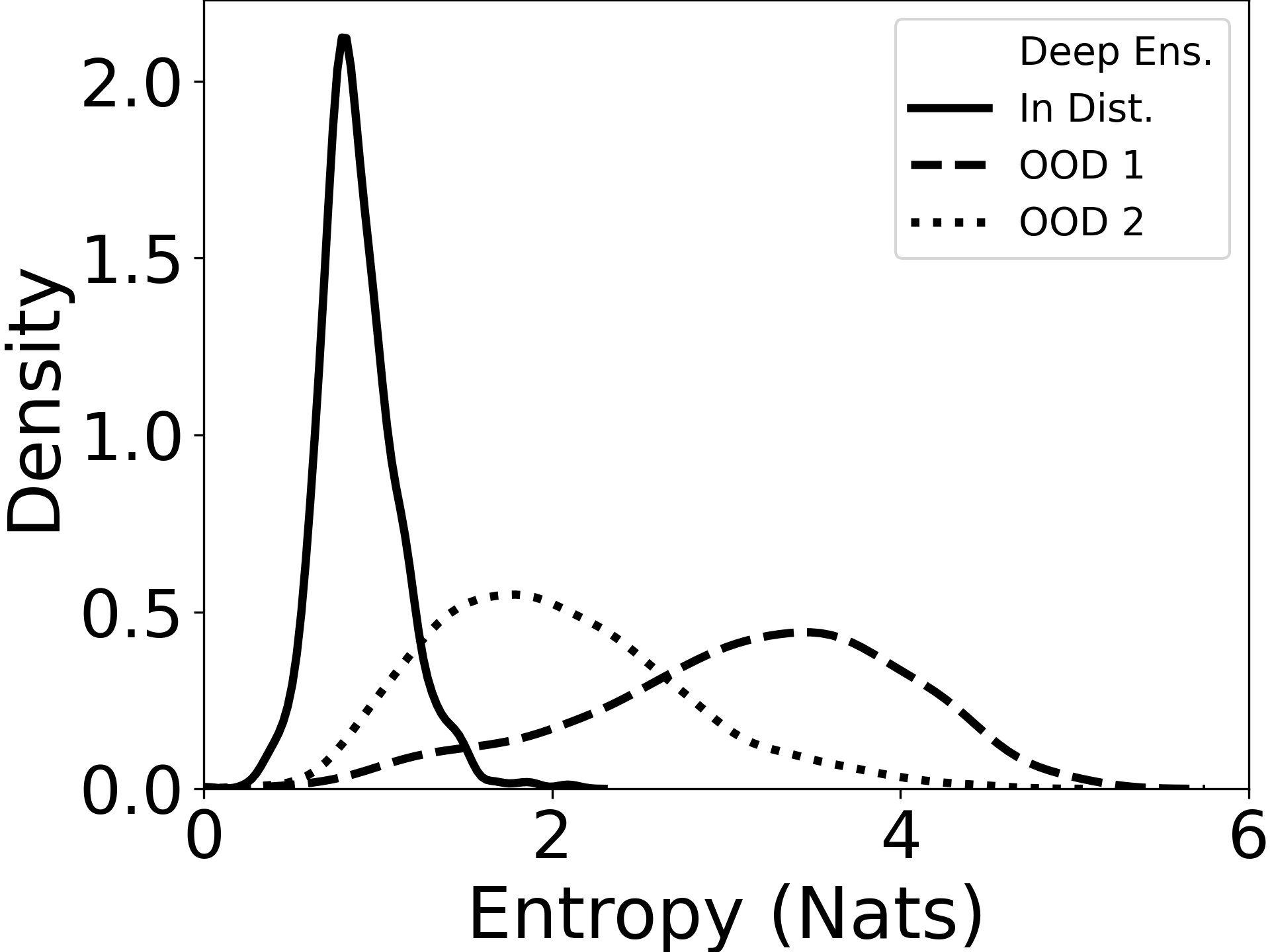}}
    ~
\subfloat{%
    \includegraphics[width=0.23\linewidth, valign=t]{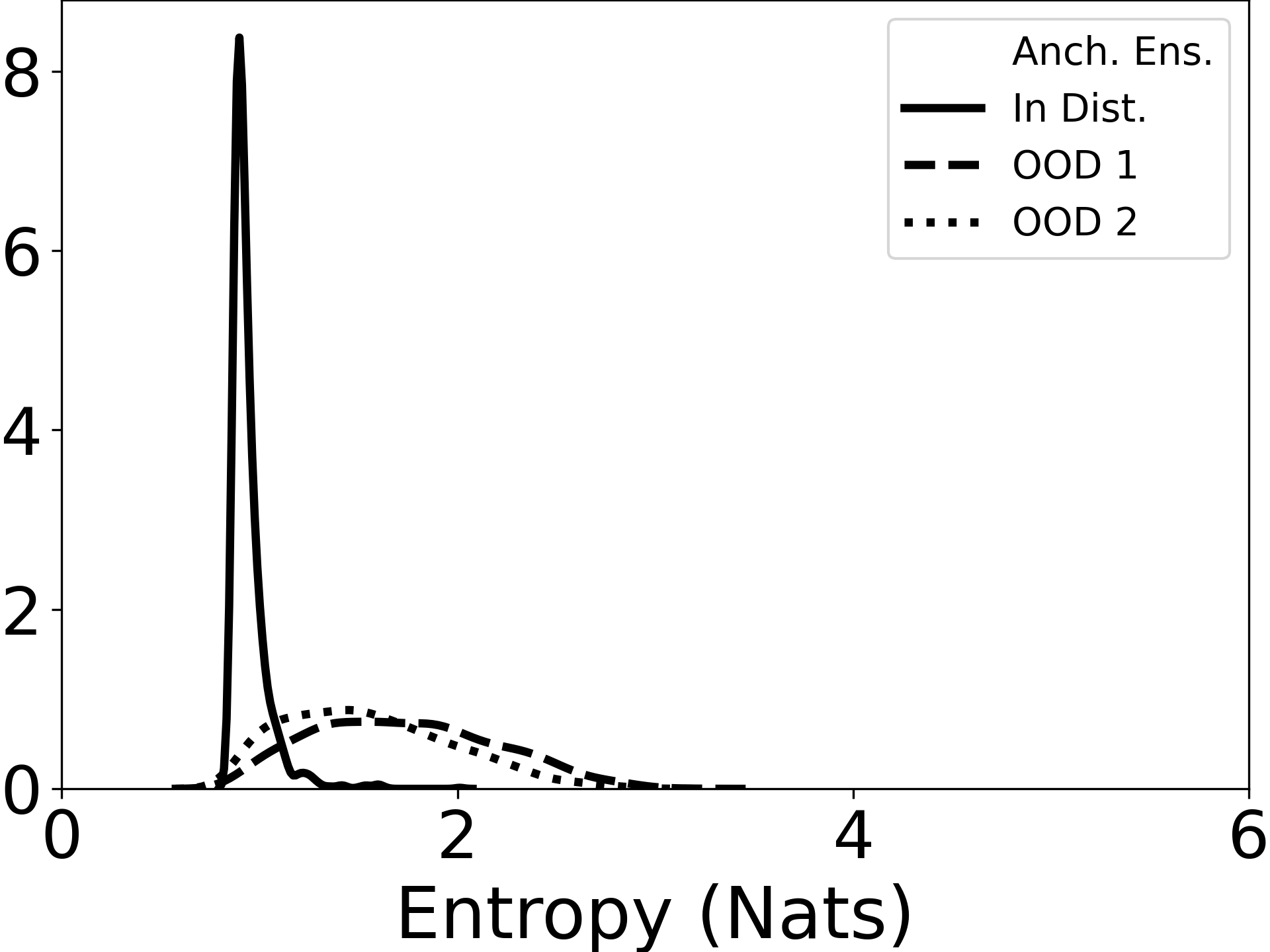}}
    ~
\subfloat{%
    \includegraphics[width=0.23\linewidth, valign=t]{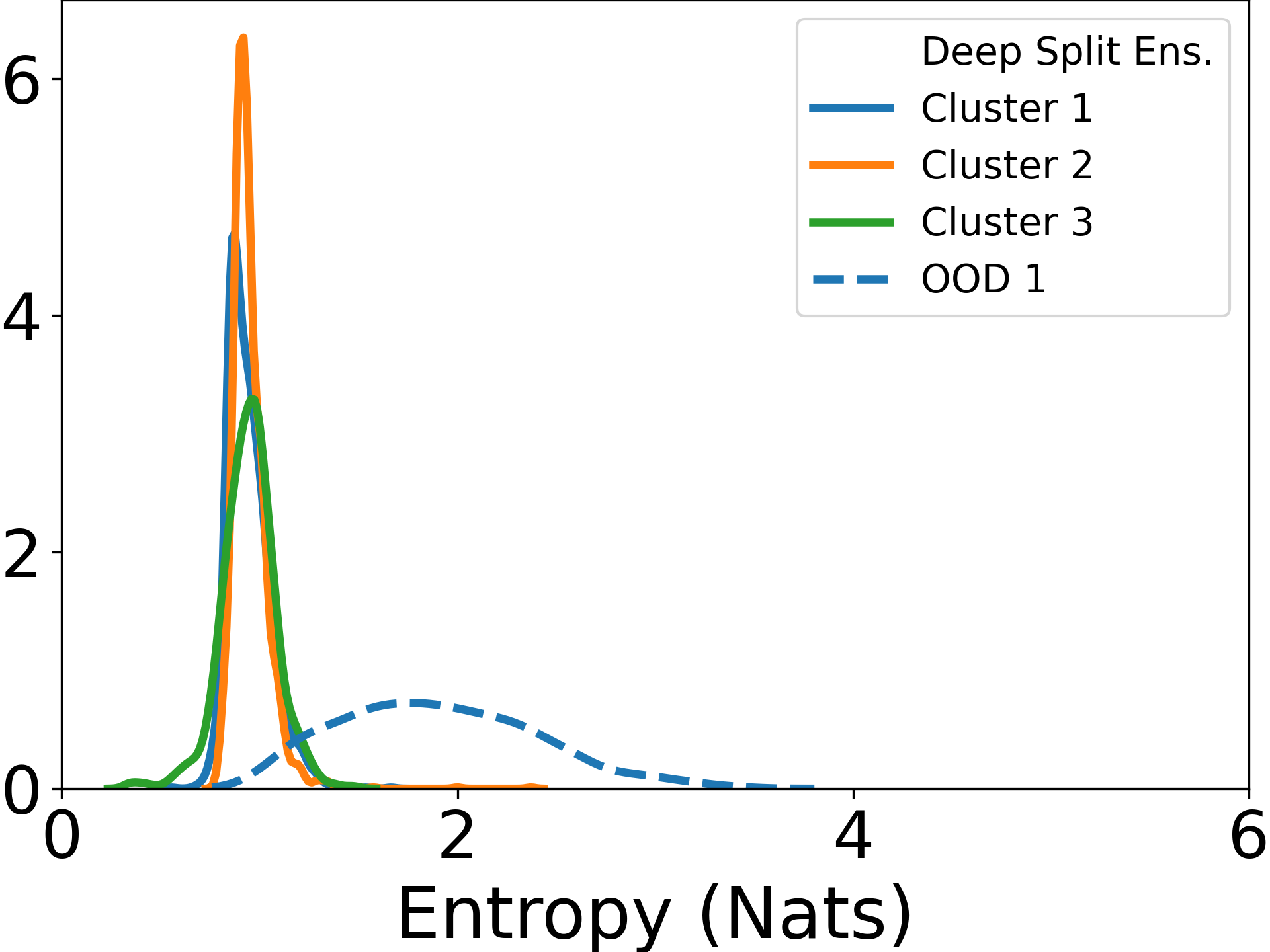}}
    ~
\subfloat{%
    \includegraphics[width=0.23\linewidth, valign=t]{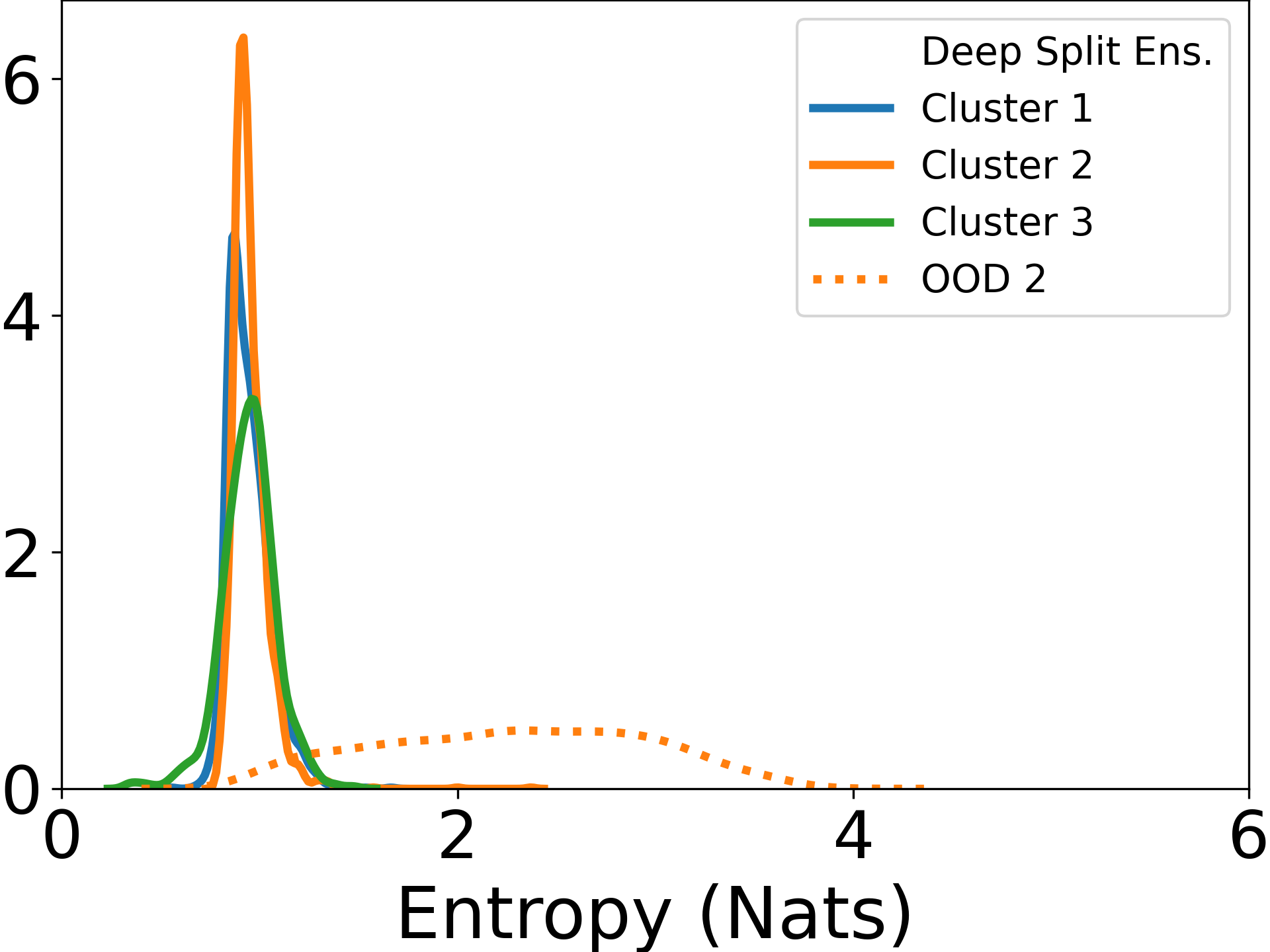}}

\caption{Entropy plots for 'Protein' and `Yacht' datasets using hierarchical clustering (Section \ref{training}), and 'Power Plant' and 'Wine' datasets using clusters from human experts (Section \ref{human}). The first two columns show the kernel density estimation (KDE) of entropy for in- distribution i.e. $\mathcal{N}(0, 1)$ and out-of-distribution samples, obtained with unified uncertainty estimation using deep ensemble and anchored ensembling respetively. The last two columns show `cluster-wise' KDE of entropy for in-distribution and out-of-distribution samples, obtained with disentangled uncertainty estimation using deep split ensembles. OOD 1 and OOD 2 refer to introducing dataset shift by inducing noise sampled from $\mathcal{N}(6, 2^2)$ into 2 random input features; the features correspond to different clusters for deep split ensembles.}
\label{entropy_plots_appendix_hc_2}
\end{figure*}

Figures \ref{entropy_plots_appendix_hc} and \ref{entropy_plots_appendix_hc_2} show the entropy plots for several datasets using hierarchical clustering and clusters from human experts. The first two columns show the kernel density estimation (KDE) of entropy for in- distribution i.e. $\mathcal{N}(0, 1)$ and out-of-distribution samples, obtained with unified uncertainty estimation using deep ensemble and anchored ensembling respetively. The last two columns show `cluster-wise' KDE of entropy for in-distribution and out-of-distribution samples, obtained with disentangled uncertainty estimation using deep split ensembles. OOD 1 and OOD 2 refer to introducing dataset shift by inducing noise sampled from $\mathcal{N}(6, 2^2)$ into 2 random input features; the features correspond to different clusters for deep split ensembles. We used the entire datasets by accumulating fold-wise test results. The shift in the KDE (for OOD samples) of the cluster-wise entropy shows that the deep split ensembles are well-calibrated.

\newpage

\subsection{65-95-99.7 rule (empirical rule) to assess calibration}\label{appendix_empirical}

\footnotetext{RMSE values for dVI are not shown as they were not reported in the paper.}

\begin{figure*}[ht!]
\centering
\subfloat{%
    \vspace{2cm}
    \includegraphics[width=0.02\linewidth, height=2.5cm, valign=t]{figures/entropy_plots/boston_title_new.png}}\hspace{0.005mm}
    ~  
\subfloat{%
    \includegraphics[width=0.31\linewidth, valign=t]{figures/empirical_rule_test/depc/boston_empirical_rule_test.png}}
    ~
\subfloat{%
    \includegraphics[width=0.31\linewidth, valign=t]{figures/empirical_rule_test/aepc/boston_empirical_rule_test.png}}
    ~
\subfloat{%
    \includegraphics[width=0.31\linewidth, valign=t]{figures/empirical_rule_test/dse/boston_empirical_rule_test.png}}
    \\
\subfloat{%
    \vspace{2cm}
    \includegraphics[width=0.02\linewidth, height=2.5cm, valign=t]{figures/entropy_plots/cement_title_new.png}}\hspace{0.005mm}
    ~  
\subfloat{%
    \includegraphics[width=0.31\linewidth, valign=t]{figures/empirical_rule_test/depc/cement_empirical_rule_test.png}}
    ~
\subfloat{%
    \includegraphics[width=0.31\linewidth, valign=t]{figures/empirical_rule_test/aepc/cement_empirical_rule_test.png}}
    ~
\subfloat{%
    \includegraphics[width=0.31\linewidth, valign=t]{figures/empirical_rule_test/dse/cement_empirical_rule_test.png}}
    \\
\subfloat{%
    \vspace{1.5cm}
    \includegraphics[width=0.02\linewidth, height=2.5cm, valign=t]{figures/entropy_plots/power_plant_title_new.png}}\hspace{0.005mm}
    ~  
\subfloat{%
    \includegraphics[width=0.31\linewidth, valign=t]{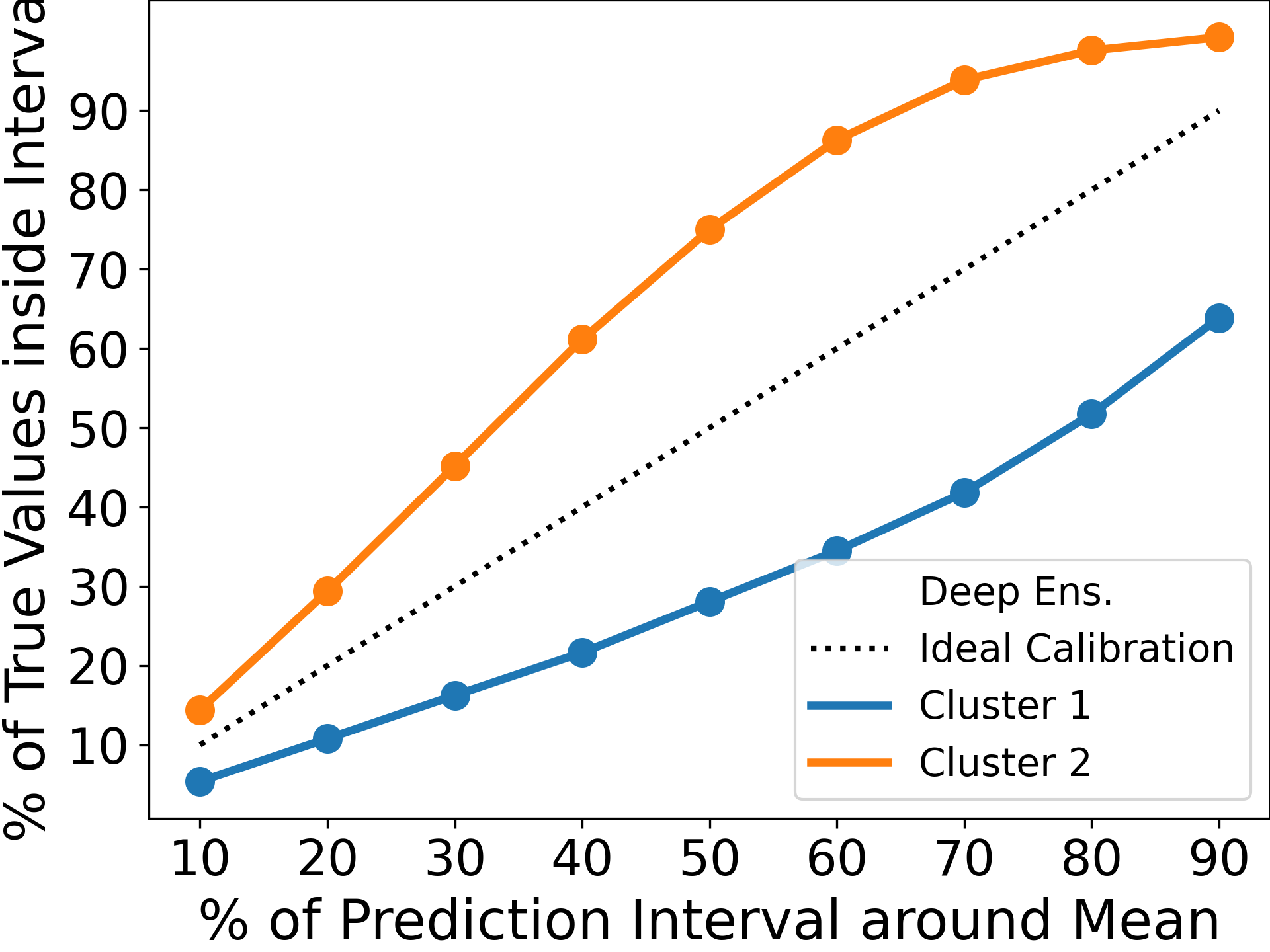}}
    ~
\subfloat{%
    \includegraphics[width=0.31\linewidth, valign=t]{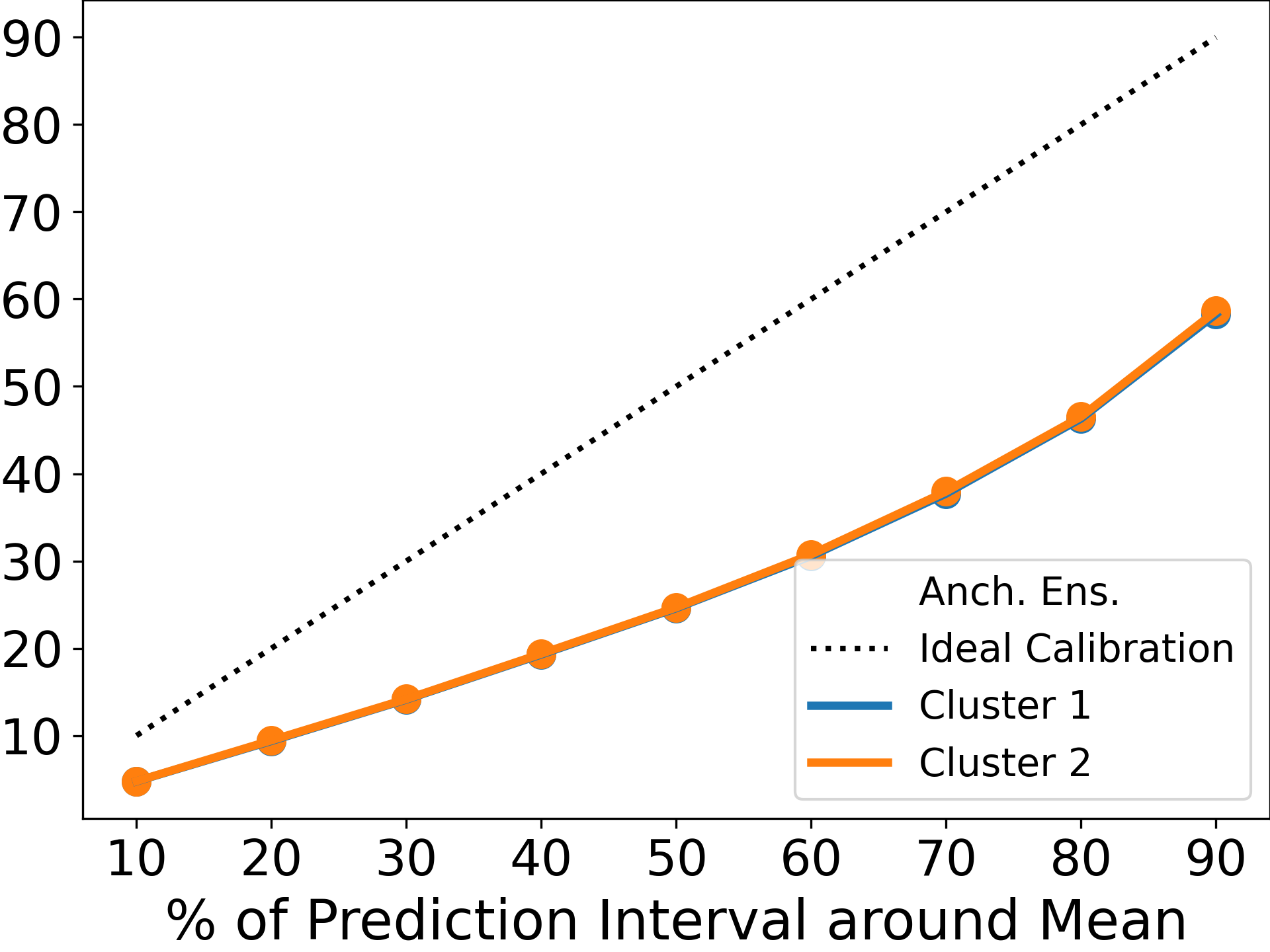}}
    ~
\subfloat{%
    \includegraphics[width=0.31\linewidth, valign=t]{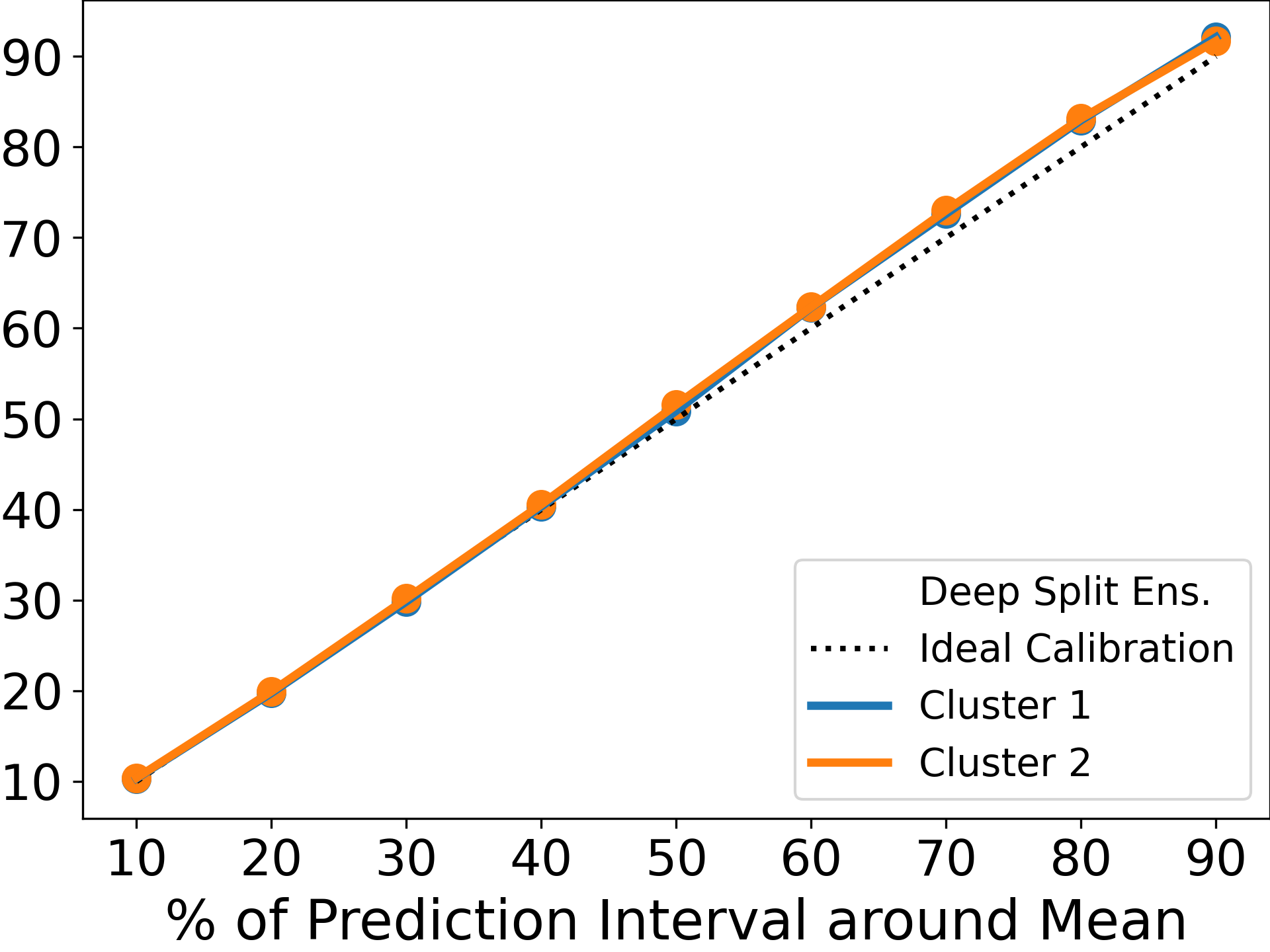}}
    \\
\subfloat{%
    \vspace{2cm}
    \includegraphics[width=0.02\linewidth, height=2.5cm, valign=t]{figures/entropy_plots/protein_title_new.png}}\hspace{0.005mm}
    ~  
\subfloat{%
    \includegraphics[width=0.31\linewidth, valign=t]{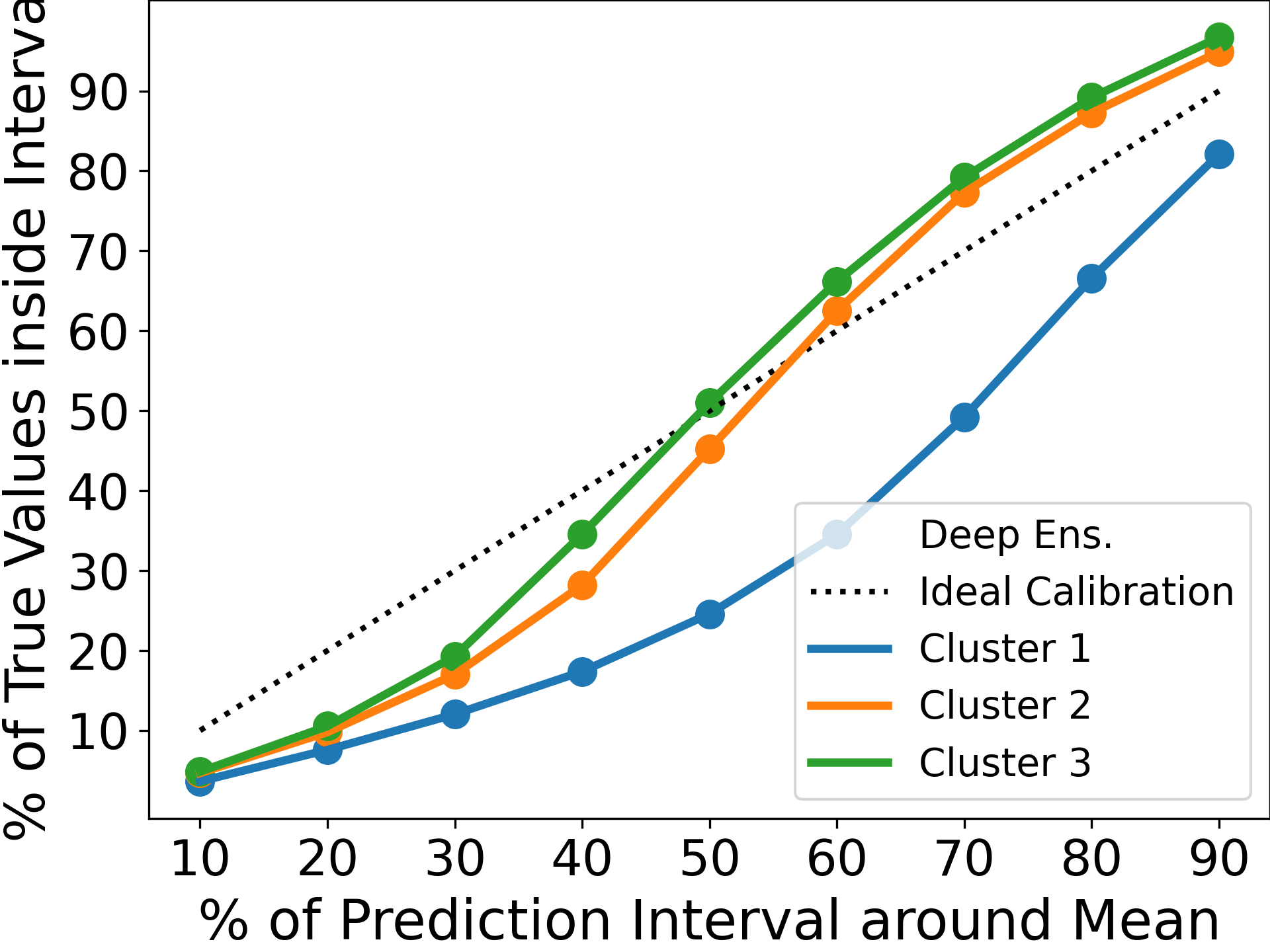}}
    ~
\subfloat{%
    \includegraphics[width=0.31\linewidth, valign=t]{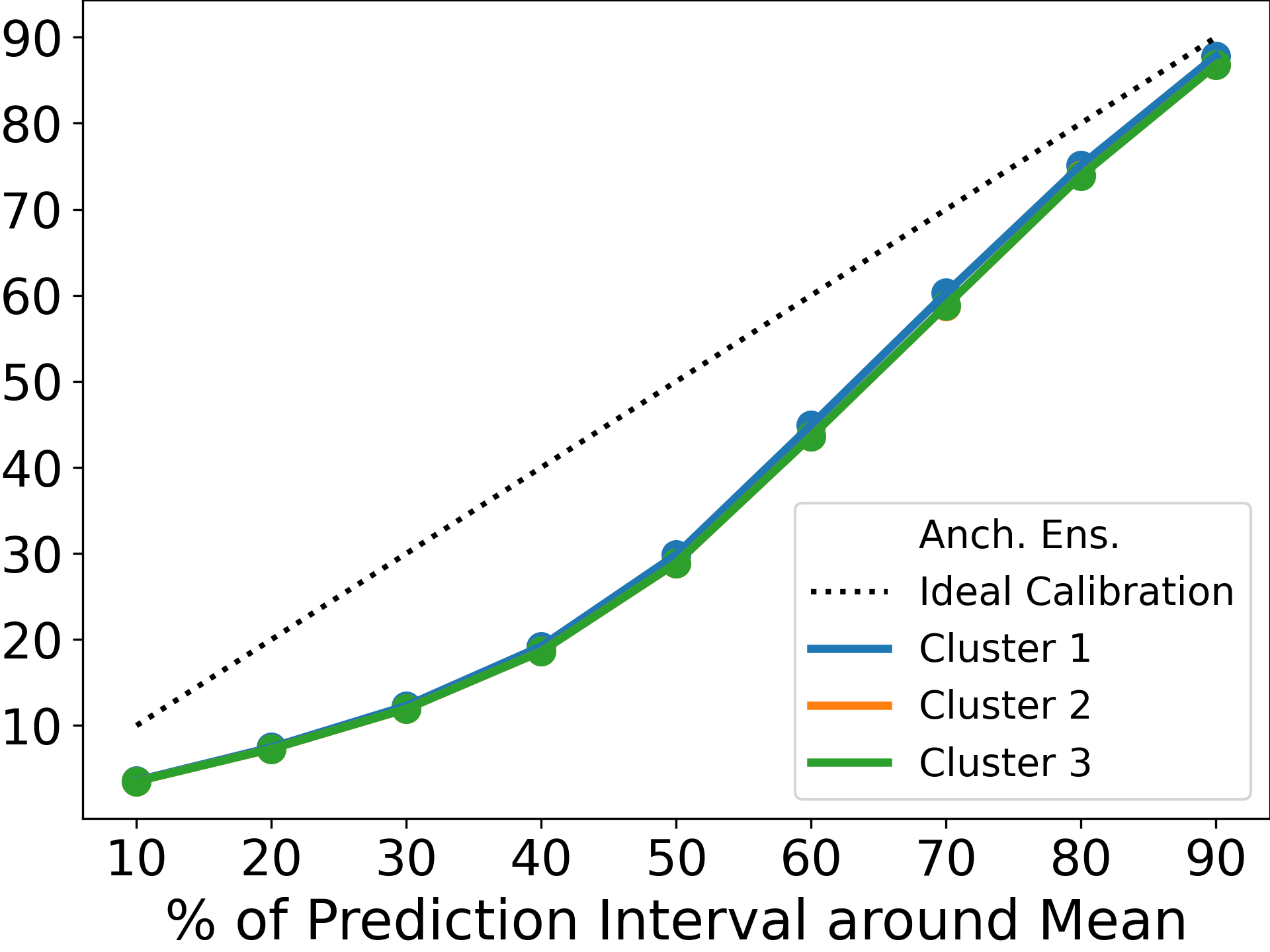}}
    ~
\subfloat{%
    \includegraphics[width=0.31\linewidth, valign=t]{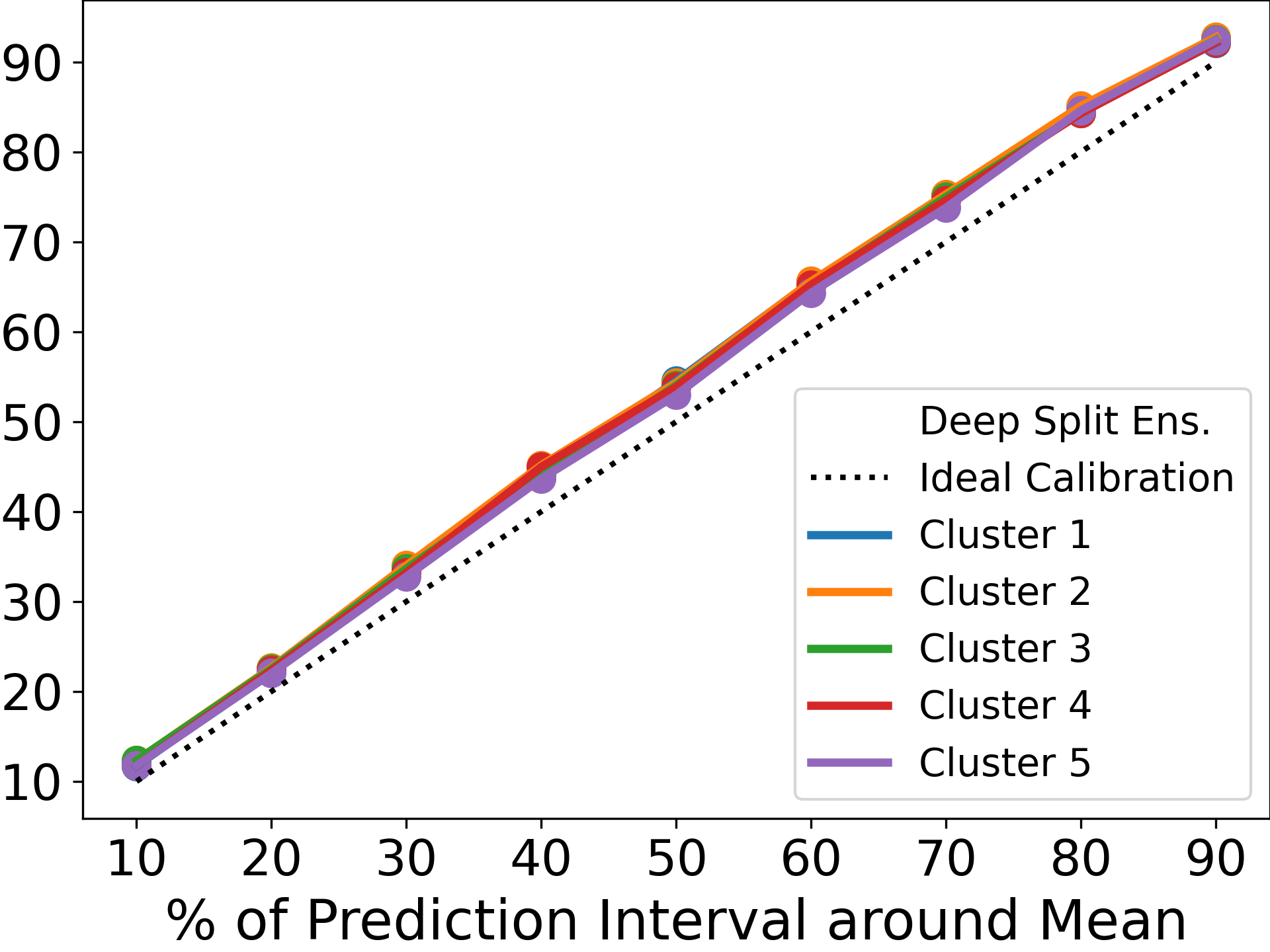}}
    \\
\subfloat{%
    \vspace{3cm}
    \includegraphics[width=0.02\linewidth, height=2.5cm, valign=t]{figures/entropy_plots/wine_title_new.png}}\hspace{0.005mm}
    ~   
\subfloat{%
    \includegraphics[width=0.31\linewidth, valign=t]{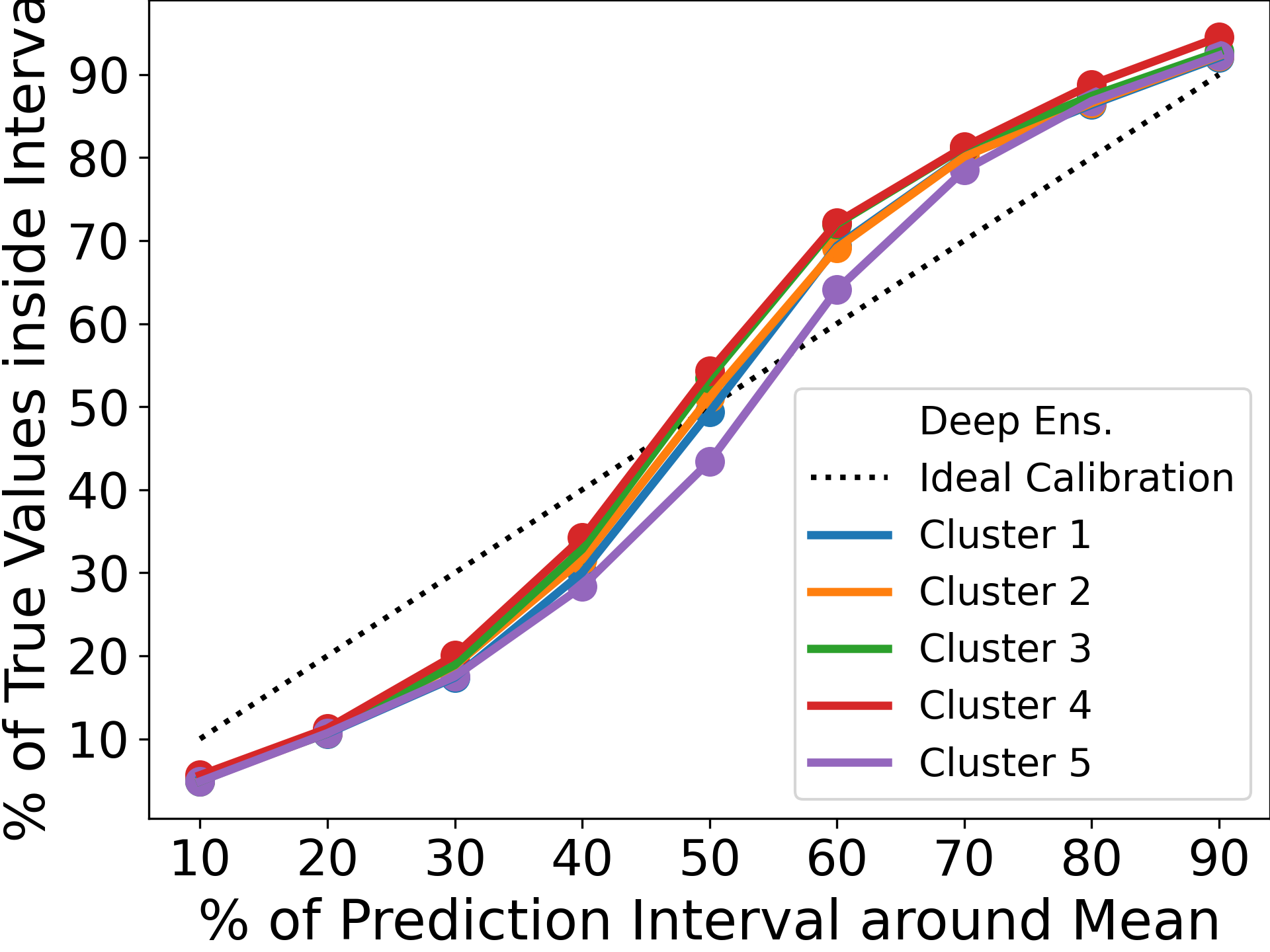}}
    ~
\subfloat{%
    \includegraphics[width=0.31\linewidth, valign=t]{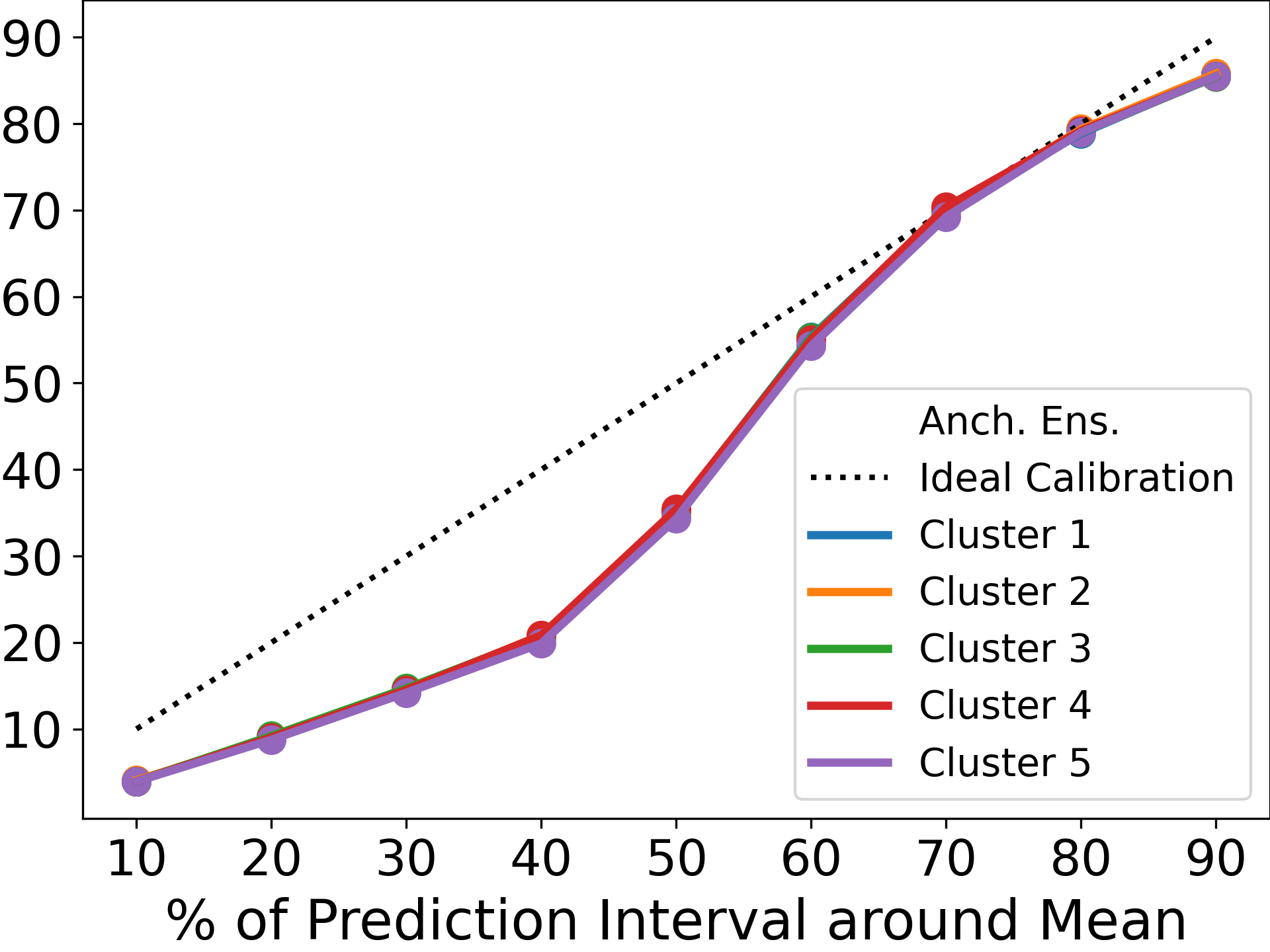}}
    ~
\subfloat{%
    \includegraphics[width=0.31\linewidth, valign=t]{figures/empirical_rule_test/dse/wine_empirical_rule_test.png}}
    ~

\caption{`Cluster-wise' calibration curves using empirical rule for `Boston', `Concrete', `Power', `Protein', and `Wine' datasets using hierarchical clustering (Section \ref{training}). The columns contain experiments using deep ensemble per input cluster (DEPC), anchored ensembling per input cluster (AEPC) and deep split ensembles respectively.}
\label{cal_quantile}
\end{figure*}

We evaluate to obtain calibration curves, where first, we compute the $x\%$ prediction interval for each test datapoint based on Gaussian quantiles using the predicted mean and variance. Then, we calculate the fraction of test observations (true values) that fall within this prediction interval. For a well-calibrated model, the observed fraction should be close to $x\%$ calculated earlier. To see how our models perform in this setting, we sweep from $x = 10\%$ to $x = 90\%$ in steps of 10, and consequently a line lying very close to the line ($y=x$) would indicate a well-calibrated model. Figure \ref{cal_quantile} shows the calibration curves for each feature cluster of DEPC, AEPC, and deep split ensembles on UCI datasets.

\end{document}